\documentclass[
    a4paper,
    doc,
    natbib
]{apa6}
\usepackage{authblk}

\usepackage[english]{babel}

\let \cite \parencite

\makeatletter

\usepackage[utf8]{inputenc} 
\usepackage[T1]{fontenc}    
\usepackage{url}            
\usepackage{booktabs}       
\usepackage{amsfonts}       
\usepackage{nicefrac}       
\usepackage{microtype}      

\usepackage[colorlinks, linkcolor=black, urlcolor=black, citecolor=black]{hyperref}

\usepackage{color}
\definecolor{black}{rgb}{0,0,0}
\usepackage{amsmath}
\usepackage{amsthm}
\usepackage{tikz}
\usetikzlibrary{decorations.pathreplacing,positioning, fit, arrows.meta, shapes}
\usetikzlibrary{shapes}
\usepackage{subcaption}
\usepackage{caption}
\usepackage{pgfplots}
\usepackage{booktabs}
\usepackage{xcolor}
\usepackage{colortbl}
\usepackage{multirow}
\usepackage{hyperref}
\usepackage{float}

\newcommand{\Figref}[1]{Figure~\ref{#1}}
\newcommand{\figref}[1]{Figure~\ref{#1}}

\newcommand{\Appref}[1]{SI}
\newcommand{\appref}[1]{SI}

\newcommand{\eg}[1]{(\ref{#1})}
\newcommand{\dasheg}[2]{\eg{#1}--\eg{#2}}

\newcommand{\tech}[1]{\emph{#1}}

\newcommand{\blue}[1]{#1}

\definecolor{ourgreen}{HTML}{4D8951}
\definecolor{darkblue}{HTML}{0499CC}
\definecolor{darkred}{HTML}{8B0000}
\definecolor{superlightgray}{HTML}{B8B8B8}

\newcommand{\softmax}{\mathbf{softmax}}
\DeclareMathOperator{\ReLU}{ReLU}

\newcommand{\LSTM}{\textbf{LSTM}}

\newcommand{\inputcolor}{black!20}
\newcommand{\outputcolor}{red!40}
\newcommand{\hiddencolor}{blue!40}
\newcommand{\hiddencolortwo}{orange!40}
\newcommand{\colorlabel}{green!40}
\newcommand{\compare}{\emph{Euclidean}}

\newcommand{\cite}{\citep}

\title{Relational reasoning and generalization using non-symbolic neural networks}
\shorttitle{Relational reasoning with neural networks}

\author[1]{Atticus Geiger}
\author[2]{Alexandra Carstensen}
\author[2]{Michael C.~Frank}
\author[1]{Christopher Potts }

\affil[1]{Stanford University Linguistics Department}
\affil[2]{Stanford University Psychology Department}

\affiliation{~}

\abstract{The notion of equality (identity) is simple and ubiquitous, making it a key case study for broader questions about the representations supporting abstract relational reasoning. Previous work suggested that neural networks were not suitable models of human relational reasoning because they could not represent mathematically identity, the most basic form of equality. We revisit this question. In our experiments, we assess out-of-sample generalization of equality using both arbitrary representations and representations that have been pretrained on separate tasks to imbue them with structure. We find neural networks are able to learn (1) basic equality (mathematical identity), (2) sequential equality problems (learning ABA-patterned sequences) with only positive training instances, and (3) a complex, hierarchical equality problem with only basic equality training instances (``zero-shot'' generalization). In the two latter cases, our models perform tasks proposed \blue{in previous work} to demarcate human-unique symbolic abilities. These results suggest that essential aspects of symbolic reasoning can emerge from data-driven, non-symbolic learning processes.}

\begin{document}

\maketitle

\newpage

\section{Introduction}\label{sec:introduction}

\newcommand{\papermode}{arxiv}

\makeatother

\newcommand{\newatticus}[1]{{\color{red}#1}}
\newcommand{\newmike}[1]{{\color{red}#1}}
\newcommand{\newalex}[1]{{\color{red}#1}}
\newcommand{\methref}{Methods}
\newcommand{\SIref}{our Appendix}
\newcommand{\SIrefnoour}{Appendix}

One of the key components of human intelligence is our ability to reason about abstract relations between stimuli. Many of the most unremarkable human activities -- scheduling a meeting, following traffic signs, assembling furniture -- require a fluency with abstraction and relational reasoning that is unmatched in nonhuman animals. An influential perspective on human uniqueness holds that relational concepts are critical to higher-order cognition \citep[e.g.,][]{gentner:2003}. By far the most common case study of abstract relations has been equality.\footnote{We use the term ``equality'' here, though different literatures have also used ``identity.''} Equality is a valuable case study because it is simple and ubiquitous, but also completely abstract in the sense that it can be evaluated regardless of the identity of the stimuli being judged.

Human equality reasoning has been studied extensively across a host of systems and tasks, with wildly variant conclusions. In some studies, equality is very challenging to learn: only great apes with either extensive language experience or specialized training succeed in matching tasks in which a \emph{same} pair, AA, must be matched to a novel same pair, BB \cite{Premack:1983,thompson:2001}. Preschool children also struggle to learn these regularities in a seemingly similar task \citep{walker:2016}. In contrast, other studies suggest that equality is simple: bees are able to learn abstract identity relationships from only a small set of training trials \cite{giurfa:2001,avargues:2011}, and human infants can generalize identity patterns \cite{anderson:2018} and succeed in relational matching tasks \cite{ferry:2015}. We take the central challenge of this literature to be characterizing the conditions that lead to success or failure in learning an abstract relation in a way that can be productively generalized to new stimuli \citep{carstensen:inpress}.

The learning task in all of these cases can be described using the predicate \emph{same} (or equivalently, =), which operates over two inputs and returns {\sc true} if they are identical in some respect, else {\sc false}. One perspective in the literature is that success in these learning tasks implies the presence of an equivalent symbolic description in the mind of the solver \cite{marcus:1999,Premack:1983}. This view does not provide a lever to distinguish which of these tasks are trivial and which are difficult, however. Further, it can fall prey to circularity: because newborns show sensitivity to identity relations \cite{gervain:2012}, then it would follow from this argument that they must have symbolic representations. If this logic applies also to bees, then we presuppose symbolic representations universally and have no account of the gradient difficulty of different tasks for different species.

An explanation of when same--different tasks are trivial and when they are difficult requires a theoretical framework beyond the symbolic/non-symbolic distinction. To make quantitative predictions about task performance, such a framework should ideally be instantiated in a computational model that takes in training data and learns a solution that generalizes when assessed with stimuli analogous to those used in experimental assessments. Symbolic computational models \cite[e.g.,][]{frank:2011} can be used to make contact with data about the breadth of generalizations that humans make. But such models require the existence of a symbolic equality predicate and hence again presuppose symbolic abilities in every case of success. Ideally, we would want a model that describes under what conditions \emph{same} is easy and under what conditions it is hard or unlearnable -- and how learning proceeds in hard cases. Here, we aim to lay the foundation for the development of such an account.

We are inspired by an emergent perspective in the animal learning literature that the representations underlying non-human animals' and human infants' successes in equality reasoning tasks are graded \cite{wasserman:2017}. This view acknowledges the increasing evidence that other species like pigeons \cite{cook2007}, crows \cite{smirnova2015}, and baboons \cite{fagot2011} can make true, out-of-sample generalizations of \emph{same} and \emph{different} relations, but it also recognizes that the observed patterns of behavior do not show the hallmarks of all-or-none symbolic representations. Instead, performance is graded. Out-of-sample generalization is possible but the level of performance depends critically on the diversity of the training stimuli \cite[e.g.,][]{castro2010}. Success requires hundreds, thousands, or even tens of thousands of training trials. And the outcome of learning is noisy and imperfect. These learning signatures appear to be a close match to the kind of learning exhibited by neural networks. Such networks are a flexible framework for arbitrary function learning which have enjoyed a huge resurgence of interest in recent years in the fields of artificial intelligence, neuroscience, and cognitive science \cite[e.g.,][]{lecun2015,saxe2019}.

In an influential rebuttal of the use of neural network models for capturing relational reasoning, \citet{marcus:1999} argued that a broad class of recurrent neural networks were unable to learn a general solution to sequential equality problems and \citet{marcus:2001} argued that feed-forward neural networks are even unable to learn a general solution \blue{even to mathematical identity, which is arguably the simplest version of \emph{same}/\emph{different} reasoning conceptually}. These claims were subsequently challenged by the presentation of evidence that some forms of neural networks are able to learn (at least aspects of) \citeauthor{marcus:1999}'s equality tasks  \cite{dienes:1999,seidenberg:1999a,seidenberg:1999b,elman:1999,negishi:1999}, yet these examples were controversial. The resulting debate \cite[reviewed in][]{alhama:2019} revealed a striking lack of consensus on some of the ground rules regarding what sort of generalization would be required to show that the learned function was suitably abstract.

In the time since these debates, successful neural network models have been developed for tasks such as natural language inference \cite{Bowman:2015, Williams:2018}, question answering \cite{Rajpurkar:2016, Rajpurkar:2018}, and visual reasoning \cite{Johnson:2017}. In many respects, all of these tasks are far more complex than equality-based tasks. In light of these findings, it may be surprising that the debate over equality-based reasoning is unresolved \cite{alhama:2019}. Yet even recent work on equality-based reasoning tasks takes as its starting point the conclusion that neural networks are unable to succeed using standard architectures and general purpose learning algorithms \cite{alhama2018,weyde2019,weyde2018,kopparti2020}. Further, though tasks and contexts vary, work in both computer vision \cite{Kim:2018,Fleuret:2011,Bengio:2016} and machine reasoning \cite{raposo2017,santoro2017,santoro2018,palm2018} has presupposed that relational reasoning generally -- and sometimes equality-based reasoning specifically -- is difficult or impossible in standard network architectures.

\begin{figure}[tp]
  \centering
  \includegraphics[scale=0.30]{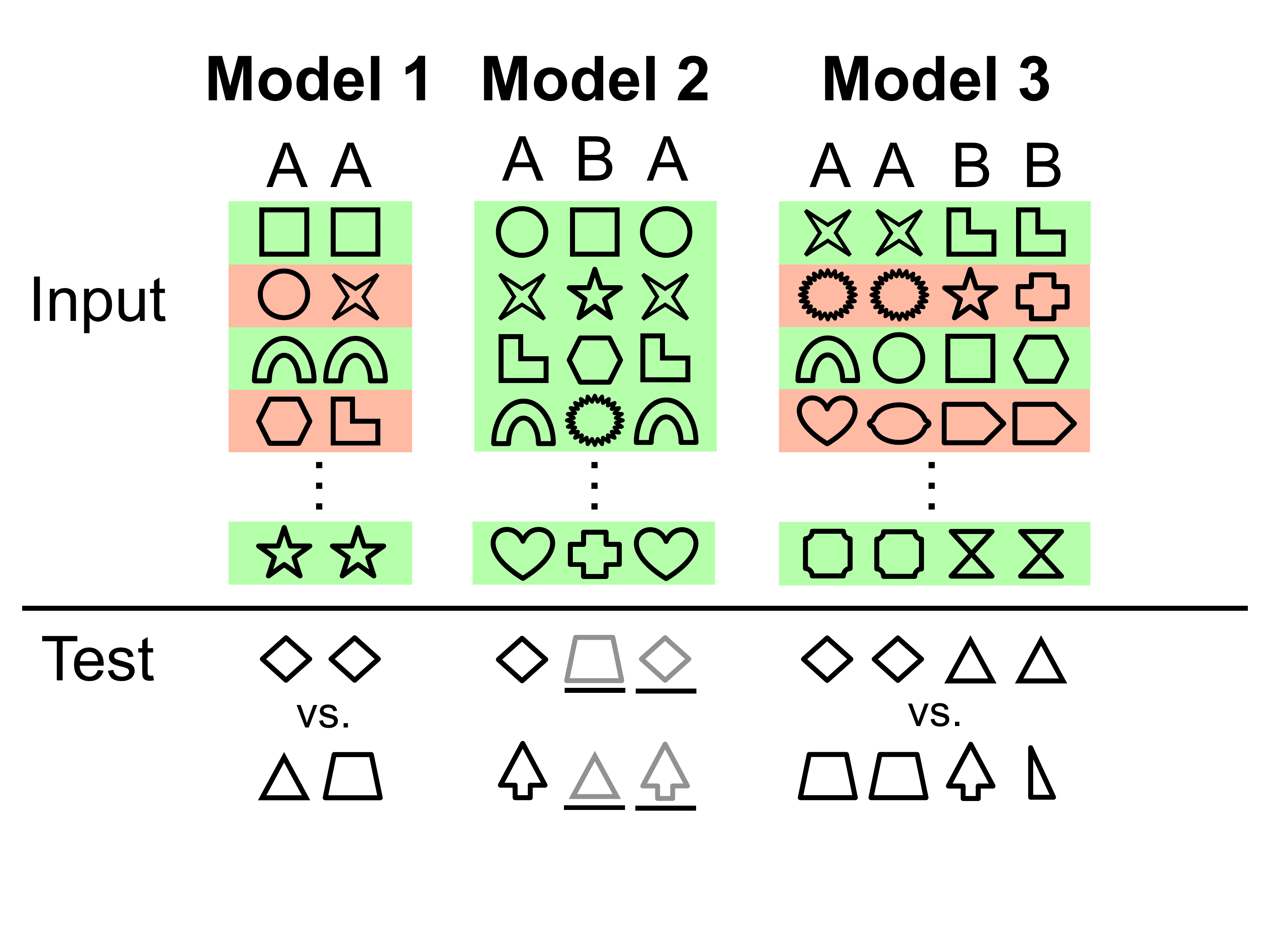}
  \caption{Relational reasoning tasks. Green and red mark positive and negative training examples, respectively. The sequential task (Model~2) uses only positive instances, and a model succeeds if, prompted with $\alpha$, it produces a sequence $\beta \ \alpha$ where $\beta \neq \alpha$. For the hierarchical task (Model~3), we show that a model trained on the basic task (Model~1) is effective with no additional training.}
  \label{fig:tasks}
\end{figure}

Modern deep learning models have been so successful that it seems odd that they would be completely unable to learn equality-based reasoning tasks. We suspect these claims remain in the literature in part because only a narrow range of network architectures and representations were explored in the earlier debate, and in part because the debate predated many important innovations in neural network design. Thus we revisit the debate here, using a broader range of architectures and representations and adopting stringent criteria for generalization. In particular we explore random and pretrained representations, which have facilitated many of the successes of modern artificial intelligence \cite{CollobertWeston:2011,Mikolov-etal:2013,pennington-socher-manning:2014:EMNLP2014,Peters-etal:2018,Devlin-etal:2019}. The use of pretrained representations to solve downstream tasks in particular is argued to be a hallmark of natural learning systems \cite{zador2019}, has been an important feature of historical models from cognitive science \cite[e.g.,][]{landauer1997,mcrae1993}, and is essential in the latest wave of state-of-the-art natural language processing models \cite{Devlin-etal:2019,Liu:2019,Radford:2019,Brown:2020}.

In our current work, we model three cases of equality-based reasoning that have featured prominently in discussions of the role of symbols in relational reasoning (\figref{fig:tasks}): (1) learning to discriminate pairs of objects that exemplify the relation \emph{same} or \emph{different}, (2) learning sequences with repeated \emph{same} elements \cite{marcus:1999}, and (3) learning to distinguish hierarchical \emph{same} and \emph{different} relations in a context with pairs of pairs exemplifying these relations \cite{Premack:1983}. Across these three models, we find strong support for their ability to learn equality relations. These results should serve to revise the conclusions of the earlier debate.

\citet{marcus:1999} and \citet{marcus:2001} showed experimentally that certain neural networks could not generalize the equality relation to stimuli unseen in training. We agree with this claim (and support it with a direct mathematical argument). But we show that the claim is specific to networks that use particular kinds of representations, rather than neural networks in general. The representations that are problematic for this task are \tech{featural} representations (a term that we define in more detail below): representations where either specific properties (e.g., shape, color) or entire items are represented as discrete dimensions in the network's inputs.

Marcus and colleagues concluded from their results that neural networks need to have primitive symbolic operators to solve equality-based relational reasoning tasks, which is a solution that has been pursued in recent machine learning research \cite{weyde2019,weyde2018, kopparti2020}. On this point, we disagree. Our experiments show that networks without such primitives can solve a range of these tasks using the sort of random or pretrained representations that are now the norm throughout artificial intelligence research. Overall, these findings suggest that some essential aspects of symbolic reasoning can emerge from entirely data-driven, non-symbolic learning processes.

Our work here makes three contributions. First, we resolve this longstanding debate by demonstrating neural networks are able to learn equality relations when provided with pretrained or random representations. Second, we modify the standard architecture of a recurrent neural network to allow it to learn the sequential equality task with no negative feedback. Negative evidence was dismissed as an unreasonably strong learning regime in the original debate over these issues \citep[e.g.,][]{marcus:1999a}, and we show that this learning regime is not necessary. Third, we show that a model pretrained on the simple equality task can achieve zero-shot generalization (generalization with no new training instances) to the hierarchical equality task, suggesting that pretraining might provide an account of how some organisms succeed on hard relational learning tasks. We believe these three contributions represent significant progress in our understanding of neural networks' ability to perform equality-based reasoning.  Taken together, these contributions lay the groundwork for further non-symbolic neural network models of relational reasoning and abstract thought more broadly.

\blue{In all our experiments, we define \emph{same} and \emph{different} in terms of mathematical identity. These are the terms of the original debate, and learning these unambiguous relations is likely a prerequisite for learning real-world equality tasks grounded in some perceptual domain, such as a vision network trained for facial recognition \citep{Deepface:2021}. Thus, following the protocols of \citet{marcus:1999} and \citet{marcus:2001} as closely as possible, we assign every object a unique representation that is reliably retrieved by the model. We demonstrate that, starting from such pre-individuation, general solutions to equality relations can be learned if the right representations are used.}

\section{General methods: Designing theoretical models of equality learning}

We begin by discussing two critical design considerations for our models: (1) the standards for generalization by which models should be evaluated and (2) the type of representations they should use. To summarize this discussion: we select generalization tasks with fully disjoint training and test vocabularies to provide the most stringent test of generalization. Next, we show analytically that featural representations, which represent either object identities or object properties as discrete representational dimensions, limit successful out-of-sample generalization. Thus, we adopt randomly initialized representations or random representations that have been pretrained for our subsequent models. The studies reported in this paper were not preregistered. The datasets used in this paper are publicly available online.\footnote{All models, datasets, and code in this paper are available at \url{https://github.com/atticusg/NeuralEqualityExperiments} \citep{ourgit}.}

\subsection{Generalization}

The standard approach to training and evaluating neural networks is to choose a dataset, divide it randomly into training and test sets, train the system on the training set, and then use its performance on the test set as a proxy for its capacity to generalize to new data. There is a sizable literature focused on ensuring that models learn from the training set in ways that allow them to generalize; for a rich overview, see \citealt{Goodfellow-et-al-2016}, Part~II.

The standard approach is fine for many purposes, but it raises concerns in a context in which we are trying to determine whether a network has truly acquired a global solution to a target function. In particular, where there is any kind of overlap between the training and test vocabularies,
we can't rule out that the network might be primarily taking advantage of idiosyncrasies in the underlying dataset to ``cheat''. This cheating would happen because the network would simply memorize aspects of the training set and learn a local approximation of the target function that happens to provide traction during assessment. In recent years, these concerns have motivated two complementary innovations in model assessment within AI: \tech{systematic generalization tasks} and \tech{adversarial testing}.

In systematic generalization tasks, one centers the model evaluation around a specific phenonemon that is not represented directly in the test set but that is determined by the training set together with an assumption about how knowledge should generalize. For example, building on ideas from \citet{Lake2018}, \citet{wu2021} define test sets around specific adjective--noun modification structures $\mathit{A\, N}$, ensuring that the training set includes only examples involving $\mathit{A\, N'}$ and $\mathit{A'\, N}$ for $\mathit{A} \neq \mathit{A}'$ and $\mathit{N} \neq \mathit{N}'$. Given a general assumption of systematicity   \citep{fodor1988}, we expect this training set to support learning to make accurate predictions for $\mathit{A\, N}$, and the design of the train--test split gives us confidence that such predictions are not based on mere memorization of $\mathit{A\, N}$ itself. For other examples involving systematic generalization tests, see \citealt{geiger2020,goodwin-etal-2020-probing,Hupkes2020,yanaka-etal-2020-neural}.

In adversarial testing (sometimes called \tech{challenge testing}), one uses behavioral assessments to identify examples and phenomena that fool top-performing models but are natural for humans. This can follow a similar pattern to systematic generalization tasks, but here the models are generally ones that have been trained on large benchmark datasets intended to circumscribe a specific capability. The intent of the adversarial test is to show that models trained on these resources have persistent gaps and other weaknesses. The intuitive idea behind adversarial testing traces to \citet{Winograd:1972} (see also \citealt{Levesque:2013}), and it has been used effectively in a number of domains, including object recognition in digital images \citep{Goodfellow-etal:2015}, basic factual question answering \citep{D17-1215}, and natural language inference \citep{glockner-etal-2018-breaking,nie2019analyzing}. More recently, adversarial testing has been extended to include training on adversarial examples in order to develop more robust models \citep{nie-etal-2020-adversarial,bartolo-etal-2020-beat,potts-etal-2020-dynasent,kiela-etal-2021-dynabench}.

In our work, we adopt elements of both these modes for assessing generalization. Following \citet{marcus:1999}, we propose that networks must be evaluated on assessment sets that are completely disjoint in every respect from the training set, all the way down to the entities involved. For example, below, we train on pairs $(a, a)$ and $(a, b)$, where $a$ and $b$ are representations from a train vocabulary $V_{T}$. At test time, we create a new assessment vocabulary $V_{A}$, derive equality and inequality pairs $(\alpha, \alpha)$ and $(\alpha, \beta)$ from that vocabulary, and assess the trained network on these new examples. This is an adversarial setting in service of a specific, systematic learning target (equality). In adopting these methods, we get a clear picture of the system's capacity to generalize, and we can safely say that its performance during assessment is a window into whether a global solution to identity has been learned. This is a very challenging setting for any machine learning model. For the sequential same--different task, we will see that it even requires us to depart from the usual formulations for predictive language models, in that such models standardly cannot even be evaluated on examples that are completely out of vocabulary.

\subsection{Representations}\label{sec:representations}

\newcommand{\mysquare}{
\begin{tikzpicture}[scale=0.5]
 \node[rectangle, fill=red!100, minimum height=3mm, minimum width=3mm]{};
\end{tikzpicture}}

\newcommand{\bluetriangle}{

\begin{tikzpicture}[scale=0.25]
 \node[regular polygon,regular polygon sides=3, fill=blue!100, inner sep=2pt]{};
\end{tikzpicture}}
\newcommand{\redtriangle}{
\begin{tikzpicture}[scale=0.5]
 \node[regular polygon,regular polygon sides=3, fill=red!100, inner sep=2pt]{};
\end{tikzpicture}}
\newcommand{\bluesquare}{
\begin{tikzpicture}[scale=0.5]
 \node[regular polygon,regular polygon sides=4, fill=blue!100, minimum height=3mm, minimum width=3mm]{};
\end{tikzpicture}}
\newcommand{\redsquare}{
\begin{tikzpicture}[scale=0.5]
 \node[regular polygon,regular polygon sides=4, fill=red!100, minimum height=3mm, minimum width=3mm]{};
\end{tikzpicture}}
\newcommand{\bluepentagon}{
\begin{tikzpicture}[scale=0.5]
 \node[regular polygon,regular polygon sides=5, fill=blue!100, minimum height=3mm, minimum width=3mm]{};
\end{tikzpicture}}
\newcommand{\redpentagon}{
\begin{tikzpicture}[scale=0.5]
 \node[regular polygon,regular polygon sides=5, fill=red!100, minimum height=3mm, minimum width=3mm]{};
\end{tikzpicture}}

\newcommand{\myrectangle}{
\begin{tikzpicture}[scale=0.25]
 \node[rectangle, fill=blue!100, minimum height=3mm, minimum width=5mm]{};
\end{tikzpicture}}

\newcommand{\mypentagon}{
\begin{tikzpicture}[scale=0.5]
 \node[regular polygon,regular polygon sides=5, minimum height=4mm, fill=red!100]{};
\end{tikzpicture}}

\newcommand{\mypolygon}{
\begin{tikzpicture}[scale=0.5]
 \node[regular polygon,regular polygon sides=7, minimum height=4mm, fill=blue!100]{};
\end{tikzpicture}}

\newcommand{\minshape}{6mm}

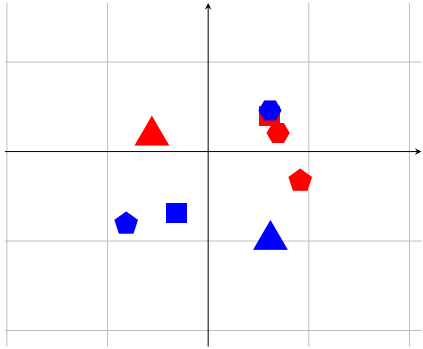
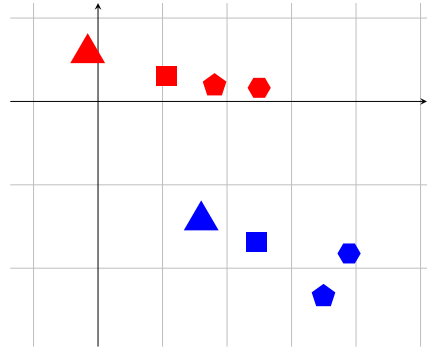
\begin{figure}[t!]
  \setlength{\arraycolsep}{2pt}
  \centering
  \small
  \begin{subfigure}[t]{0.05\textwidth}
    \centering
    \renewcommand{\arraystretch}{1.18}
    \begin{tabular}[b]{@{} c @{}}
      \midrule
      \redtriangle \\
      \bluetriangle \\
      \redsquare \\
      \bluesquare \\
      $\vdots$ \\
      \bluepentagon \\
      \bottomrule
    \end{tabular}
    \caption{}
    \label{fig:entities}
  \end{subfigure}
  \begin{subfigure}[t]{0.22\textwidth}
    \centering
    \renewcommand{\arraystretch}{1.18}
    $\begin{array}[b]{*{6}{c}}
       \toprule
       \redtriangle & \bluetriangle & \redsquare & \bluesquare & \redpentagon & \bluepentagon \\
       \midrule
       1 & 0 & 0 & 0& 0& 0 \\
       0 & 1 & 0 & 0& 0& 0 \\
       0 & 0 & 1 & 0& 0& 0 \\
       0 & 0 & 0 & 1& 0& 0 \\
       \multicolumn{6}{c}{\vdots} \\
       0 & 0 & 0 & 0& 0& 1 \\
       \bottomrule
     \end{array}%
    $
    \caption{Localist.}
    \label{fig:reps:localist}
  \end{subfigure}
  \begin{subfigure}[t]{0.2\textwidth}
    \centering
    \renewcommand{\arraystretch}{1.18}
    $\begin{array}[b]{*{3}{r}}
       \toprule
       \rotatebox{90}{red} & \rotatebox{90}{blue} & \rotatebox{90}{sides}\\
       \midrule
       0 & 1 & 3  \\
       1 & 0 & 3  \\
       0 & 1 & 4  \\
       1 & 0 & 4 \\
       \multicolumn{3}{c}{\vdots} \\
       1 & 0 & 5 \\
       \bottomrule
     \end{array}%
   $
   \caption{Property.}
   \label{fig:reps:symbolic}
 \end{subfigure}
 \begin{subfigure}[t]{0.2\textwidth}
   \centering
   \renewcommand{\arraystretch}{1.18}
   $\begin{array}[b]{rr}
      \toprule
      d_{1} &  d_{2} \\
      \midrule
      -0.3 &    0.1 \\
      0.3 &   -0.5 \\
      0.3 &    0.2 \\
      -0.2 &   -0.3 \\
      0.5 &   -0.2 \\
      -0.4 &   -0.4 \\
      \bottomrule
    \end{array}%
   $
   \caption{Random.}
   \label{fig:reps:random}
 \end{subfigure}
 \begin{subfigure}[t]{0.2\textwidth}
   \centering
   \renewcommand{\arraystretch}{1.18}
   $\begin{array}[b]{rr}
      \toprule
      d_{1} &  d_{2} \\
      \midrule
      -0.1 &    0.6 \\
      0.8 &   -1.4 \\
      0.5 &    0.3 \\
      1.2 &   -1.7 \\
      0.9 &    0.2 \\
      1.7 &   -2.3 \\
      \bottomrule
    \end{array}
    $
   \caption{Pretrained.}
   \label{fig:reps:pretrained}
 \end{subfigure}

 \vspace{12pt}

 \begin{subfigure}[t]{0.49\linewidth}
   \centering
   \begin{tikzpicture}[scale=0.8]
     \begin{axis}[
       title={},
       grid=major,
       axis lines=middle,
       no markers,
       every axis plot/.append style={ultra thick},
       xmin=-1.01, xmax=1.06,
       ymin=-1.09, ymax=0.83,
       ticks=none
       ]
       \node[regular polygon, regular polygon sides=3, fill=red!100] at (axis cs:-0.27955937, 0.08926568) {};
       \node[regular polygon, regular polygon sides=3, fill=blue!100] at (axis cs:0.30943045, -0.49350125) {};
       \node[regular polygon, regular polygon sides=4, fill=red!100] at (axis cs:0.30581924, 0.1981394) {};
       \node[regular polygon, regular polygon sides=4, fill=blue!100] at (axis cs:-0.15974948, -0.3445205) {};
       \node[regular polygon, regular polygon sides=5, fill=red!100] at (axis cs:0.45721307, -0.16340545) {};
       \node[regular polygon, regular polygon sides=5, fill=blue!100] at (axis cs:-0.40725416, -0.40328363) {};
       \node[regular polygon, regular polygon sides=6, fill=red!100] at (axis cs:0.34749436, 0.10372603) {};
       \node[regular polygon, regular polygon sides=6, fill=blue!100] at (axis cs:0.30712828, 0.22973178) {};
     \end{axis}
   \end{tikzpicture}
   \caption{The matrix in \figref{fig:reps:random}.}
   \label{fig:tsne-random}
 \end{subfigure}
 \begin{subfigure}[t]{0.49\linewidth}
   \centering
   \begin{tikzpicture}[scale=0.8]
     \begin{axis}[
       title={},
       no markers,
       grid=major,
       axis lines=middle,
       every axis plot/.append style={ultra thick},
       xmin=-0.68, xmax=2.55,
       ymin=-2.94, ymax=1.18,
       ticks=none
       ]
       \node[regular polygon, regular polygon sides=3, fill=red!100] at (axis cs:-0.080068775, 0.5792134) {};
       \node[regular polygon, regular polygon sides=3, fill=blue!100] at (axis cs:0.79974085, -1.4261162) {};
       \node[regular polygon, regular polygon sides=4, fill=red!100] at (axis cs:0.52764297, 0.30605778) {};
       \node[regular polygon, regular polygon sides=4, fill=blue!100] at (axis cs:1.2276263, -1.6893328) {};
       \node[regular polygon, regular polygon sides=5, fill=red!100] at (axis cs:0.9033405, 0.1921083) {};
       \node[regular polygon, regular polygon sides=5, fill=blue!100] at (axis cs:1.7477895, -2.3355992) {};
       \node[regular polygon, regular polygon sides=6, fill=red!100] at (axis cs:1.2491779, 0.16325347) {};
       \node[regular polygon, regular polygon sides=6, fill=blue!100] at (axis cs:1.9460244, -1.8231807) {};
     \end{axis}
   \end{tikzpicture}
   \caption{The matrix in \figref{fig:reps:pretrained}.}
   \label{fig:tsne-pretrained}
 \end{subfigure}
 \caption{Each matrix is a method for representing the shapes in \figref{fig:entities}, where each row is a vector representation of one shape. Localist and property are featural representations where each vector dimension encodes the value of a single, semantically interpretable property. Random and pretrained are non-featural representations where the values of properties are encoded implicitly in two dimensions. Random representations and localist representations encode only identity, whereas property representations and pretrained representations encode color and number of sides.}
 \label{fig:reps}
\end{figure}

Essentially all modern machine learning models represent objects using vectors of real numbers. However, there are important differences in how these vectors are used to encode the properties of objects. The method of representation chosen for a particular model impacts both whether there is a natural notion of similarity between entities and whether the model can generalize to examples unseen in training. These two attributes are deeply related; if there is a natural notion of similarity between vector representations, then models can generalize to inputs with representations that are similar to those seen in training.

We characterize two broad approaches to such property encoding -- which we call \tech{featural representations} and \tech{non-featural representations} -- and argue that the differences between them have not been given sufficient attention in the debate about the ability of neural networks to perform relational reasoning. We acknowledge that a dimension of any vector representation is a ``feature'', but we adopt a usage that is common in cognitive science, namely, that a feature is an interpretable semantic primitive.\footnote{The term \tech{distributed representations} is used to refer jointly to what we call property representations, random representations, and pretrained representations. We opted not to use this term because it does not seperate property representations from random and pretrained representations, which is the relevant division here. Distributed representations are often contrasted in the neural network literature with \tech{local} or \tech{localist} representations; as discussed below, here we define these terms specifically to refer to representations whose features correspond to specific entities.}

We ground our discussion in a hypothetical universe of blocks which vary by shape and color. \Figref{fig:entities} is a partial view of them, and \figref{fig:reps:localist}--\figref{fig:reps:pretrained} present four different ways of encoding the properties of these objects in vectors.

\subsubsection{Featural representations}

The defining characteristic of \tech{featural} vector representations is that each dimension encodes the value of a single, semantically intepretable property. The properties can be binary, integer-valued, or real-valued.

We use the term \tech{localist} for the special case of featural representations in which only objects are represented and there is a feature corresponding to each object. In \figref{fig:reps:localist}, each column represents the property of being an object, and every object is represented as a vector that has a single unit with value 1 (which is why these representations are often called ``one-hot'' vectors). There is no shared structure across objects; all are equally (un)related to each other as far as the model is concerned.

We will refer to featural representations that are not localist as \tech{property representations}. Here, column dimensions encode specific, meaningful properties of objects. In our example, we can represent the properties of being red and being blue with two different binary features, and the property of having a certain number of sides as a single integer feature, as in \figref{fig:reps:symbolic}. Unlike with localist representations, objects in this space can have complex relationships to each other, as encoded in the shared structure given by the columns.

\subsubsection{Non-featural representations}\label{sec:nonfeatural}

A \tech{non-featural representation} is a vector that encodes property values implicitly across many dimensions. Perhaps the simplest non-featural representations are \tech{completely random} vectors, as in \figref{fig:reps:random}. Random representations can be seen as the non-featural counterpart to localist representations. In both of these representation schemes, all the objects are equally (un)related to each other, since column-wise patterns are unlikely in random representations and, to the extent that they are present, they exist completely by chance. However, in random representations, all the column dimensions can contribute meaningfully to identifying objects, whereas a localist representation has only one vector unit that determines the identity of any given object.

\subsubsection{Pretraining}\label{sec:pretraining}

\begin{figure}[tp]
  \centering
  \resizebox{325pt}{!}{%
    \begin{tikzpicture}[
      font=\sf \large,
      >=LaTeX,
      rep/.style={
        rectangle,
        rounded corners=3mm,
        draw,
        very thick,
        minimum height =1cm,
        minimum width=1.61cm
      },
      function/.style={
        ellipse,
        draw,
        inner sep=1pt
      },
      gt/.style={
        rectangle,
        draw,
        minimum width=5mm,
        minimum height=4mm,
        inner sep=1pt
      },
      function/.style={
        rectangle,
        draw,
        minimum width=5mm,
        minimum height=4mm,
        inner sep=1pt
      },
      arrowconcat/.style={
        rounded corners=.25cm,
        dashed,
        thick,
        ->,
      },
      arrowfunction/.style={
        rounded corners=.25cm,
        thick,
        ->,
      }
      ]

      \node [rep, fill=\inputcolor] (input) at (6,1){$E[i]$} ;

      \node [rep, fill=\hiddencolor] (hidden) at (6,5){$h_{i}$} ;
      \node [gt, minimum width=1cm] (relu) at (6, 3-0.1) {$\ReLU( E[i]W_{xh} + b_{h})$};
      \draw [arrowfunction] (input) -- (relu) -- (hidden);

      \node [gt, minimum width=1cm] (softmax1) at (0,7-0.1) {$\softmax(h_{i}W_{hy}^{1} + b_{y}^{1})$};
      \node [gt, minimum width=1cm] (softmax2) at (6,7-0.1) {$\softmax(h_{i}W_{hy}^{2} + b_{y}^{2})$};
      \node [gt, minimum width=1cm] (softmax3) at (12,7-0.1) {$\softmax(h_{i}W_{hy}^{3} + b_{y}^{3})$};

      \node [rep, fill=\outputcolor] (output1) at (0,9){$y_{i,1}$} ;
      \node [rep, fill=\outputcolor] (output2) at (6,9){$y_{i,2}$} ;
      \node [rep, fill=\outputcolor] (output3) at (12,9){$y_{i,3}$} ;

      \draw [arrowfunction] (hidden) -- (softmax1) -- (output1);
      \draw [arrowfunction] (hidden) -- (softmax2) -- (output2);
      \draw [arrowfunction] (hidden) -- (softmax3) -- (output3);

    \end{tikzpicture}
  }
  \caption{Pretraining architecture shown for three pretraining tasks. The network is trained as a classifier, but the target of learning is the matrix of input representations $E[i]$, which are imbued with structure from the tasks via backpropagation.}
  \label{fig:models:pretraining}
\end{figure}
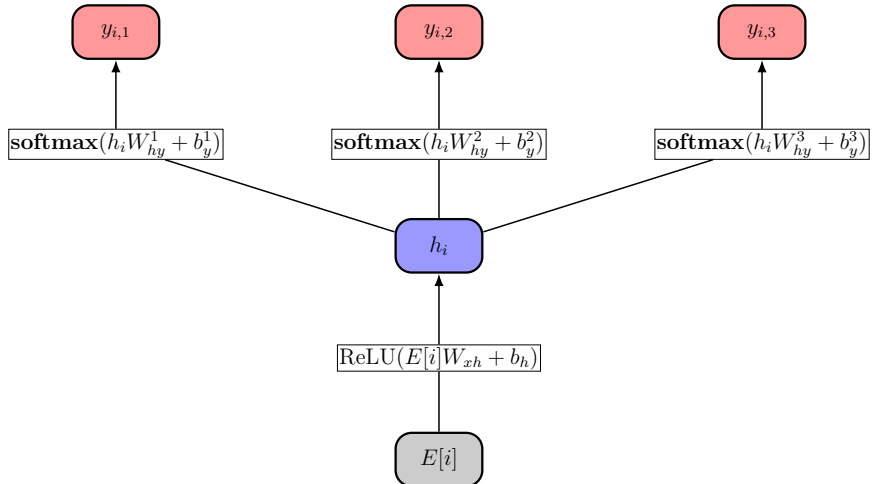

Random representations are a starting point that encodes object identity, but we can \tech{pretrain} these representations via a learning process, imbuing them with rich structure that implicitly encodes property values across many dimensions. \Figref{fig:reps:pretrained} provides a simple example. This matrix is the results of pretraining the representations in \figref{fig:reps:random} on three tasks: whether the object is blue, whether the object is red, and the number of sides the object has. Superficially, the two matrices look equally random, but the random representations in \figref{fig:tsne-random} have no such structure, while the pretrained representations in \figref{fig:tsne-pretrained} do. For example, there is a line that separates blue and red objects.

In our main experiments, we explore pretraining as a path to more efficient learning. There are many potential approaches to pretraining that could be effective in this context. We adopt a simple one based in a multi-task classifier. Intuitively, this works as follows: random representations (vectors) are paired with  labels corresponding to properties like those in our illustration: whether the object is blue, whether the object is red, and so forth. The multi-task classifier is trained as usual to predict these labels. However, rather than using the trained classifier to predict such labels, we focus instead on the way the input random representations are affected (pretrained) by this learning process. Once the process is complete, the richly structured inputs that result are then used in other tasks.

In detail, our pretraining model is a feed-forward network with a multi-task objective. For an example $i$ and task $j$, the model is defined as follows:
\begin{align}
  h_{i} &= \ReLU( E[i]W_{xh} + b_{h}) \\
  y_{i,j} &= \softmax(h_{i}W_{hy}^{j} + b_{y}^{j})
\end{align}
Here, $E[i]$ is the vector representation for example $i$ in an embedding matrix $E$ of dimension $|V| \times m$, where $V$ is the ``vocabulary'' of vectors to be pretrained. These representations are multiplied by a vector of weights $W_{xh}$ of dimension $m \times n$ and a bias vector $b_{h}$ of dimensions $n$ is added to the result. These two steps create a linear projection of the input representation, and the bias term is the value of this linear projection when the input representation is the zero vector. Then, the non-linear activation function $\ReLU(x) = \max(0, x)$ is applied element-wise to this linear projection. This non-linearity is what gives the neural model more expressive power than a logistic regression \citep{Cybenko:1989,Hornik:Stinchcombe:White:1989}. The result is the hidden representation $h_{i}$.

The hidden representation is the input to a separate classification layer for each class $j$. Each of these classification layers is defined by weights $W_{hy}^{j}$ and bias $b_{y}^{j}$. Each of these again defines a linear projection, now of $h_{i}$. The result is fed through the softmax activation function: $\softmax(x)_{i} = \frac{\exp{x_{i}}}{\sum_{j} \exp{x_{j}}}$. This creates a probability distribution over the classes (in our tasks, `positive' and `negative'). For a given input and task $j$, the model computes this probability distribution and the input is categorized as the class for $j$ with the higher probability.

\Figref{fig:models:pretraining} depicts this model in detail for 3 tasks ($J = 3$). The diagram helps reveal that the embedding $E$ and the parameters $W_{xh}$ and $b_{h}$ are jointly trained against all the tasks, whereas the parameters $h_{i}W_{hy}^{j}$ and $b_{y}^{j}$ are specific to each task $j$.

The overall objective of the model is to maximize the sum of the task objective functions. For $|V|$ examples, $J$ tasks, and $K_{j}$ the number of classes for task $j$:
\begin{equation}
  \max(\theta)
  \quad
  \frac{1}{|V|}
  \sum_{i=1}^{|V|}
  \sum_{j=1}^{J}
  \sum_{k=1}^{K_{j}}
  y^{i,j,k} \log \left( h_{\theta}(i)^{j,k} \right)
\end{equation}
where $y^{i,j,k}$ is the correct label for example $i$ in task $j$ and $h_{\theta}(i)^{j,k}$ is the predicted value for example $i$ in task $j$. We use $\theta$ to abbreviate all the model parameters: $E$, $W_{xh}$, $b_{h}$, and $W_{hy}^{j}$ and $b_{y}^{j}$ for each task $j$.

Our motivation for pretraining is to update the embedding $E$ so that its representations encode properties that can be used by subsequent models. To achieve this, we backpropagate errors through the network and into $E$. This backpropagation is the step that pretrains $E$ itself.

For the experiments reported below, we initialize $E$ randomly and then, for pretraining on $J$ tasks, we create a random binary vector of length $J$ for each row in $E$. Each dimension (task) in $J$ is independent of the others. We always pretrain for 10 epochs, where each epoch consists of an example for each item in the vocabulary. This choice is motivated primarily by computational costs; additional pretraining epochs greatly increase experiment run-times, though they do have the potential to imbue the representations with even more useful structure. Pretraining with the optimal hyperparameter settings always led to perfect accuracy on the pretraining tasks.

Pretraining need not be restricted to input representations; all the parameters of a model can be pretrained, offering the possibility that networks might be used as modular components to solve more complex tasks. We realize this possibility with our Experiment~4, where a model pretrained on a simple equality is used as a modular component to compute hierarchical equality.

\subsubsection{Localist and binary property representations prevent generalization}

Featural representations -- both localist and property -- have the appealing property that they are easy for researchers to interpret because of the tight correspondence between column dimensions and properties. However, this transparency actually inhibits neural networks from discovering general solutions. The core insight is that networks cannot learn anything about column dimensions that are not represented in their training data; whatever weights are associated with those dimensions are unchanged by the learning process, so predictions about those dimensions remain random at test time.

In order to see this, we need attend to the details of how neural models are trained. All the neural models in this paper are trained using the back-propagation learning algorithm. An easily observed fact about this algorithm is that, if a unit of the input vector is always zero during training, then any weights connected to that unit and only that unit will not change from their initialized values during training. This means that, when a standard neural model is evaluated on an input vector that has a non-zero value for a unit that was zero throughout training, untrained weights are used to make predictions, and so the network's behavior is unpredictable.

Of all the representation schemes we consider, localist representations are the one that most severely limit generalization. No two representations share a non-zero unit, and so when models are presented with inputs unseen in training, untrained weights are used and the resulting behavior is unpredictable. In addition, all such representations are orthogonal and equidistant from one another, so there is no notion of similarity, and consequently there is no usable transfer of information from training to assessment.

Property representations with binary features also limit generalization, though less severely than localist representations. Localist representations prevent generalization to \emph{entities} unseen during training, while binary feature representations prevent generalization to \emph{properties} unseen during training. For example, if color and shape are represented as binary features, and a red square and blue circle are seen in training, then a model could generalize to the unseen entities of a blue circle or a red square. However, if no entity that is a circle is seen during training, then the binary feature representing the property of being a circle is zero throughout training and untrained weights are used when the model is presented with a entity that is a circle during testing, which once again results in unpredictable behavior.

Property representations with analog features do not inhibit generalization in the same way. If height is represented as an analog feature, then a single unit represents all height values and is always non-zero. Of course, if the height feature is held out entirely during training (i.e., seen only at test time), then generalization will be limited just as it is for binary property representations.

Importantly, non-featural representations do not inhibit generalization in these ways, because all units for all representations are non-zero and the network can learn parameters that create complex associations between these entities and its task labels. These representations then can be generalized to new entities, even those unseen during training. \blue{In machine learning, learned non-featural representations are now the norm, not only because they address this limitation but also because they have proven superior even in settings where all column dimensions are well represented in the training data.}

Recent work in machine learning \cite{weyde2019,weyde2018,kopparti2020} attempts to overcome the analytic limitations of binary featural representations by modifying standard neural architectures to have symbolic primitives or changing network weight priors. In our work, we instead opt for non-featural representations, which do not have this analytic limitation and are the norm in state-of-the-art artificial intelligence models. There is no need to introduce symbolic primitives or modify network weight priors when non-featural representations are used.

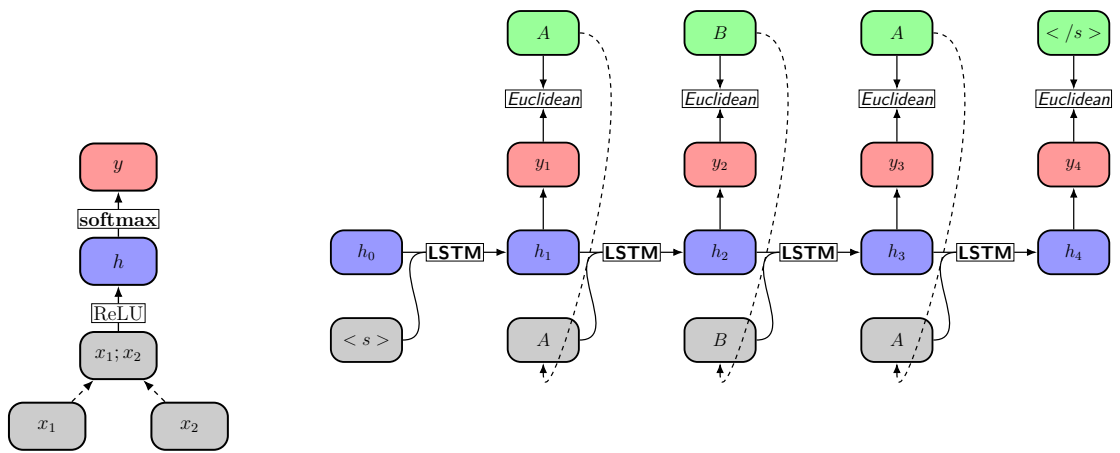
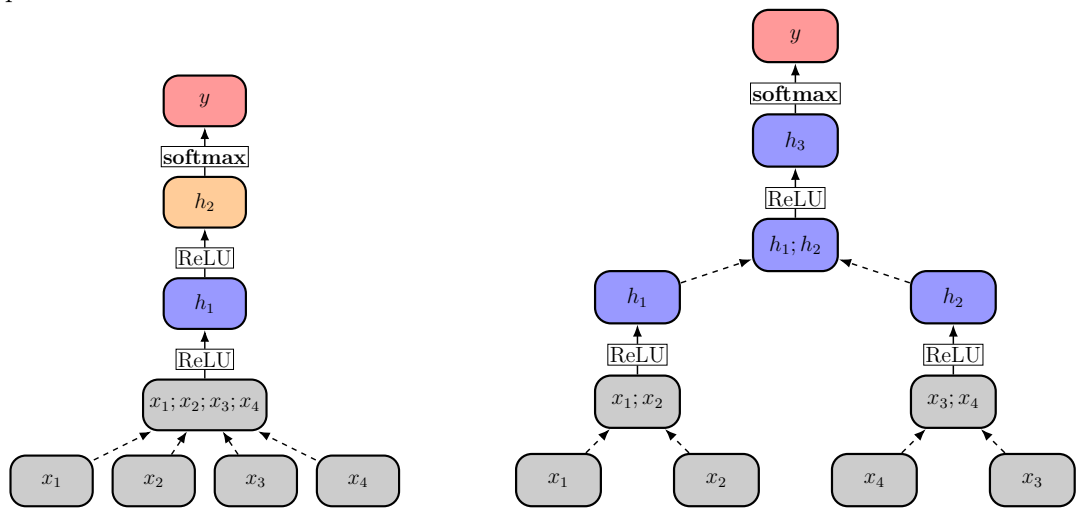
\begin{figure}

\begin{subfigure}[t]{0.28\textwidth}
  \centering
  \resizebox{0.7\textwidth}{!}{%
    \begin{tikzpicture}[
      font=\sf \large,
      >=LaTeX,
      rep/.style={
        rectangle,
        rounded corners=3mm,
        draw,
        very thick,
        minimum height =1cm,
        minimum width=1.61cm
      },
      function/.style={
        ellipse,
        draw,
        inner sep=1pt
      },
      gt/.style={
        rectangle,
        draw,
        minimum width=5mm,
        minimum height=4mm,
        inner sep=1pt
      },
      function/.style={
        rectangle,
        draw,
        minimum width=5mm,
        minimum height=4mm,
        inner sep=1pt
      },
      arrowconcat/.style={
        rounded corners=.25cm,
        dashed,
        thick,
        ->,
      },
      arrowfunction/.style={
        rounded corners=.25cm,
        thick,
        ->,
      }
      ]

      \node [rep, fill=\inputcolor] (input1) at (-1.5,1.5){$x_1$} ;
      \node [rep, fill=\inputcolor] (input2) at (1.5,1.5){$x_2$} ;
      \node [rep, fill=\inputcolor] (concat) at (0,3){$x_1;x_2$} ;
      \draw [arrowconcat] (input1) -- (concat);
      \draw [arrowconcat] (input2) -- (concat);

      \node [rep, fill=\hiddencolor] (hidden) at (0,5){$h$} ;
      \node [gt, minimum width=1cm] (relu) at (0,4-0.1) {$\ReLU$};
      \draw [arrowfunction] (concat) -- (relu) -- (hidden);

      \node [rep,fill=\outputcolor] (output) at (0,7){$y$} ;
      \node [gt, minimum width=1cm] (softmax) at (0,6-0.1) {$\softmax$};
      \draw [arrowfunction] (hidden) -- (softmax) -- (output);

    \end{tikzpicture}
  }
  \caption{The single layer network computing equality from the first experiment.}
  \label{fig:models:equality}
\end{subfigure}
\hfill
\begin{subfigure}[t]{0.68\textwidth}
  \resizebox{\textwidth}{!}{%
    \begin{tikzpicture}[
      font=\sf \large,
      >=LaTeX,
      rep/.style={
        rectangle,
        rounded corners=3mm,
        draw,
        very thick,
        minimum height =1cm,
        minimum width=1.61cm
      },
      function/.style={
        ellipse,
        draw,
        inner sep=1pt
      },
      gt/.style={
        rectangle,
        draw,
        minimum width=5mm,
        minimum height=4mm,
        inner sep=1pt
      },
      function/.style={
        rectangle,
        draw,
        minimum width=5mm,
        minimum height=4mm,
        inner sep=1pt
      },
      arrowconcat/.style={
        rounded corners=.25cm,
        dashed,
        thick,
        ->,
      },
      arrowfunction/.style={
        rounded corners=.25cm,
        thick,
        ->,
      }
      ]

      \node [rep, fill=\inputcolor] (input1) at (-10,4){$<s>$} ;
      \node [rep, fill=\inputcolor] (input2) at (-6,4){$A$} ;
      \node [rep, fill=\inputcolor] (input3) at (-2,4){$B$} ;
      \node [rep, fill=\inputcolor] (input4) at (2,4){$A$} ;

      \node [rep, fill=\hiddencolor] (hidden0) at (-10,6){$h_0$} ;
      \node [rep, fill=\hiddencolor] (hidden1) at (-6,6){$h_1$} ;
      \node [rep, fill=\hiddencolor] (hidden2) at (-2,6){$h_2$} ;
      \node [rep, fill=\hiddencolor] (hidden3) at (2,6){$h_3$} ;
      \node [rep, fill=\hiddencolor] (hidden4) at (6,6){$h_4$} ;

      \node [gt] (LSTM0) at (-8,6){$\LSTM$} ;
      \node [gt] (LSTM1) at (-4,6){$\LSTM$} ;
      \node [gt] (LSTM2) at (0,6){$\LSTM$} ;
      \node [gt] (LSTM3) at (4,6){$\LSTM$} ;

      \node [rep, fill=\outputcolor] (output1) at (-6,8){$y_1$} ;
      \node [rep, fill=\outputcolor] (output2) at (-2,8){$y_2$} ;
      \node [rep, fill=\outputcolor] (output3) at (2,8){$y_3$} ;
      \node [rep, fill=\outputcolor] (output4) at (6,8){$y_4$} ;

      \node [gt] (compare1) at (-6,9.5){$\compare$} ;
      \node [gt] (compare2) at (-2,9.5){$\compare$} ;
      \node [gt] (compare3) at (2,9.5){$\compare$} ;
      \node [gt] (compare4) at (6,9.5){$\compare$} ;

      \node [rep, fill=\colorlabel] (label1) at (-6,11){$A$} ;
      \node [rep, fill=\colorlabel] (label2) at (-2,11){$B$} ;
      \node [rep, fill=\colorlabel] (label3) at (2,11){$A$} ;
      \node [rep, fill=\colorlabel] (label4) at (6,11){$</s>$} ;

      \draw [arrowfunction] (label1) -- (compare1);
      \draw [arrowfunction] (label2) -- (compare2);
      \draw [arrowfunction] (label3) -- (compare3);
      \draw [arrowfunction] (label4) -- (compare4);

      \draw [arrowfunction] (output1) -- (compare1);
      \draw [arrowfunction] (output2) -- (compare2);
      \draw [arrowfunction] (output3) -- (compare3);
      \draw [arrowfunction] (output4) -- (compare4);

      \draw [arrowfunction] (hidden1) -- (output1);
      \draw [arrowfunction] (hidden2) -- (output2);
      \draw [arrowfunction] (hidden3) -- (output3);
      \draw [arrowfunction] (hidden4) -- (output4);

      \draw [thick] (input1) to[out=0,in=-180, distance=1cm] (LSTM0);
      \draw [thick] (input2) to[out=0,in=-180, distance=1cm] (LSTM1);
      \draw [thick] (input3) to[out=0,in=-180, distance=1cm] (LSTM2);
      \draw [thick] (input4) to[out=0,in=-180, distance=1cm] (LSTM3);

      \draw [arrowfunction] (hidden0) -- (LSTM0) -- (hidden1);
      \draw [arrowfunction] (hidden1) -- (LSTM1)-- (hidden2);
      \draw [arrowfunction] (hidden2) -- (LSTM2)-- (hidden3);
      \draw [arrowfunction] (hidden3) -- (LSTM3)-- (hidden4);

      \draw[->,dashed,thick] (label1) to[out=0,in=-90, distance=1.99cm] (input2);
      \draw[->,dashed,thick] (label2) to[out=0,in=-90, distance=1.99cm] (input3);
      \draw[->,dashed,thick] (label3) to[out=0,in=-90, distance=1.99cm] (input4);
    \end{tikzpicture}
  }
  \caption{The recurrent LSTM network producing ABA sequences from the second experiment.}
  \label{fig:reps:sequence}
\end{subfigure}

\begin{subfigure}[t]{0.49\textwidth}
  \centering
  \resizebox{0.7\textwidth}{!}{%
    \begin{tikzpicture}[
      font=\sf \large,
      >=LaTeX,
      rep/.style={
        rectangle,
        rounded corners=3mm,
        draw,
        very thick,
        minimum height =1cm,
        minimum width=1.61cm
      },
      function/.style={
        ellipse,
        draw,
        inner sep=1pt
      },
      gt/.style={
        rectangle,
        draw,
        minimum width=5mm,
        minimum height=4mm,
        inner sep=1pt
      },
      function/.style={
        rectangle,
        draw,
        minimum width=5mm,
        minimum height=4mm,
        inner sep=1pt
      },
      arrowconcat/.style={
        rounded corners=.25cm,
        dashed,
        thick,
        ->,
      },
      arrowfunction/.style={
        rounded corners=.25cm,
        thick,
        ->,
      }
      ]

      \node [rep, fill=\inputcolor] (input1) at (-3,1.5){$x_1$} ;
      \node [rep, fill=\inputcolor] (input2) at (-1,1.5){$x_2$} ;
      \node [rep, fill=\inputcolor] (input3) at (1,1.5){$x_3$} ;
      \node [rep, fill=\inputcolor] (input4) at (3,1.5){$x_4$} ;
      \node [rep, fill=\inputcolor] (concat) at (0,3){$x_1;x_2;x_3;x_4$} ;
      \draw [arrowconcat] (input1) -- (concat);
      \draw [arrowconcat] (input2) -- (concat);
      \draw [arrowconcat] (input3) -- (concat);
      \draw [arrowconcat] (input4) -- (concat);

      \node [rep, fill=\hiddencolor] (hidden) at (0,5){$h_1$} ;
      \node [gt, minimum width=1cm] (relu) at (0,4-0.1) {$\ReLU$};
      \draw [arrowfunction] (concat) -- (relu) -- (hidden);

      \node [rep, fill=\hiddencolortwo] (hidden2) at (0,7){$h_2$} ;
      \node [gt, minimum width=1cm] (relu2) at (0,6-0.1) {$\ReLU$};
      \draw [arrowfunction] (hidden) -- (relu2) -- (hidden2);

      \node [rep,fill=\outputcolor] (output) at (0,9){$y$} ;
      \node [gt, minimum width=1cm] (softmax) at (0,8-0.1) {$\softmax$};
      \draw [arrowfunction] (hidden2) -- (softmax) -- (output);

    \end{tikzpicture}
  }
  \caption{The two layer network computing hierarchical equality from the third experiment.}
  \label{fig:models:premack-deep}
\end{subfigure}
\hfill
\begin{subfigure}[t]{0.49\textwidth}
  \centering
  \resizebox{\textwidth}{!}{%
    \begin{tikzpicture}[
      font=\sf \large,
      >=LaTeX,
      rep/.style={
        rectangle,
        rounded corners=3mm,
        draw,
        very thick,
        minimum height =1cm,
        minimum width=1.61cm
      },
      function/.style={
        ellipse,
        draw,
        inner sep=1pt
      },
      gt/.style={
        rectangle,
        draw,
        minimum width=5mm,
        minimum height=4mm,
        inner sep=1pt
      },
      function/.style={
        rectangle,
        draw,
        minimum width=5mm,
        minimum height=4mm,
        inner sep=1pt
      },
      arrowconcat/.style={
        rounded corners=.25cm,
        dashed,
        thick,
        ->,
      },
      arrowfunction/.style={
        rounded corners=.25cm,
        thick,
        ->,
      }
      ]

      \node [rep, fill=\inputcolor] (input1) at (-4.5,1.5){$x_1$} ;
      \node [rep, fill=\inputcolor] (input2) at (-1.5,1.5){$x_2$} ;
      \node [rep, fill=\inputcolor] (concat) at (-3,3){$x_1;x_2$} ;
      \draw [arrowconcat] (input1) -- (concat);
      \draw [arrowconcat] (input2) -- (concat);

      \node [rep, fill=\hiddencolor] (hidden) at (-3,5){$h_1$} ;
      \node [gt, minimum width=1cm] (relu1) at (-3,4-0.1) {$\ReLU$};
      \draw [arrowfunction] (concat) -- (relu1) -- (hidden);

      \node [rep, fill=\inputcolor] (input3) at (4.5,1.5){$x_3$} ;
      \node [rep, fill=\inputcolor] (input4) at (1.5,1.5){$x_4$} ;
      \node [rep, fill=\inputcolor] (concat2) at (3,3){$x_3;x_4$} ;
      \draw [arrowconcat] (input3) -- (concat2);
      \draw [arrowconcat] (input4) -- (concat2);

      \node [rep, fill=\hiddencolor] (hidden2) at (3,5){$h_2$} ;
      \node [gt, minimum width=1cm] (relu2) at (3,4-0.1) {$\ReLU$};
      \draw [arrowfunction] (concat2) -- (relu2) -- (hidden2);

      \node [rep,fill=\hiddencolor] (hiddenconcat) at (0,6){$h_1;h_2$};
      \draw [arrowconcat] (hidden2) -- (hiddenconcat);
      \draw [arrowconcat] (hidden) -- (hiddenconcat);

      \node [rep, fill=\hiddencolor] (hidden3) at (0,8){$h_3$} ;
      \node [gt, minimum width=1cm] (relu3) at (0,7-0.1) {$\ReLU$};
      \draw [arrowfunction] (hiddenconcat) -- (relu3) -- (hidden3);

      \node [rep,fill=\outputcolor] (output) at (0,10){$y$} ;
      \node [gt, minimum width=1cm] (softmax) at (0,9-0.1) {$\softmax$};
      \draw [arrowfunction] (hidden3) -- (softmax) -- (output);

    \end{tikzpicture}
  }
  \caption{The single layer network pretrained on equality computing hierarchical equality from the fourth experiment.}
  \label{fig:models:premack}
\end{subfigure}
  \caption{The neural models used in each of our four experiments.}
  \label{fig:models}
\end{figure}

\section{Experiment 1: A feed-forward neural model of same--different relations}\label{sec:equality}

We first investigate whether a basic supervised feed-forward neural network can learn the equality relation in the strict setting we describe above, where the training and test vocabularies are disjoint.

\subsection{Methods}

Our basic model for equality is a feed-forward neural network with a single hidden representation layer. The model is depicted graphically in \Figref{fig:models:equality} and defined as follows:
\begin{align}
h &= \ReLU([a;b]W_{xh} + b_{h}) \label{eq:x2h}\\
y &= \softmax(hW_{hy} + b_{y}) \label{eq:h2y}
\end{align}
The input is a pair of vectors $(a, b)$, which correspond to the two stimulus objects. These vectors are non-featural representations that do not have features encoding properties of the objects or their identity (though they may be pretrained). These are concatenated to form a single vector $[a;b]$ of dimension $2m$, which is the simplest way of merging the two representations to form a single input.

This representation is multiplied by a matrix of weights $W_{xh}$ of dimension $2m \times n$ and a bias vector $b_{h}$ of dimension $n$ is added to this result, where $n$ is the hidden layer dimensionality, with the $\ReLU$ activation applied element-wise to the result. This yields the hidden representation $h$, which is multiplied by a second matrix of weights $W_{hy}$, dimension $n \times 2$, and a bias term $b_{y}$ (dimension $2$) is added to this. A final softmax activation function creates a probability distribution over the classes. For a given input, the model computes this probability distribution and the input is categorized as the class with the higher probability.

During training, this model is presented with positive and negative labeled examples and the parameters $W_{xh}$, $W_{hy}$, $b_{y}$, and $b_{h}$ are learned using backpropagation with a cross entropy loss function. This function is defined as follows, for a corpus of $N$ examples and $K$ classes:
\begin{equation}
\max(\theta)
\quad
\frac{1}{N}
\sum_{i=1}^{N}
\sum_{k=1}^{K}
y^{i,k} \log(h_{\theta}(i)^{k})
\end{equation}
where $\theta$ abbreviates the model parameters ($W_{xh}$, $W_{hy}$, $b_{y}$, $b_{h}$), $y^{i,k}$ is the actual label for example $i$ and class $k$, and $h_{\theta}(i)^{k}$ is the corresponding prediction.

During testing, this model is tasked with categorizing inputs unseen during training. It is straightforward to show that a network like this is capable of learning equality as we have defined it. Indeed, in \SIref, we provide one possible analytic solution to the equality relation using this neural model. Here we illustrate with a small example network that maps all identity pairs to $[0.5, 0.5]$ and all non-identity pairs to $[y, 1-y]$ where $y > 0.5$, which supports a trivial classification rule:
\setlength{\arraycolsep}{4pt}
\begin{equation}
\softmax\left(
  \ReLU\left(
    [a;b]
    \left(
      \begin{array}[c]{@{} *{4}{r} @{}}
        1 & 0 & -1 & 0 \\
        0 & 1 & 0 & -1 \\
        -1 & 0 & 1 & 0 \\
        0 & -1 & 0 & 1
      \end{array}
    \right)
    +
    \mathbf{0}
  \right)
  \left(
    \begin{array}[c]{@{} r r @{}}
      1 & 0 \\
      1 & 0\\
      1 & 0\\
      1 & 0\\
    \end{array}
  \right)
  +
  \mathbf{0}
\right)
\end{equation}

This result shows that equality in our sense is learnable in principle, but it doesn't resolve the question of whether networks can find this kind of solution given finite training data. To address this issue, we train networks on a stream of pairs of random vectors. Half of these are identity pairs $(a, a)$, labeled with `positive', and half are non-identity pairs $(a, b)$, labeled with `negative'. Trained networks are assessed on the same kind of balanced dataset, with vectors that were never seen in training so that, as discussed earlier, we get a clear picture of whether they have found a generalizable solution.

We implemented this network using the multi-layer perceptron from scikit-learn \citep{scikit-learn}, and we conducted a wide hyperpameter search. See \SIref\ for a full specification of our experimental protocol. In the main text, we graph results for the single best hyperparameter setting, reflecting our goal of assessing the possibility of high performance in this task from a network of this type. All the models are trained on balanced datasets, and they never see the same input more than once. The test set is a fixed set of 500 vectors, disjoint from the training sets and also balanced across the two classes.

\subsection{Results}

\begin{figure}
 \centering
 \includegraphics[width=0.6\linewidth]{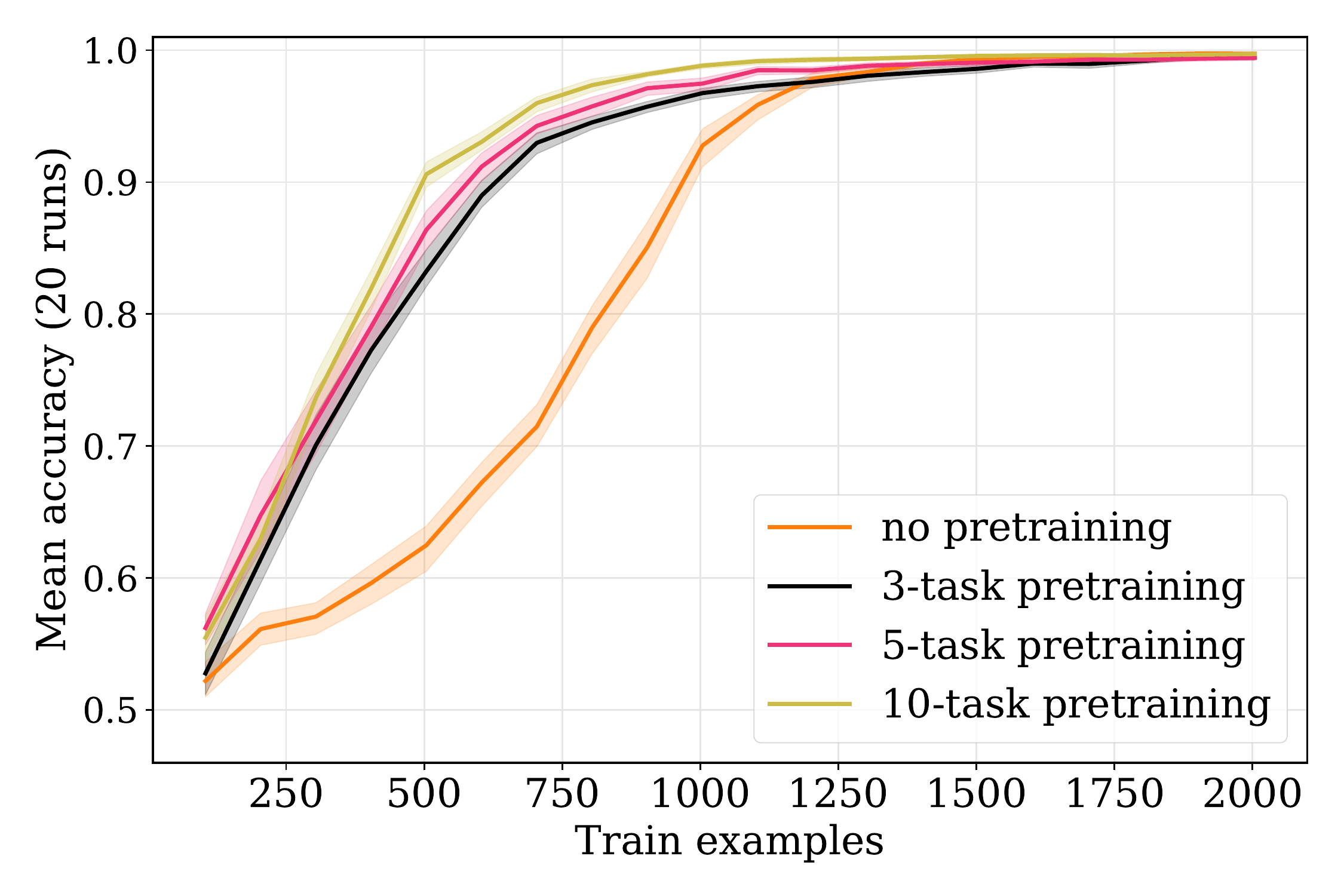}
 \caption{Experiment 1 results: feed-forward neural networks with one hidden layer trained on our simple equality task (10-dimensional inputs, 100-dimensional hidden layers). The `no pretraining' model is provided random representations and the `$k$-task pretraining' models are provided random representations that are grounded in $k$ binary property domains.}
 \label{fig:basic-equality-pretrain}
\end{figure}

\Figref{fig:basic-equality-pretrain} shows our results, for a model with an embedding dimension of 10 and a hidden-layer dimension of 100. (In the Appendix, \Figref{fig:model1} provides results for other network dimensionalities, and \figref{fig:train-results:e1} provides comparable results for evaluations on the training set.) The representations used in these experiments are random representations that were pretrained using a linear classifier for 0, 3, 5, or 10 different independent binary feature discrimination tasks (with 0 corresponding to purely random representations). For example, following \figref{fig:reps}, a three-task model might be trained to encode binary properties that we might gloss as `being blue', `having four sides', and `being red'. For all representations, this neural model reached above-chance performance almost immediately, but required upwards of 1,000 examples to achieve near perfect accuracy.   Interestingly, we observed a clear speed-up with pretraining, with more pretraining tasks resulting in the largest gains. It seems that, by grounding our representations in ``property domains'' (as represented by the different task dimensions), we imbue them with implicit structure that makes learning easier.

\subsection{Discussion}

Our assessment pairs have nothing in common with the training pairs except insofar as both involve vectors of real numbers of the same dimensionality. During training, the network is told (via labels) which pairs are equality pairs and which are not, but the pairs themselves contain no information about equality per se. It thus seems fair to us to say that these networks have learned equality -- or at least how to simulate that relation with near perfect accuracy. Further, the use of representations that are structured by pretraining results in faster learning.

\section{Experiment 2: A recurrent neural model of  sequential same--different}\label{sec:lms}

Our first model is simple and successfully learns equality. However, this model is supervised with both positive and negative evidence. In the initial debate around these issues, supervision with negative evidence was dismissed as an unreasonably strong learning regime \citep[e.g.,][]{marcus:1999a}. While this argument likely holds true for language learning \citep[in which supervision is generally agreed not to be binary or direct;][]{brown:1970,chouinard2003}, it is not necessarily true for learning more generally. Nevertheless, learning of sequential rules without negative feedback is possible for infants \cite{marcus:1999,rabagliati:2019}. In experiments of this type, infants are presented with a set of positive examples of the form ABA. Then their responses are measured for examples using entirely new stimuli that either conform to the ABA regularity or not. Our next model explores whether neural networks can learn this sequential equality task in the same challenging regime with no negative supervision.

\subsection{Methods}

To explore learning with only positive instances, we use a neural LSTM language model, a recurrent network with the ability to selectively forget and remember information \cite{hochreiter:1997}. Language models are sequential: at each timestep, they predict an output given their predictions about the preceding timesteps. As typically formulated, the prediction function is just a classifier: at each timestep, it predicts a probability distribution over the entire vocabulary of options, and the item with the highest probability is chosen as a symbolic output. This output becomes the input at the next timestep, and the process continues.

As we noted above, this standard formulation will not work in situations in which we want to make predictions about test items with an entirely disjoint vocabulary from the training sample. The classifier function will get no feedback about these out-of-vocabulary items during training, and so it will never predict them during testing.

To address this issue, we reformulate the prediction function. Our proposal is to have the model predict output vector representations -- instead of discrete vocabulary items -- at each timestep. During training, the model seeks to minimize the distance between these output predictions and the representations of the actual output entities. During assessment, we take the prediction to be the item in the entire vocabulary (training and assessment) whose representation is closest to the predicted vector (in terms of Euclidean distance). This fuzzy approach to prediction creates enough space for the model to predict sequences from an entirely new vocabulary.

The specific model we use for this is as follows:
\begin{align}
h_{t} &= \LSTM(x_{t}, h_{t-1}) \label{eq:lstm-recur}\\
y_{t} &= h_{t}W + b\label{eq:lstm-predict}
\end{align}
This holds for $t > 0$, and we set $h_{0} = \mathbf{0}$. $\LSTM$ is a long short-term memory cell \cite{hochreiter:1997}. We visualize this model in \figref{fig:reps:sequence}.

The input is a sequence of vectors $x_1, x_2, x_3, \dots$, each of dimension $m$, which correspond to a sequence of stimulus objects. These vectors are, again, non-featural representations that do not have features encoding properties of the objects or their identity, though they may be pretrained to encode such properties more abstractly.

At each timestep $t$, the input vector $x_t$ is fed into the $\LSTM$ cell along with the previous hidden representation $h_{t-1}$. The defining feature of an $\LSTM$ is the ability to decide whether to store information from the current input, $x_t$, and whether to remember or forget the information from the previous timestep $h_{t-t}$. The output of the $\LSTM$ cell is the hidden representation for the current time step $h_t$. The dimension of the hidden representations is $n$. The hidden representation is multiplied by a matrix $W$ with dimensionality $n \times m$ to produce $y_t$. This result, $y_t$, is a linear projection of the hidden representation into the input vector space, which is necessary because $y_t$ is a prediction of what the next input, $x_{t+1}$, will be.

The objective function is as follows:
\begin{equation}
\max(\theta)
\quad
-\frac{1}{N}
\sum_{i=1}^{N}
\sum_{t=1}^{T_{i}}
\left\| h_{\theta}\left(x^{i, 0:{t-1}}\right) - x^{i,t} \right\|^{2}
\end{equation}
for $N$ examples. Here, $T_{i}$ is the length of example $i$. As before, $\theta$ abbreviates the parameters of the model as specified in \dasheg{eq:lstm-recur}{eq:lstm-predict}. We use $h_{\theta}(x^{i, 0:{t-1}})$ for the vector predicted by the model for example~$i$ at timestep~$t$, which is compared to the actual vector at timestep $t$ via squared Euclidean distance (i.e., the mean squared error).

We implemented this recurrent LSTM network for the sequential ABA task using PyTorch \citep{pytorch}. As in our previous experiment, we conducted a wide hyperparameter search (see \SIref\ for details). In the main text, we graph results for the single best hyperparameter setting. We trained and tested on vocabularies of size 20, which are reconstructed between each run.

Our \SIrefnoour\
provide an analytic solution to the ABA task using this model. To see how well the model performs in practice, we trained networks on sequences \texttt{<s> a b a </s>}, where $\texttt{b} \neq \texttt{a}$. We show the network every such sequence during training, from an underlying vocabulary of 20 items (creating a total of 380 examples). To assess how well the model learns this pattern, we seed it with \texttt{<s> x} where \texttt{x} is an item from a disjoint vocabulary from that seen in training, and we say that a prediction is accurate if the model continues with \texttt{y x </s>}, where \texttt{y} is any character (from the training or assessment vocabulary) except \texttt{x}. In our experiments, we use a test vocabulary of 26 items, which creates 52 distinct \texttt{y x </s>} continuations and hence 52 distinct test examples.

\subsection{Results}

\begin{figure}
 \centering
 \includegraphics[width=0.6\linewidth]{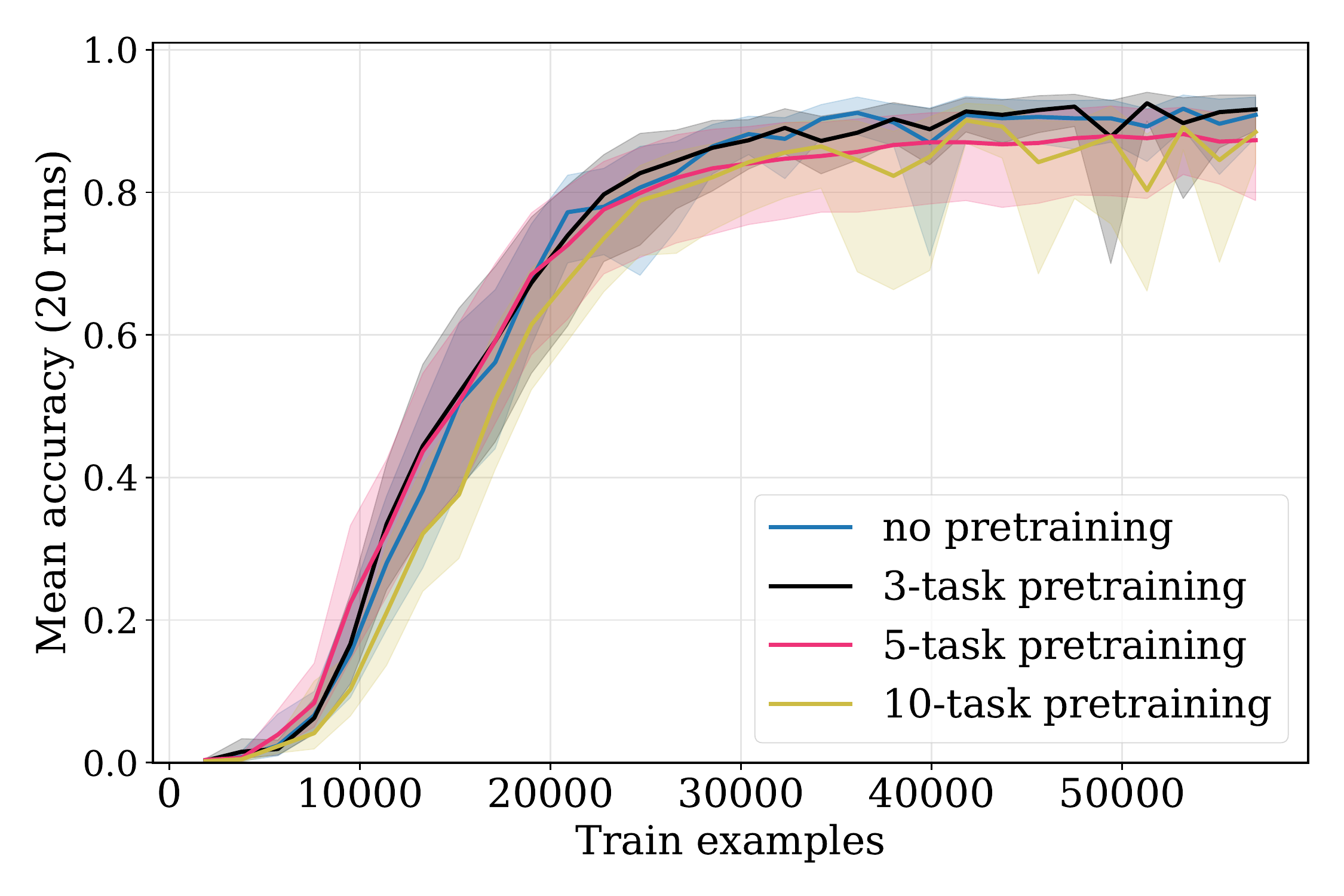}
 \caption{Experiment 2 results: LSTM recurrent neural networks trained on our sequential equality task (2-dimensional inputs, 100-dimensional hidden layers). The `no pretraining' model is provided random representations and the `$k$-task pretraining' models are provided random representations that are grounded in $k$ binary property domains via pretraining. All the training examples are presented at once over multiple epochs.}
 \label{fig:fuzzy-lm-pretrain-results}
\end{figure}

\Figref{fig:fuzzy-lm-pretrain-results} shows our results, for networks with 2-dimensional embeddings and 100-dimensional hidden representations. (\Figref{fig:model2} provides results for a wide range of network dimensionalities.) Unlike for the previous equality experiment, we found that we had to allow the model to experience multiple epochs of training on the same set in order to succeed, and tens of thousands of training examples were necessary. We considered a range of representations (as in Experiment~1), and the model was again successful with all representations. However, unlike in Experiment~1, we do not find evidence in favor of pretraining for this task.

\Figref{fig:train-results:e2} provides training-set results that are comparable to those in \figref{fig:fuzzy-lm-pretrain-results}. These models very quickly learn to perform with near perfect accuracy on the training set, creating a large gap between training-set performance and test-set performance. This contrasts with our other models, for which the differences between these two modes of evaluation are small. This difference likely traces to the nature of language modeling tasks (compared with classifiers) as well as our decision to allow the model to experience multiple epochs of the same training data.

\subsection{Discussion}

These sequential models are given no negative examples and they must predict into a totally new vocabulary. Despite these challenges, they succeed at learning the underlying patterns in our data. On the other hand, the learning process is slow and data-intensive. We hypothesized that grounding representations in property domains via pretraining might lead to noticeable speed-ups, as it did in for our simple same-different task (Experiment~1), but we did not see this effect in practice. We speculate that there may be model variants that reduce these demands, given that learning is in principle possible in this architecture, but we leave them to future work.

\section{Experiment 3: A feed-forward neural model of hierarchical same--different relations}\label{sec:premack}

Given the strong results found for simple equality relations, we can ask whether more challenging equality problems are also learnable in our setting. The hierarchical equality task used by \citet{Premack:1983} is an interesting test case: given a pair of pairs $((a,b), (c,d))$, the label is `positive' if $(a = b)$ and $(c = d)$ or $(a \neq b)$ and $(c \neq d)$. The mixed cases (i.e., $(a = b)$ and $(c \neq d)$; $(a \neq b)$ and $(c = d)$) are labeled `negative'.

\citeauthor{Premack:1983} suggested that the ability exemplified by this task -- reasoning about hierarchical \emph{same} and \emph{different} relations -- could represent a form of symbolic abstraction uniquely enabled by language. Given the non-symbolic nature of our models, our simulations provide a test of this hypothesis, though we should look critically at their ability to find good solutions with reasonable amounts of training data.

\subsection{Methods}

We can approach this task using the same model and methods as we used for equality, with the relatively minor change of providing the network four vector representations instead of two. We found that feed-forward neural networks with only one hidden layer required nearly 100,000 training examples to solve this task (see \figref{fig:model1:premack}).  We hypothesized that a single hidden layer network might be suboptimal here. This task is intuitively hierarchical: if one works out the equality labels for each of the two pairs, then the further classification decision can be done entirely on that basis. Our current neural network might be too shallow to find this kind of decomposition. To address this issue, we use a feed forward network with two hidden layers. We visualize this model in \Figref{fig:models:premack-deep} and we define
\begin{align}
h_{1} &= \ReLU([a;b;c;d]W_{xh} + b_{h_{1}}) \label{eq:x2h1}\\
h_{2} &= \ReLU(h_{1}W_{hh} + b_{h_{2}}) \label{eq:x2h2}\\
y &= \softmax(h_{2}W_{hy} + b_{y}) \label{eq:h2y2}
\end{align}

We implemented this network using PyTorch and conducted a wide hyperparameter seach (see \SIref\ for details). In the main text, we graph results for the single best hyperparameter setting. The training and testing protocol is the same as in Experiment~1, and we further ensure that the train and test sets are balanced across the four distinct input types for the hierarchical task (same/same, different/different, same/different, and different/same).

\subsection{Results}

\Figref{fig:premack-h2-pretrain} shows our results for networks with 10-dimensional embeddings and 100-dimensional hidden layers. (\Figref{fig:model3a} shows results for other network dimensionalities.) We again considered a range of representations, and again the network succeeded across this range, with pretraining increasing performance dramatically. The network required more than 20,000 training instances to reach top performance, and upwards of 10,000 examples with pretrained representations.

\subsection{Discussion}

Our model with two hidden layers requires vastly more data than human participants get in similar experiments. For example, sequential rule learning experiments typically involve short exposures in the range of dozens to hundreds of examples \cite[e.g.,][]{marcus:1999,endress2005}. Thus, it is worth asking whether there are other solutions that would be more data efficient and more in line with human capabilities. We next seek to further capitalize on the hierarchical nature of this task by defining a modular pretraining regime in which previously learned capabilities are recruited for new tasks.

\section{Experiment 4: A tree-structured model of hierarchical same--different relations pretrained on simple equality}\label{sec:modular}

Our successful results training neural networks on simple equality suggested another strategy for solving the hierarchical equality task. Rather than requiring our networks to find solutions from scratch, we pretrained them on basic equality tasks and then used those parameters as a starting point for learning hierarchical equality. This set of simulations was conceptually similar to our previous experiments with pretrained input representations, but now we pretrained an entire subpart of the model, rather than just input representations. This approach parallels the experimental paradigm used by \citet{thompson:1997}, in which chimpanzees that received pretraining on a basic equality (\emph{same}/\emph{different} judgment) task -- but not naive champanzees -- succeed in a hierarchical equality task.

\begin{figure*}
  \centering
  \begin{subfigure}[t]{0.48\linewidth}
    \centering
    \includegraphics[width=1\linewidth]{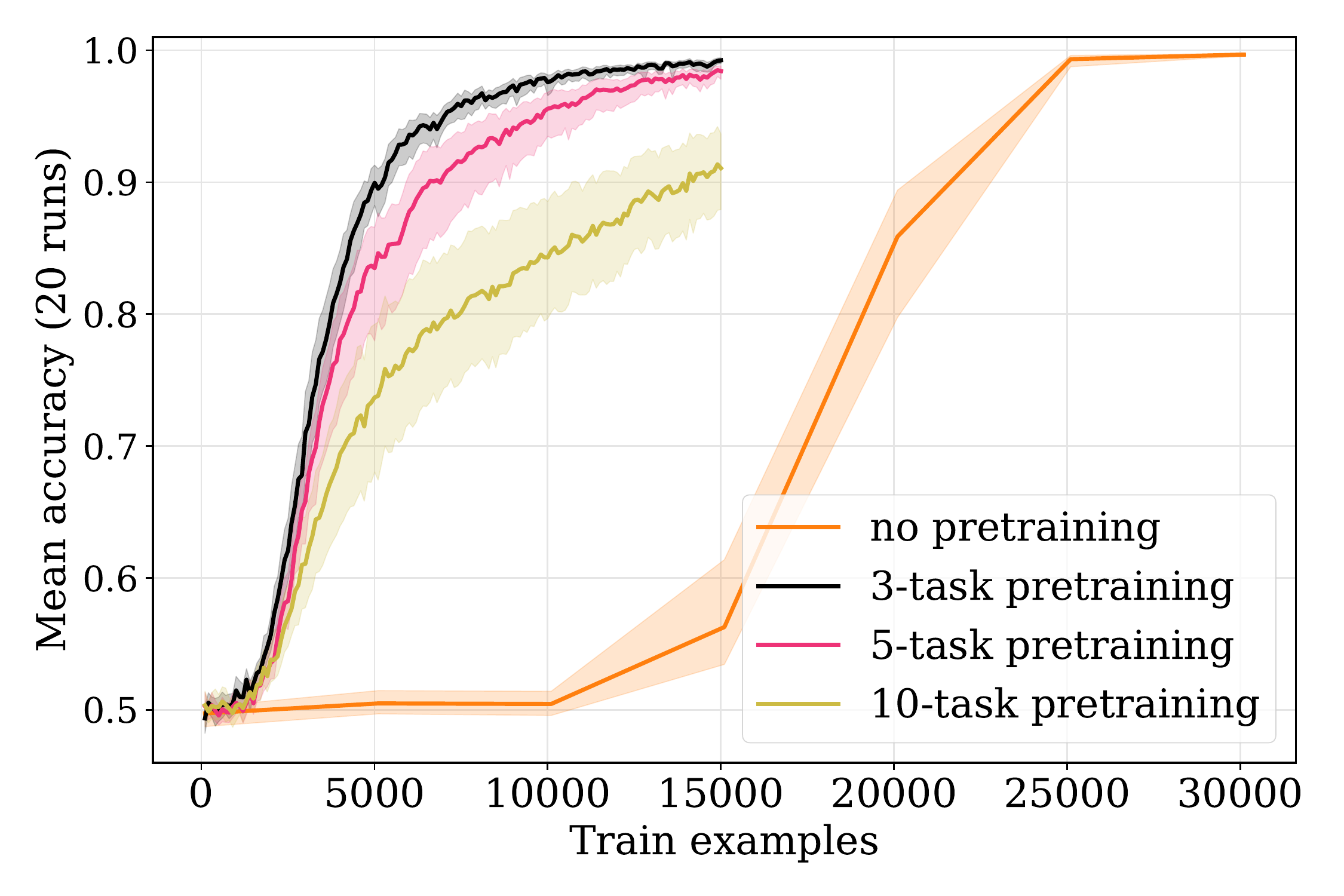}
    \caption{Experiment 3 results: feed-forward neural networks with two hidden layers trained on our hierarchical equality task (10-dimensional inputs; 100-dimensional hidden layers).}
    \label{fig:premack-h2-pretrain}
  \end{subfigure}
  \hfill
  \begin{subfigure}[t]{0.48\textwidth}
    \centering
    \includegraphics[width=1\linewidth]{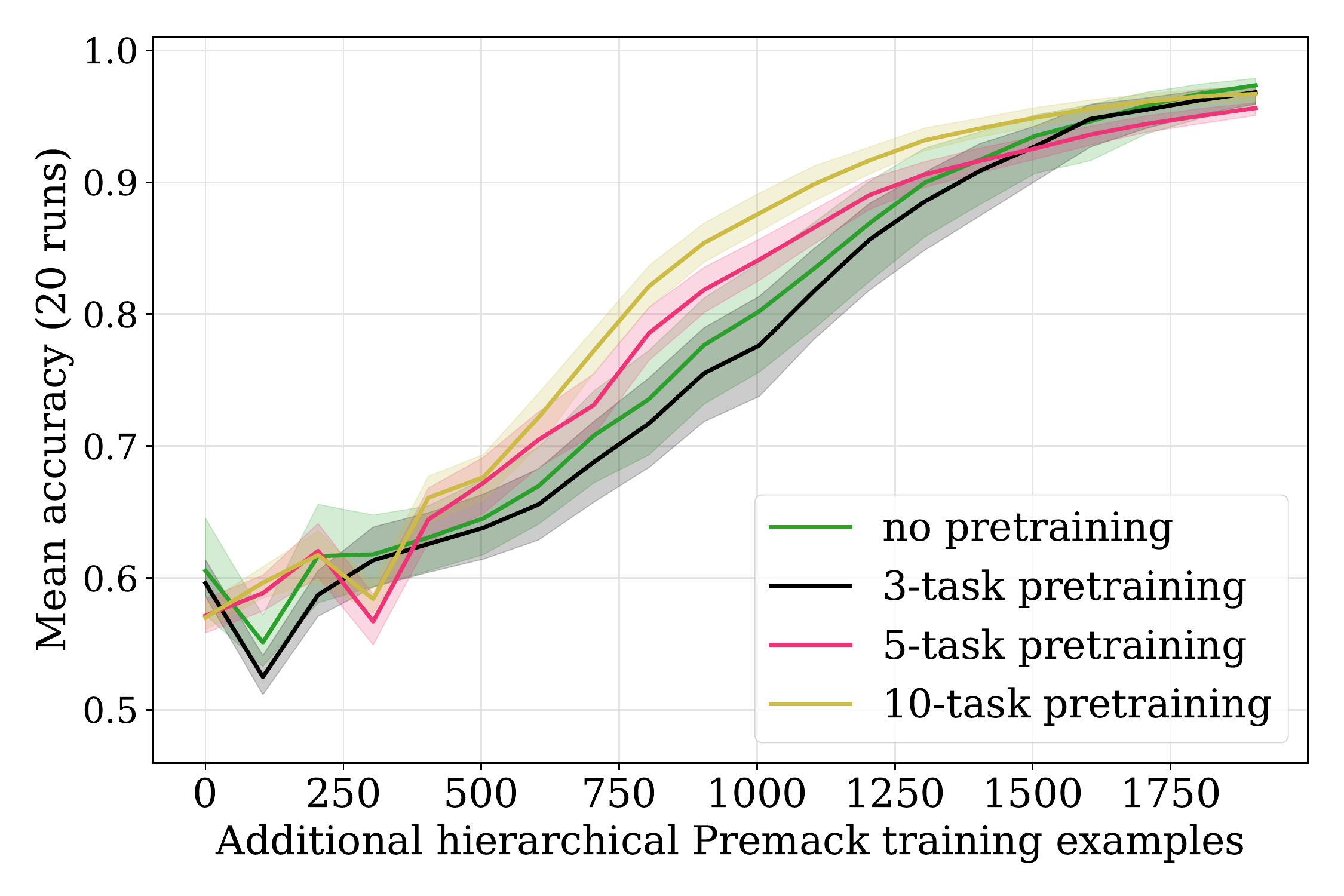}
    \caption{Experiment 4 results: simple equality networks (with 25-dimensional inputs and, in turn, 50-dimensional hidden representations) applied to the hierarchical equality task as pretrained components. Even with no additional training instances for this task (``zero shot''), all models achieve greater than chance accuracy, and even modest amounts of additional training on the task lead to excellent performance. Additionally pretraining the input representations leads to modest improvements.}
    \label{fig:premack-pretraining-results}
  \end{subfigure}
  \caption{Results for (a) hierarchical sequential equality task without pretraining on simple equality and (b) with pretraining on simple equality. The `no pretraining' models are provided random representations and the `$k$-task pretraining' models are provided random representations that are grounded in $k$ binary property domains via pretraining learning tasks. The Experiment~3 model requires vastly more task-specific training data than the Experiment~4 model to achieve good results, which helps to show the value of modular reuse of pretrained components.}
\end{figure*}

\subsection{Methods}

The hierarchical equality task requires computing the equality relation three times: compute whether the first two inputs are equal, compute whether the second two inputs are equal, then compute whether the truth-valued outputs of these first two computations are equal. We propose to use the same network pretrained on basic equality to perform all three equality computations. The model is depicted in \figref{fig:models:premack} and defined as follows:
\begin{align}
h_1 &= \ReLU([a;b]W_{xh} + b_{h}) \label{eq:e4:1}\\
h_2 &= \ReLU([c;d]W_{xh} + b_{h})\\
h_3 &= \ReLU([h_1;h_2]W_{xh} + b_{h}) \\
y &= \softmax(h_3W_{hy} + b_{y}) \label{eq:e4:y}
\end{align}
Here, $W_{xh}$, $W_{hy}$, $b_h$, and $b_y$ are the parameters from the model in equations \dasheg{eq:x2h}{eq:h2y} already trained on basic equality. Crucially, the same parameters, $W_{xh}$ and $b_h$, are used three times: twice to compute representations encoding whether a pair of input entities are equal ($h_1$, $h_2$), and once to compute a representation ($h_{3}$) encoding whether the truth values encoded by $h_1$ and $h_2$ are equal. This final representation is then used to compute a probability distribution over two classes, and the class with the higher probability is predicted by the model.\footnote{\blue{A clear limitation of this approach is that it presupposes the hierarchical nature of the task rather than learning it from the data. Recent developments with modular networks and routing networks might open up a path to removing this limitation \citep{Andreas-etal:2016,chang2018automatically,Cases-etal:2019,Kirsch-etal:2018}. We briefly return to this point in our `General discussion' below.}}

We implemented this network using PyTorch using the same protocols as in Experiments 1--3 (see \SIref\ for details). In the main text, we graph results for the single best hyperparameter setting. The training and testing protocol matches the one used in Experiments~3.

\subsection{Results}

\Figref{fig:premack-pretraining-results} shows our results for networks with 25-dimensional inputs, which sets the dimensionality of the hidden representations to 50. (See \Figref{fig:model3a} for results for other network dimensionalities.) All the models have above chance performance after being trained only on the simple equality task -- that is, they achieve zero-shot generalization to the hierarchical task and within two thousand examples, the models achieve near perfect accuracy. Additionally pretraining the input representations leads to modest but consistent gains in terms of learning speed. (In \figref{fig:train-results:e4}, we provide comparable training set assessments; in that context, pretraining has the opposite effect, slightly slowing down learning.)

\subsection{Discussion}

It is remarkable that a model trained only on equality between entities is able to get traction on a problem that requires determining whether equality holds between the truth values encoded in two learned representations. What is more, even small amounts of additional training data lead to excellent performance on the task. This shows the value of pretraining regimes involving entire networks, and it begins to show how trained networks can serve as modular components that are useful for solving more complex tasks.

Intuitively, this problem presents four different input class-types: the two `positive' classes (same/same and different/different), and the two `negative' classes (same/different and different/same). We see noteworthy variation in model performance across these classes: where there is little or no training data, accuracy is high for the `negative' classes and low for the `positive' ones. This pattern stabilizes quickly, and then we find that the different/different class is persistently the most difficult, which aligns with general qualitative findings from behavioral experiments. \Figref{fig:premack-by-class} depicts these findings for our best model.

\section{General Discussion}

Equality is a key case study for understanding the origins of human relational reasoning. This case study has been puzzling for symbolic accounts of reasoning because such accounts do not provide a compelling explanation for why some equality tasks are so easy to learn and others are so hard. In addition, evidence of graded learning and generalization in non-human species suggests that a gradual learning account might provide more traction in explaining the empirical data \cite{wasserman:2017}. Inspired by this work, we revisited a long-standing debate about whether neural network models can learn equality relations from data \cite{alhama:2019}.

Our work here makes three contributions to this debate. First, we show that non-featural representations -- both random and pretrained -- allow standard neural networks to learn simple, sequential, and hierarchical equality tasks. Both the research that originated this debate \cite{marcus:1999,marcus:2001,dienes:1999,seidenberg:1999a,seidenberg:1999b,elman:1999,negishi:1999} and more recent work \cite{weyde2019, weyde2018, kopparti2020, alhama2018} only involve experiments where featural representations are used. We suggest that this choice led directly to the negative conclusions from this body of work. Second, we show that neural networks can achieve high test accuracy on the sequential equality task with no negative feedback, suggesting that a negative feedback learning regime is not critical for learning equality. Finally, we show that a neural network trained only on simple equality can generalize to hierarchical equality, even in a ``zero-shot'' evaluation. Although pretrained representations sometimes led to faster learning, they were not a necessary component for models to succeed, and success was possible even using random representations.

\subsection{The implications of pretraining}

In some settings, our current models require many more training instances than humans seem to need. However, our pretraining approach suggests a path forward: by using pretrained models as modular components, we can get traction on challenging tasks without any training specifically for those tasks. In some cases, even a small amount of additional training can make a substantial difference.

One implication of our pretraining findings is that it should be possible to scaffold non-human animals' performance in complex, hierarchical equality tasks via training on simpler ones. Indeed, \citet{smirnova2015} show just this result in crows, consistent with our findings. Although we do not discount the potential role of linguistic labels in informing adult humans' expertise in such tasks \cite{gentner:2003}, pretraining also provides a potential account of how infants and young children might succeed in a range of equality reasoning tasks without access to specific linguistic symbols like ``same'' \cite{walker:2016,ferry:2015,hochmann2016}.

While pretraining in the sparse, artificial contexts of our models is far from accurately capturing the experience of infants and other animals, we believe it is reasonable to think of cognitive systems as “pretrained” by their perceptual experiences and by their evolved perceptual architecture. This “pretraining” might lead them to have continuous representations of perceptual information that are non-featural in the sense of our models, but which still encode psychologically meaningful properties. In this sense, human generalizations could be similar to those made by our models in that both encode meaningful properties of stimuli in a way that is not perfectly individuated or discrete. Further, both can make rapid (and sometimes one-shot) generalizations, especially in cases where the agent has had prior experience with the stimuli \citep{rabagliati:2019}. This kind of viewpoint on our pretraining simulations is consistent with a range of views of how complex human abilities are created from simpler one \cite[e.g.,][]{frank2008,heyes2018}.

We have not offered a theory of how an agent might come to be able to combine pretrained components to solve more complex tasks, however. Rather, we simply pieced these components together by hand to achieve our desired outcomes. A natural next step would be to learn how to effectively combine these modular pieces. The artificial intelligence literature currently offers a range of techniques that could be used for this purpose, including Neural Module Networks \citep{Andreas-etal:2016}, Compositional Recursive Learners \citep{chang2018automatically}, Modular Networks \cite{Kirsch-etal:2018}, and Recursive Routing Networks \citep{Cases-etal:2019}. These techniques might provide a fruitful avenue for simulating the emergence of more complex cognitive abilities.

\subsection{Beyond exact equality}

We have so far confined our discussion to idealized notions of \emph{same} and \emph{different} that are defined in terms of complete mathematical identity. These are the terms of the original debate involving \citet{marcus:1999}, \citet{dienes:1999}, \citet{seidenberg:1999a,seidenberg:1999b}, \citet{elman:1999}, \citet{negishi:1999}, and others. We hope to have established that neural networks can solve this version of the problem. However, the cognitive notions of \emph{same} and \emph{different} are richer and more multifaceted than this.

Our models and findings hold for the case where judgments are made about one and the same image, but it is an open question how to extend them to situations in which one has two distinct images of the same entity and the intended label is ``same''. Even this basic expansion of the problem space poses many challenging new questions concerning context dependence (e.g., when are two different apples of the same type judged to be ``the same'') and human fallibility (one could fail to perceive relevant differences or fail to ignore irrelevant ones), and thus the very notion of success is also likely to become more graded and context dependent. \citet{webb2021emergent} begin to explore this problem, and those initial results give us confidence that the kinds of networks we propose here could be conditioned contextually to provide flexible, context-sensitive interpretations for sameness and graded similarity \cite{medin1993}, holding the promise of unifying models of identity, similarity, and related notions.

More broadly still, our work suggests a possible way forward in understanding the acquisition of logical semantics. Graded logical functions like those our models learned here could form the foundation for a semantics of words like ``same'' \cite{potts2019}. Such an option is appealing because it escapes from the circularity of defining the semantics of linguistic symbols as originating in a mental primitive {\sc same}. A semantics for ``same'' requires defining its inputs and outputs as well as how it composes with other symbols. Further, understanding the pragmatics of ``same'' then requires understanding how exact the similarity must be for the statement to be true in particular contexts.

\subsection{Conclusions}

Even the best present-day deep learning models often fail to achieve systematic and general solutions \citep{Lake2018, Marcus2018,geiger2019,Hupkes2020,linzen2020,wu2021}. However, as our experiments demonstrate, abstract equality reasoning is too simple of a domain to expose this failure. Indeed, at this point, even very complicated problems in logical semantics might not suffice to make this point, given the high level of performance that can sometimes be achieved \citep{Bowman:Potts:Manning:2015,mul2019siamese}.

Earlier debates about the nature of equality computations centered around the question of whether models included explicitly symbolic elements. We believe ours do not; but it is of course possible to quibble with this judgment. For example, since the supervisory signals used in Experiments 1 and 3 are generated based on a symbolic rule, perhaps that makes these models symbolic under some definition. (Of course, the same argument could be applied to the supervision signal that is provided to crows, baboons, and human children in some tasks).

This kind of argument increasingly seems terminological, rather than substantive, however. Neural models may learn symbolic computations in ways that do not contain explicit operators but that nevertheless have elements that play the same causal role. \citet{geiger2020} and \citet{geiger2021} perform causal abstraction analyses of trained neural models and find that they define and use symbolic variables at an abstract level of computation. In related work, \citet{Geiger:Wu:Lu-etal:2021} train neural models to implement abstract symbolic computations in order to solve systematic generalization tasks. These techniques leverage recent innovations in models of causality and abstraction \citep{chalupka16,rubenstein17,beckers2019}, and they provide more evidence for the classic view that neural networks are capable of implementing symbolic computations \citep{smolensky1988,fodor1988} which has received recent focus again  \citet{Piantadosi:2021,Deepmind:2021}. So an advocate for a symbolic view could even say that such networks (and likely ours as well) have learned symbols, though this move might lead to more confusion than clarity with respect to the original debate.

In the end, the primary goal is not to resolve the question of whether explicit symbolic representations are necessary (although by most reasonable definitions we believe we have presented a negative answer to that question). Rather, the hope is to create an explicit learning theory for relational reasoning, which might in turn explain the rich and puzzling empirical landscape \cite{carstensen:inpress}. Our hope is that -- by providing a range of architectures that learn equality functions -- the work described here takes a first step in this direction.

\bibliographystyle{acl_natbib}
\bibliography{relational-learning-bib}

\newpage

\appendix

\section{Model optimization details}\label{app:optimization}

The feed forward networks for basic and hierarchical equality were implemented using the multi-layer perception from scikit-learn and a cross entropy function was used to compute the prediction error. The recurrent LSTM network for the sequential ABA task was implemented using PyTorch and a mean squared error function was used to compute the prediction error. The networks for pretraining representations and for the hierarchical equality task were also implemented using PyTorch, with cross-entropy loss functions used to compute the prediction errors.

For all models, Adam optimizers \citep{Kingma:Ba:2015} were used. For all models, we used a batch size of 1 and ran a hyperparameter search over learning rate values of \{0.00001, 0.0001, 0.001\} and l2 normalization values of \{0.0001, 0.001, 0.01\} for each hidden dimension and input dimension. We considered hidden dimensions of \{2, 10, 25, 50, 100\} and input dimensions of \{2, 10, 25, 50, 100\}. In the main text, we graph results for the single best hyperparameter setting, averaged over 20 runs with different random model and input intializations. See `Additional results plots' below for details on how model performance is affected by changes in hidden dimensionality and input dimensionality.

\section{An analytic solution to identity with a feed-forward network}\label{app:equality-solution}

We now show that our feed-forward networks with one hidden layer can solve the same--different problem we pose, in the following sense: for any set of inputs, we can find parameters $\theta$ that perfectly classify those inputs. At the same time, we also show that there are always additional inputs for which $\theta$ makes incorrect predictions.

Here are the parameters of a feed forward neural network that performs a binary classification task
\[
  \texttt{ReLu}
  \left(
    \begin{pmatrix} x_1 \\ x_2  \end{pmatrix}^T
    \begin{pmatrix} W^{11} & W^{12}\\ W^{21}& W^{22} \end{pmatrix}
  \right)
  \begin{pmatrix} v^{11} & v^{12} \\ v^{21} & v^{22} \end{pmatrix} +
  \begin{pmatrix}b_1 &b_2 \end{pmatrix} =
  \begin{pmatrix} o_1 & o_2\end{pmatrix}\]
where, if $n$ is the dimension of entity embeddings used, then
\begin{align*}
  x, y,v^{11}, v^{12}, v^{21}, v^{22} &\in \mathbb{R}^{n \times 1} \\
  W^{11}, W^{12},W^{21}, W^{22} &\in \mathbb{R}^{n \times n} \\
  b_1, b_2, o_1, o_2 &\in \mathbb{R}
\end{align*}
Given an input $(x_1,x_2)$, if the output $o_1$ is larger than $o_2$, then one class is predicted; if the output $o_2$ is larger that $o_1$, then the other class is predicted. When the two outputs are equal, the network has predicted that both classes are equally likely and we can arbitrarily decide which class is predicted. In this case, the output $o_1$ predicts the two inputs, $x_1$ and $x_2$, are in the identity relation and the output $o_2$ predicts the two inputs are not.

\begin{figure}[h]
  \centering
  \newcommand\X{2}
  \newcommand\E{0.03}
  \begin{tikzpicture}[scale=0.7]
    \centering
    \filldraw[fill=black, opacity=0.1]
    (-5,-5 + \X)--(5-\X,5) -- (5,5) -- (5,5 - \X)--(-5+ \X,-5)--(-5,-5)--(-5,-5 + \X);
    \draw[<->,ultra thick] (-5,0)--(5,0) node[right]{$x_1$};
    \draw[<->,ultra thick] (0,-5)--(0,5) node[above]{$x_2$};
    \draw[<->,thick] (-5,-5)--(5,5) ;
    \draw[<->,thick, dashed] (-5,-5 + \X)--(5-\X,5) ;
    \draw[<->,thick,dashed] (-5+ \X,-5)--(5,5 - \X) ;
    \draw [decorate,decoration={brace,amplitude=12pt},xshift=-0pt,yshift=0pt]
    (3 - \X/2 + \E ,3 + \X/2- \E) -- (3-\E,3+\E) node [black,midway, xshift=12pt,yshift=12pt]
    { \rotatebox{-45}{$\scriptstyle b_1 - b_2$}};
  \end{tikzpicture}
  \caption{A visual representation of how the analytic solution to identity of a single layer feed forward network changes depending on the values of two bias terms, $b_1,b_2$.}
  \label{fig:analyticff}
\end{figure}
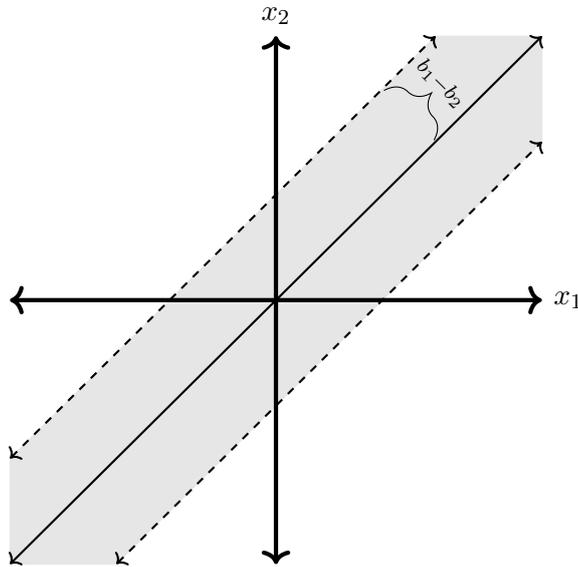

Now we specify parameters to provide an analytic solution to the identity relation using this network:
\[
  \texttt{ReLu}
  \left(
    \begin{pmatrix} x_1 \\ x_2 \end{pmatrix}^T
    \begin{pmatrix} I & -I\\ -I& I \end{pmatrix}
  \right)
  \begin{pmatrix} \vec{1} & \vec{0} \\ \vec{1} & \vec{0} \end{pmatrix} +
  \begin{pmatrix}b_1 &b_2 \end{pmatrix} =
  \begin{pmatrix} o_1 & o_2\end{pmatrix}
\]
where $I$ is the identity matrix, $-I$ is the negative identity matrix, and $\vec{1}$ and $\vec{0}$ are the two vectors in $\mathbb{R}^n$ that have all zeros and all ones, respectively. The output values, given an input, are
\[ o_1 = \sum_{i = 1}^{n}|(x_1)_i- (x_2)_i|+ b_1 \qquad \quad  o_2 = b_2\]
where two parameters are left unspecified, $b_1, b_2$. We present a visualization in Figure~\ref{fig:analyticff} of how the analytic solution to identity of this network changes depending on the values of two bias terms. In this example, the network receives two one-dimensional inputs, $x_1$ and $x_2$. If the ordered pair of inputs is in the shaded area on the graph, then they are predicted to be in the identity relation. If in the unshaded area, they are predicted not to be. The dotted line is where the network predicts the two classes to be equally likely.

The network predicts $x_1$ and $x_2$ to be in the identity relation if $\sum_{i = 1}^{n}|x_i- y_i| < b_1-b_2$ which is visualized as the points between two parallel lines above and below the solution line $x_1 = x_2$. As the difference $b_1-b_2$ gets smaller and smaller, the two lines that bound the network's predictions get closer and closer to the solution line. However, as long as $b_1-b_2$ is positive, there will always be inputs of the form $(r,r+(b_1-b_2)/2)$ that are false positives. For any set of inputs, we can find bias values that result in the network correctly classifying those inputs, but for any bias values, we can find an input that is incorrectly classified by those values. In other words, we have an arbitrarily good solution that is never perfect. We provide a proof below that there is no perfect solution and so this is the best outcome possible. However, if we were to decide that, if the network predicts that an input is equally likely in either class, then this input is predicted to be in the identity relation, we could have a perfect solution with $b_1= b_2$.

Here is proof that a perfect solution is not possible. A basic fact from topology is that the set $\{x: f(x) < g(x)\}$ is an open set if $f$ and $g$ are continuous functions. Let $N_{o_1}$ and $N_{o_2}$ be the functions that map an input $(x_1,x_2)$ to the output values of the neural network, $o_1$ and $o_2$, respectively. These functions are continuous. Consequently, the set $C = \{(x_1,x_2): N_{o_2}(x_1,x_2) < N_{o_1}(x_1,x_2)\}$, which is the set of inputs that are predicted to be in the equality relation, is open.

With this fact, we can show that, if the neural network correctly classifies any point on the solution line $x_1 = x_2$, then it must incorrectly classify some point not on the solution line. Suppose that $C$ contains some point $(x,x)$. Then, by the definition of an open set, $C$ contains some $\epsilon$ ball around $(x,x)$, and therefore $C$ contains $(x,x+\epsilon)$, which is not on the solution line $x_1=x_2$. Thus, $C$ can never be equal to the set $\{(x_1,x_2):x_1=x_2\}$. So, because $C$ is the set of inputs classified as being in the equality relation by the neural network, a perfect solution cannot be achieved. Thus, we can conclude our arbitrarily good solution is the best we can do.

\section{An analytic solution to ABA sequences}\label{sec:analyticlm}

Here are the parameters of a long short term memory recurrent neural network (LSTM):
\begin{align*}
  f_t &= \sigma(x_t W_f  + h_{t-1} U_f  + b_f) \\
  i_t &= \sigma(x_t W_i  + h_{t-1} U_i + b_i) \\
  o_t &= \sigma(x_t W_o +  h_{t-1} U_o + b_o)\\
  c_t &= f_t \circ c_{t-1} + i_t \circ \ReLU(x_tW_c + h_{t-1}U_c + b_c) \\
  h_t &= o_t \circ \ReLU(c_t) \\
  y_t &= h_tV
\end{align*}
where, if $n$ is the representation size and $d$ is the network hidden dimension, then
\begin{align*}
  x_t \in \mathbb{R}^n, f_t, i_t, o_t,  h_t, c_t &\in \mathbb{R}^d\\
  W \in \mathbb{R}^{n \times d}, U &\in \mathbb{R}^{d \times d}\\
  V \in \mathbb{R}^{d \times n}, b &\in \mathbb{R}^d
\end{align*}
and $\sigma$ is the sigmoid function. The initial hidden state $h_0$ and initial cell state $c_0$ are both set to be the zero vector. We say that an LSTM model with specified parameters has learned to produce ABA sequences if the following holds: when the network is seeded with some entity vector representation as its first input, $x_1$, then the output $y_1$ is not equal to $x_1$ and at the next time step the output $y_2$ is equal to $x_1$.

We let $d = 2n + 1$ and assign the following parameters, which provide an analytic solution to producing ABA sequences:
\begin{align*}
  f_t &= \sigma(x_t \textbf{0}_{n\times d} + h_{t-1}\textbf{0}_{d \times d} + \textbf{N}_{d}) \\
  i_t &= \sigma(x_t\textbf{0}_{n\times d} + h_{t-1}\begin{bmatrix} -4 \dots -4 \\ \textbf{0}_{2n\times n} \end{bmatrix}  + \textbf{N}_d)   \\
  o_t &= \sigma(x_t\textbf{0}_{n\times d} +  h_{t-1} \begin{bmatrix} 1 \dots 1 \\ \textbf{0}_{2n\times n} \end{bmatrix} + \textbf{0}_{d})\\
  c_t &= f_t \circ c_{t-1} + \\
  & i_t \circ \ReLU\left(x_t \begin{bmatrix} 0 &  & \\ \vdots & -I_{n \times n} & I_{n \times n}\\ 0 & &  \end{bmatrix} + h_{t-1}\textbf{0}_{d \times d} + \begin{bmatrix} N & 0 \dots 0\end{bmatrix}\right) \\
  h_t &= o_t \circ \ReLU(c_t)\\
  y &= h_t \begin{bmatrix} 0 \dots 0 \\ -I_{n \times n} \\ I_{n \times n}  \end{bmatrix}
\end{align*}
Where $\textbf{0}_{j\times k}$ is the $j \times k$ zero matrix, $\textbf{m}_k$ is a $k$ dimensional vector with each element having the value $m$, $I_{n\times n}$ is the $n \times n$ identity matrix, and $N$ is some very large number.  Now we show that these parameters achieve an increasingly good solution as $N$ increases. When a value involves the number $N$, we will simplify the computation by saying what that value is equal to as $N$ approaches infinity. We begin with an arbitrary input $x_1$ and the input and hidden state intialized to zero vectors:
\[
  h_0 = \textbf{0}_d \qquad c_0 = \textbf{0}_d
\]
The gates at the first time step are easy to compute, as the cell state and hidden state are zero vectors so the gates are equal to the sigmoid function applied to their respective bias vectors. The forget gate is completely open, the output gate is partially open, and the input gate is fully open:
\begin{align*}
  f_1 &= \sigma(\textbf{N}_d) \approx \textbf{1}_d \\
  o_1 &= \sigma(\textbf{0}_d) = \textbf{0.5}_d \\
  i_1 &=  \sigma(\textbf{N}_d) \approx \textbf{1}_d
\end{align*}
Then we compute the cell and hidden states at the first timestep. The cell state encodes the information of the input vector, so it can be used to recover the vector at a later time step and receives no information from the previous cell state despite the forget gate being open, because the previous cell state is a zero vector. The hidden state is the cell state scaled by one half.
\begin{align*}
  c_1 &= \textbf{1}_d\circ \textbf{0}_d+ \\
      & \textbf{1}_d \circ \ReLU\left(x_1\begin{bmatrix} 0 &  & \\ \vdots & -I_{n \times n} & I_{n \times n}\\ 0 & &  \end{bmatrix} + \begin{bmatrix} N & 0 \dots 0\end{bmatrix}\right) \\
      & = \ReLU\left( \begin{bmatrix} N & -x_1 & x_1 \end{bmatrix}\right) \\[1ex]
  h_1 &= \textbf{0.5}_d\ReLU \left(\ReLU( \begin{bmatrix} N & -x_1 & x_1 \end{bmatrix})\right) \\
      &= \textbf{0.5}_d\circ \ReLU( \begin{bmatrix} N & -x_1 & x_1 \end{bmatrix})
\end{align*}
At the next time step, the forget gate remains fully open, the output gate changes from partially open to fully open, and the input gate changes from fully open to fully closed:
\begin{align*}
  f_2 &= \textbf{1}_d \\[2ex]
  o_2 &= \sigma(x_2\textbf{0}_{n\times d} + h_{1}\begin{bmatrix} 1 \dots 1 \\ \textbf{0}_{2n\times n} \end{bmatrix}  + \textbf{0}_d) \\
  &= \sigma(\textbf{0.5}_d\circ \ReLU\left( \begin{bmatrix} N & -x_1 & x_1 \end{bmatrix}\right)\begin{bmatrix} 1 \dots 1 \\ \textbf{0}_{2n\times n} \end{bmatrix}  + \textbf{0}_d)  \\
  &=  \sigma(\textbf{0.5}_d\circ \textbf{N}_d) \approx \textbf{1}_d \\[2ex]
  i_2 &= \sigma(x_2\textbf{0}_{n\times d} + h_{1}\begin{bmatrix} -4 \dots -4 \\ \textbf{0}_{2n\times n} \end{bmatrix}  + \textbf{N}_d)  \\
  &= \sigma(\textbf{0.5}_d\circ \ReLU\left( \begin{bmatrix} N & -x_1 & x_1 \end{bmatrix}\right)\begin{bmatrix} -4 \dots -4 \\ \textbf{0}_{2n\times n} \end{bmatrix}  + \textbf{N}_d)   \\
  &=  \sigma(\textbf{0.5}_d\circ\textbf{-4N}_d + \textbf{N}_d) \approx \textbf{0}_d
\end{align*}
Then we compute the cell and hidden states for the second timestep. Because the forget gate is completely open and the input gate is completely closed, the cell state remains the same. Because the output gate is completely open, the hidden state is the same as the cell state.
\begin{align*}
  c_2 &= \textbf{1}_d \circ \ReLU \left( \begin{bmatrix} N & -x_1 & x_1 \end{bmatrix}\right) +\\
      & \textbf{0}_d\circ \ReLU(x_2\begin{bmatrix} 0 &  & \\ \vdots & -I_{n \times n} & I_{n \times n}\\ 0 & &  \end{bmatrix} + \begin{bmatrix} N & 0 \dots 0\end{bmatrix}) \\
      &= \ReLU \left( \begin{bmatrix} N & -x_1 & x_1 \end{bmatrix}\right) \\[2ex]
  h_2 &= \textbf{1}_d \circ \ReLU\left(\ReLU ( \begin{bmatrix} N & -x_1 & x_1 \end{bmatrix})\right) \\
      &= \ReLU \left( \begin{bmatrix} N & -x_1 & x_1 \end{bmatrix}\right)
\end{align*}
With the hidden states for the first and second time steps, we can compute the output values and find that the output at the first time step is the initial input vector scaled by one half and the output at the second time step is the initial input vector.

\begin{align*}
  y_1 &= h_1\begin{bmatrix} 0 \dots 0 \\ -I_{n \times n} \\ I_{n \times n}  \end{bmatrix} = \textbf{0.5}_d\circ \ReLU\left( \begin{bmatrix} N & -x_1 & x_1 \end{bmatrix}\right) \begin{bmatrix} 0 \dots 0 \\ -I_{n \times n} \\ I_{n \times n}  \end{bmatrix} \\
      &= \textbf{0.5}_d\circ x_1\\
  y_2 &= h_2\begin{bmatrix} 0 \dots 0 \\ -I_{n \times n} \\ I_{n \times n}  \end{bmatrix} = \texttt{ReLu} ( \begin{bmatrix} N & -x_1 & x_1 \end{bmatrix}) \begin{bmatrix} 0 \dots 0 \\ -I_{n \times n} \\ I_{n \times n}  \end{bmatrix} = x_1
\end{align*}
Then, because $y_1 = \textbf{0.5}_d\circ x_1 \not = x_1$ and $y_2 = x_1$, this network produces ABA sequences.

\section{Additional results plots}

\subsection{Experiment~1 results for different hidden dimensionalities}\label{app:model1-results}

\begin{figure}
  \centering

  \begin{subfigure}{0.45\linewidth}
    \includegraphics[width=1\textwidth]{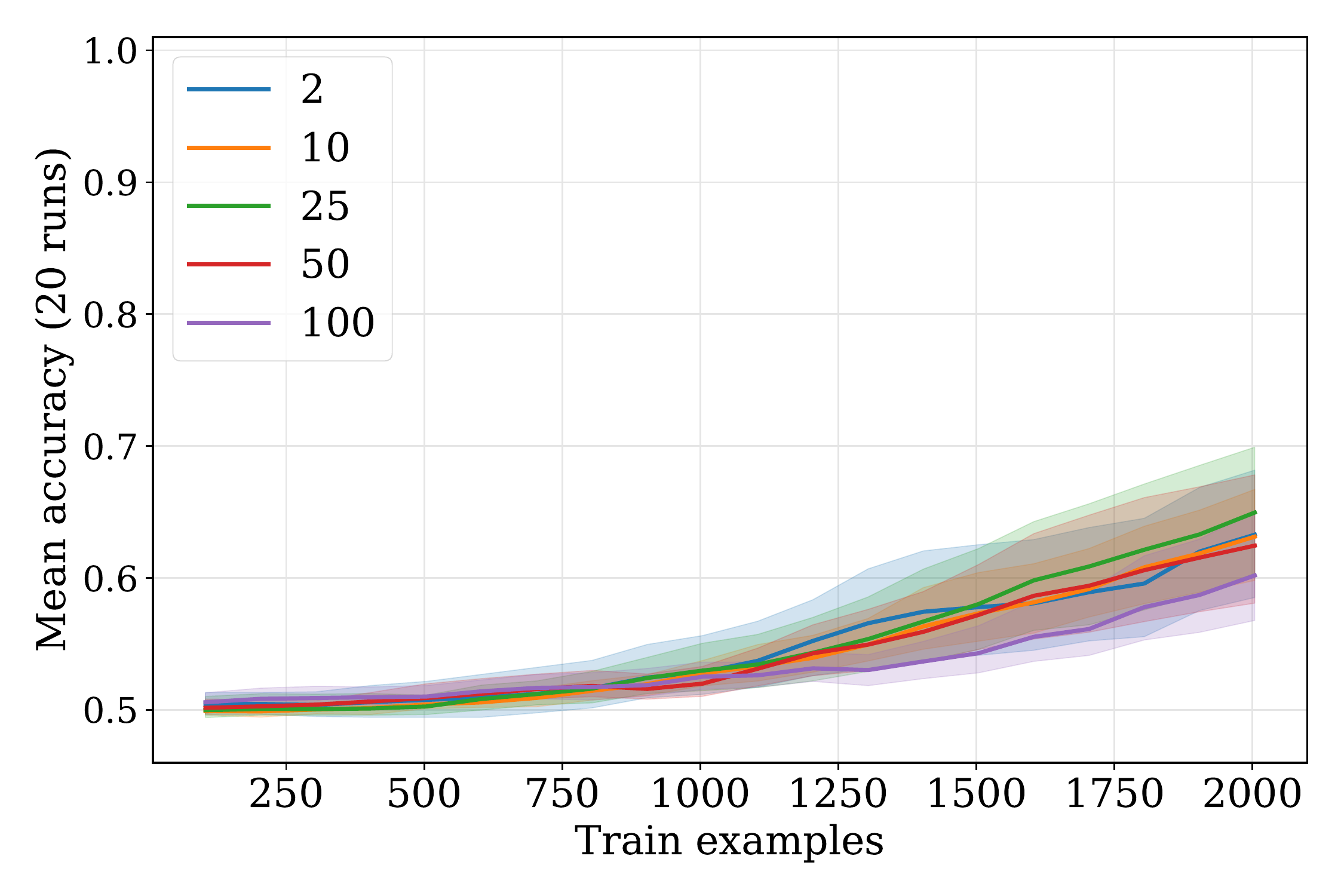}
    \caption{Hidden dimensionality 2.}
  \end{subfigure}
  \hfill
  \begin{subfigure}{0.45\linewidth}
    \includegraphics[width=1\textwidth]{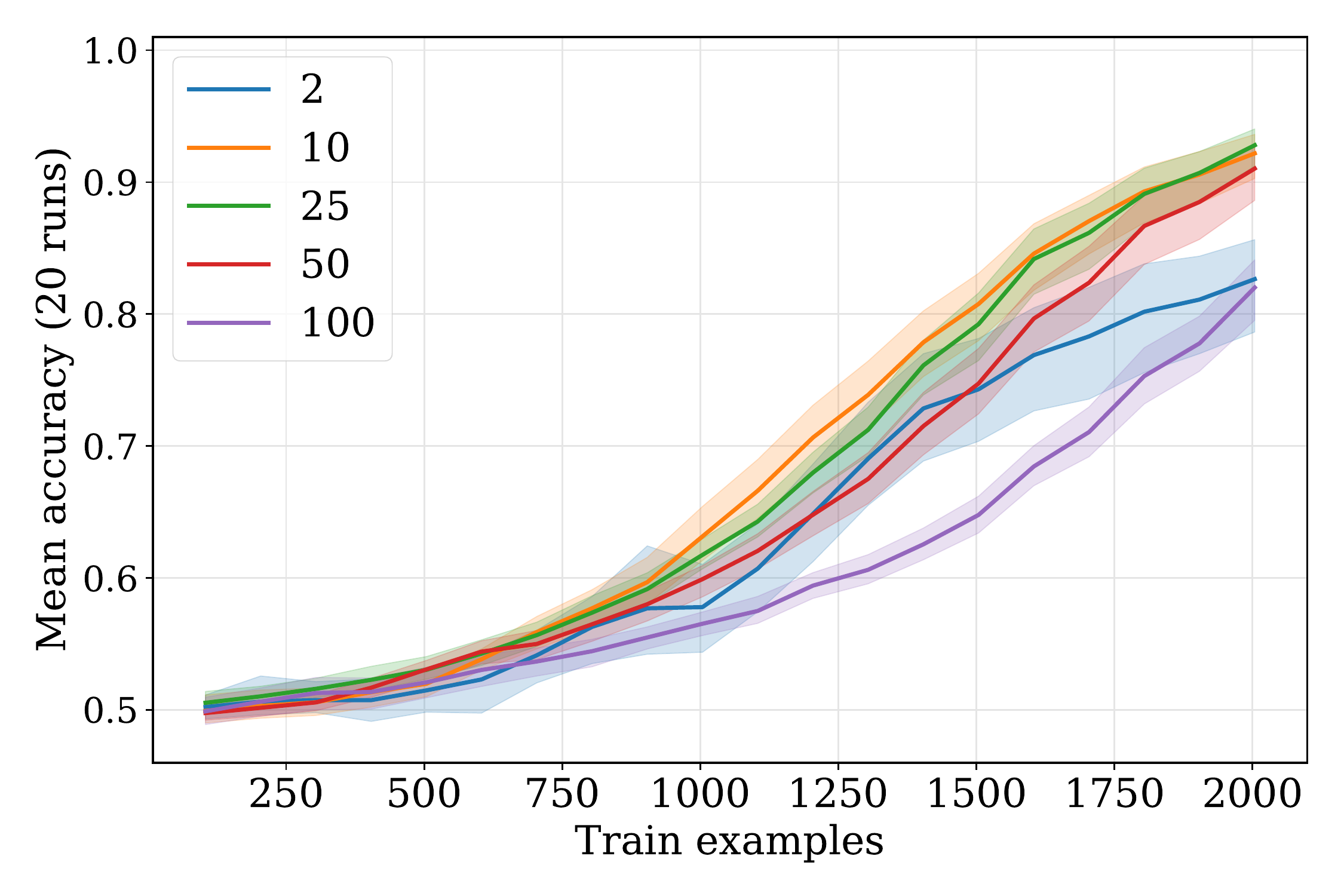}
    \caption{Hidden dimensionality 10.}
  \end{subfigure}

  \vspace{24pt}

  \begin{subfigure}{0.45\linewidth}
    \includegraphics[width=1\textwidth]{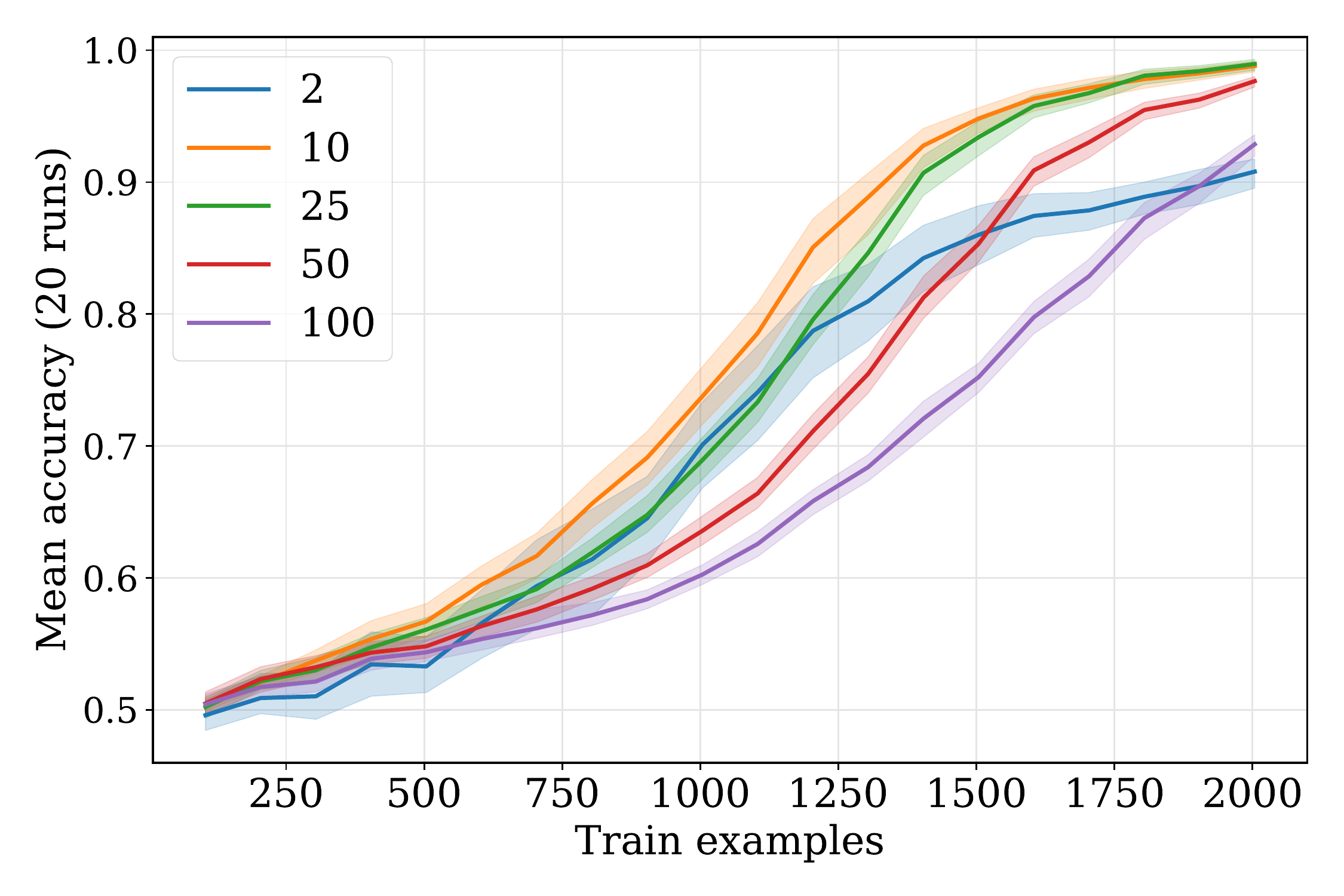}
    \caption{Hidden dimensionality 25.}
  \end{subfigure}
  \hfill
  \begin{subfigure}{0.45\linewidth}
    \includegraphics[width=1\textwidth]{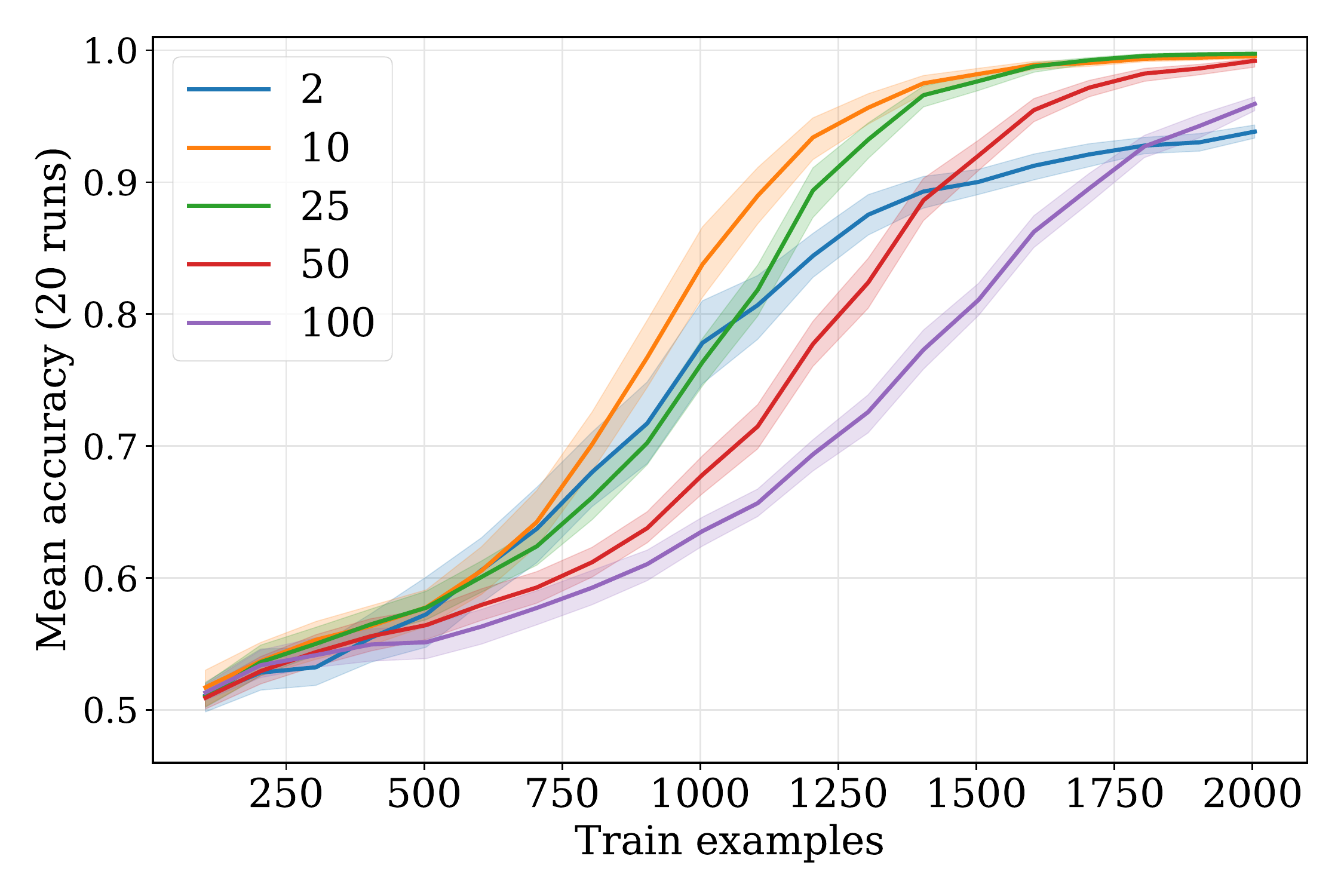}
    \caption{Hidden dimensionality 50.}
  \end{subfigure}

  \vspace{24pt}

  \begin{subfigure}{0.45\linewidth}
    \includegraphics[width=1\textwidth]{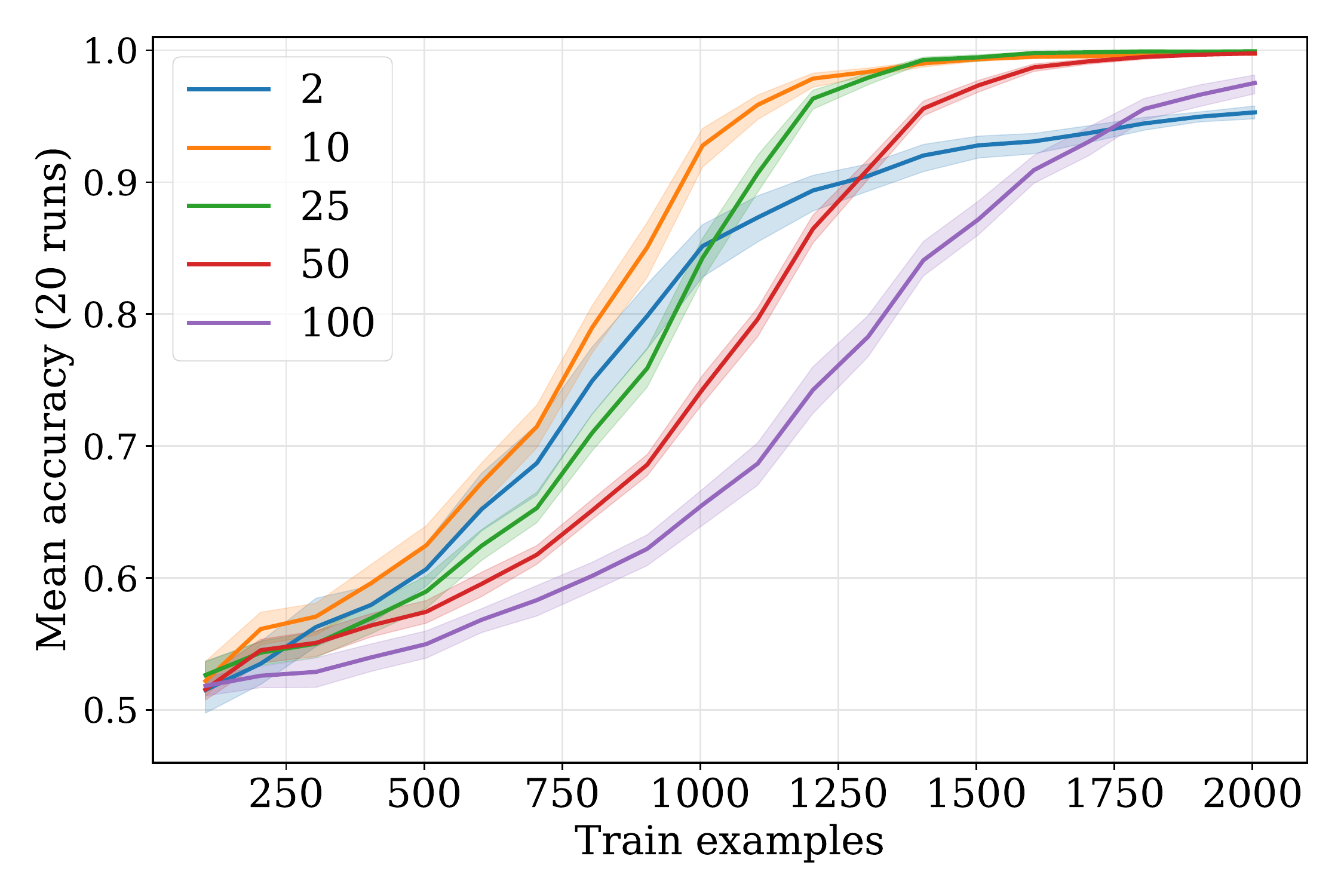}
    \caption{Hidden dimensionality 100.}
    \label{fig:model1-rep}
  \end{subfigure}
  \caption{Experiment~1 model for basic same--different. Lines correspond to different input dimensions. These models were provided random input representations.}
  \label{fig:model1}
\end{figure}

\Figref{fig:model1} explores a wider range of hidden dimensionalities for our Experiment~1 model (feed-forward network with a single hidden layer; equations~\dasheg{eq:x2h}{eq:h2y}) applied to the basic same--different task. The lines correspond to different embedding dimensionalities.

\subsection{Experiment~2 results for different hidden dimensionalities}

\begin{figure}
  \centering

  \begin{subfigure}{0.45\linewidth}
    \includegraphics[width=1\textwidth]{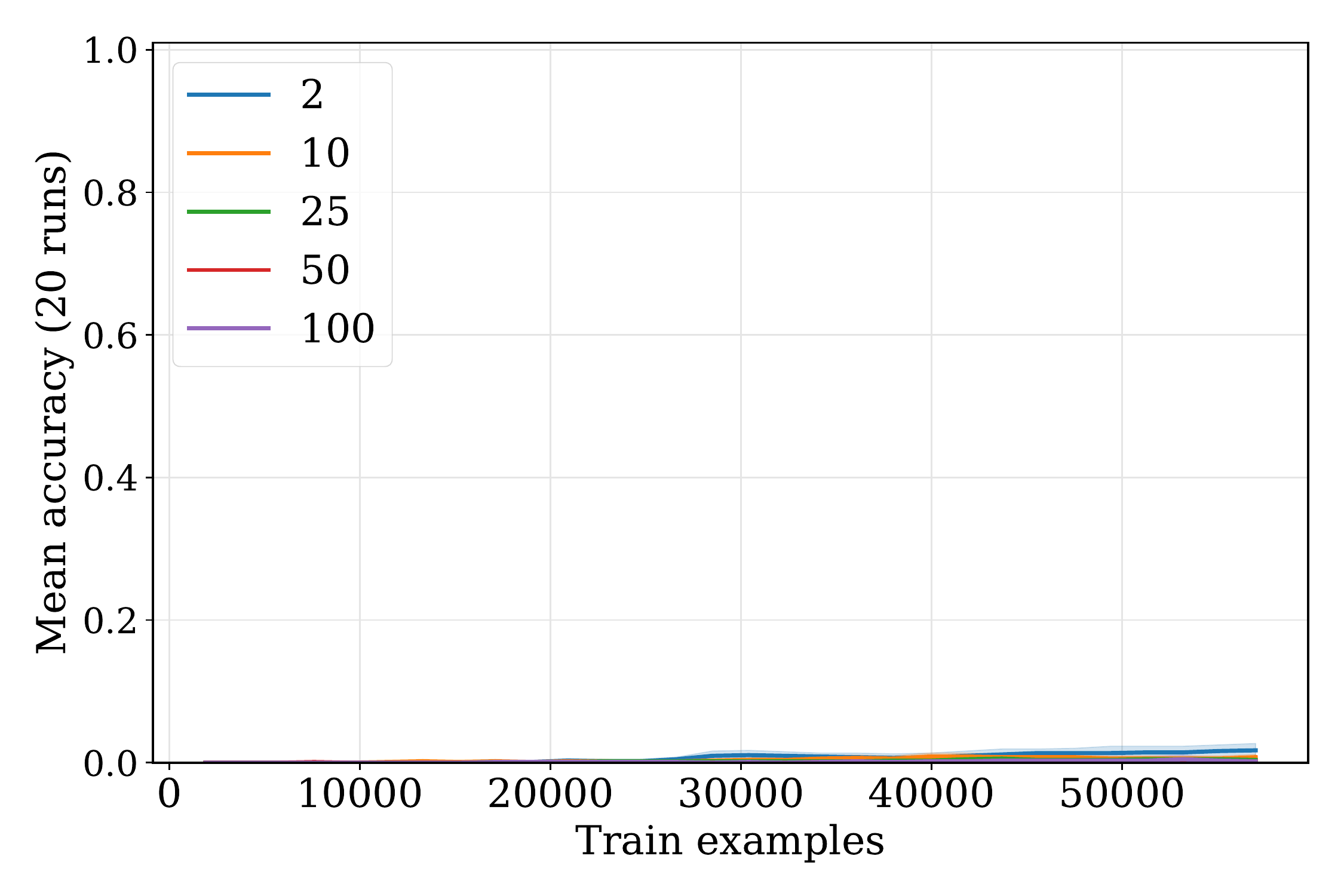}
    \caption{Hidden dimensionality 2.}
  \end{subfigure}
  \hfill
  \begin{subfigure}{0.45\linewidth}
    \includegraphics[width=1\textwidth]{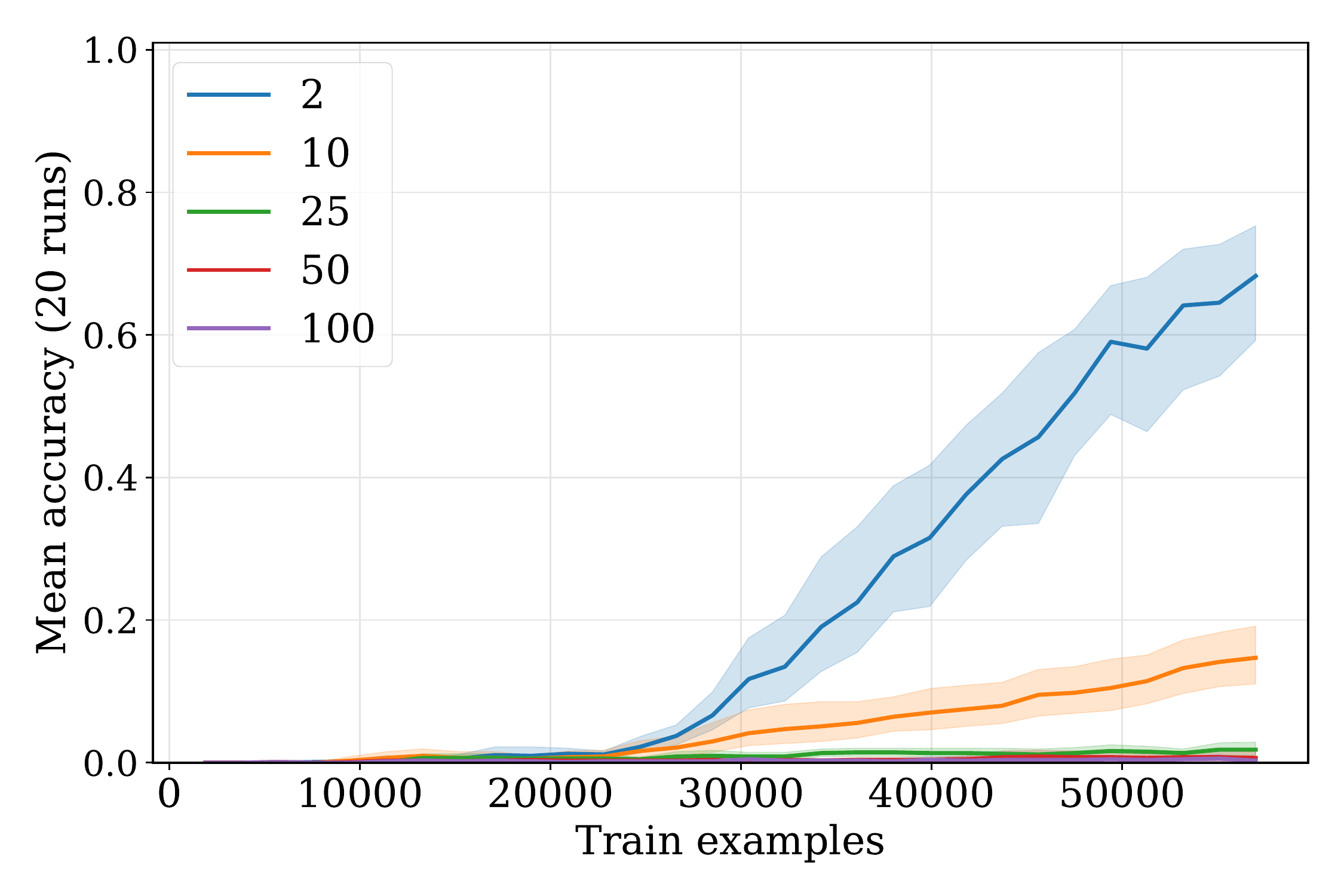}
    \caption{Hidden dimensionality 10.}
  \end{subfigure}

  \vspace{24pt}

  \begin{subfigure}{0.45\linewidth}
    \includegraphics[width=1\textwidth]{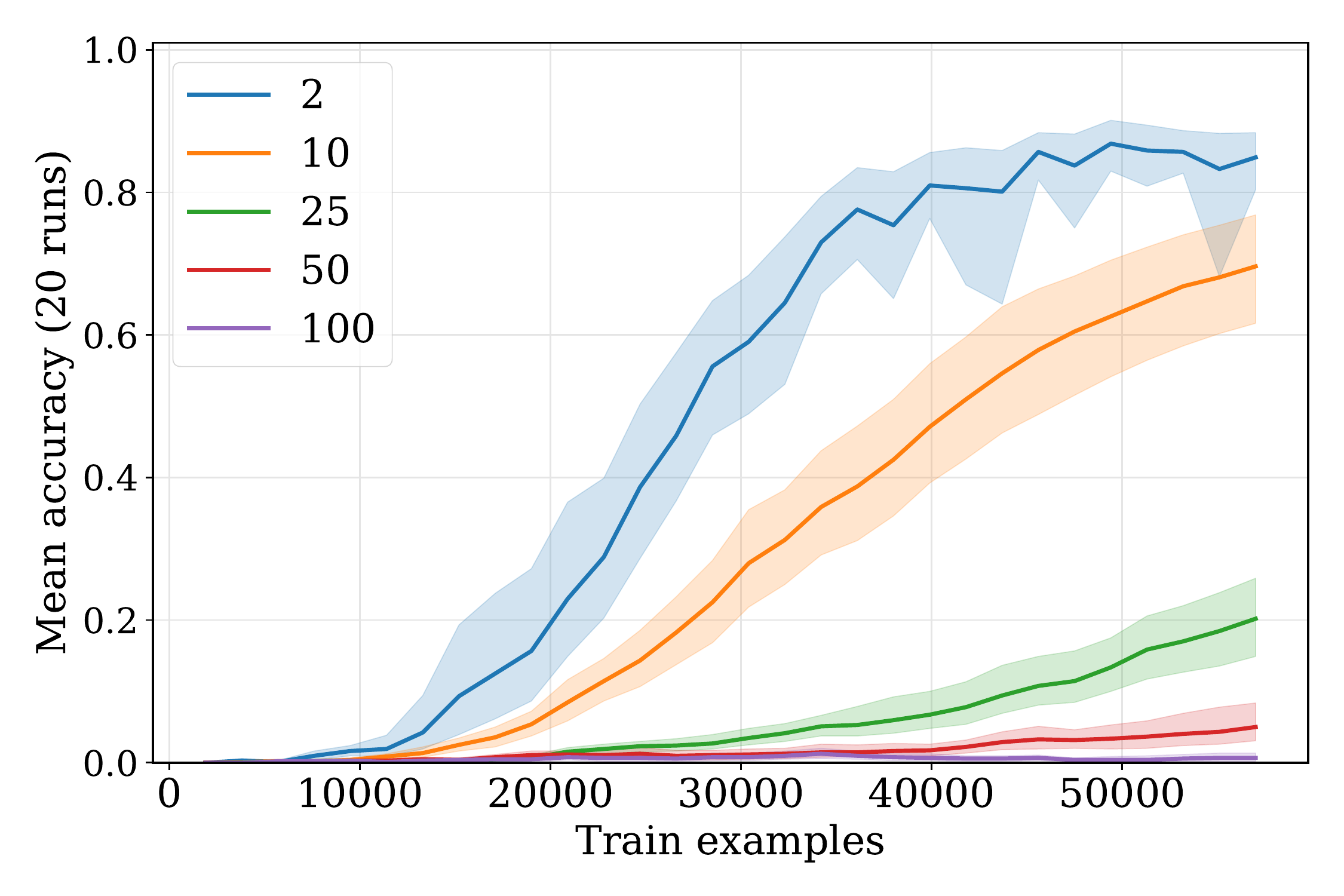}
    \caption{Hidden dimensionality 25.}
  \end{subfigure}
  \hfill
  \begin{subfigure}{0.45\linewidth}
    \includegraphics[width=1\textwidth]{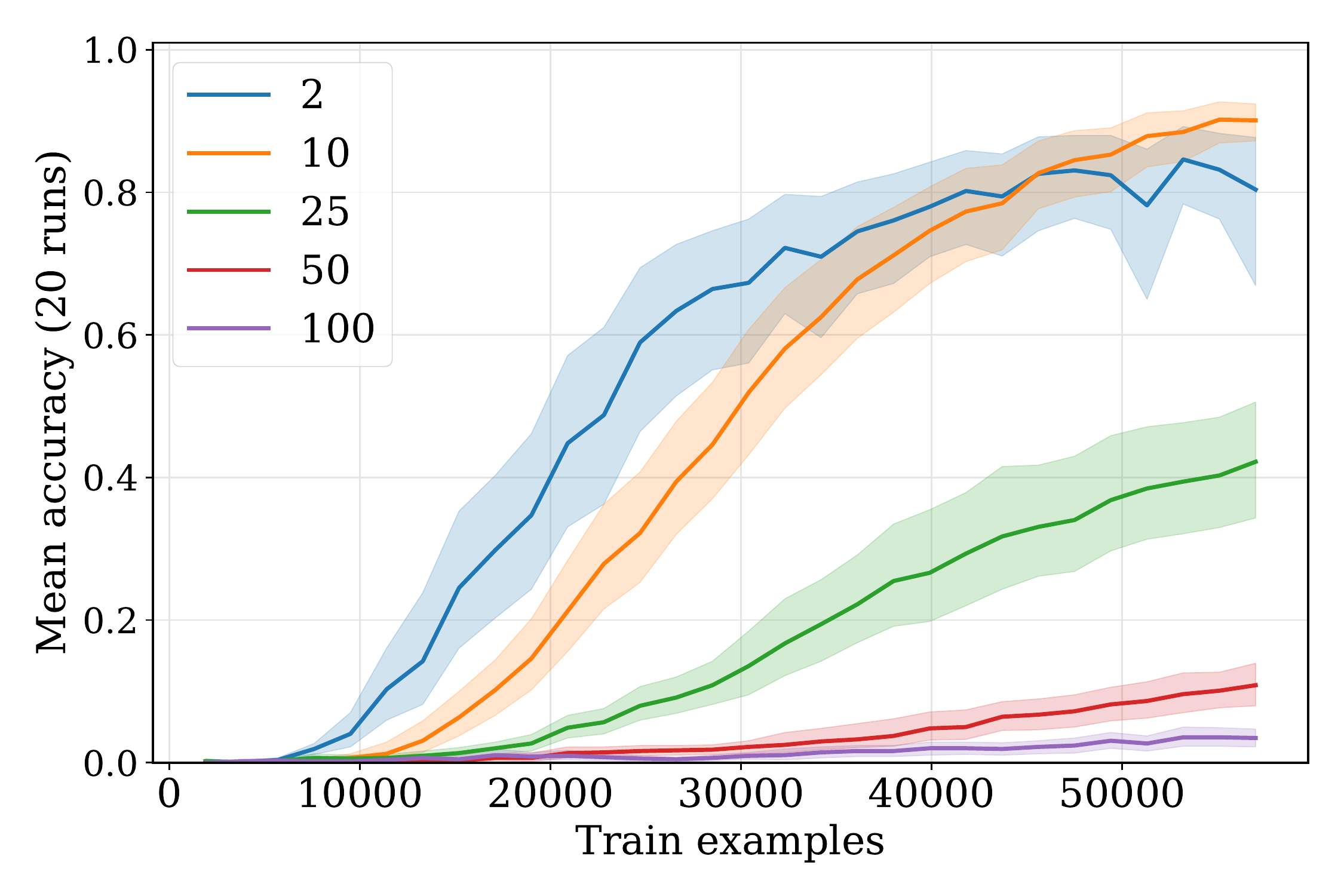}
    \caption{Hidden dimensionality 50.}
  \end{subfigure}

  \vspace{24pt}

  \begin{subfigure}{0.45\linewidth}
    \includegraphics[width=1\textwidth]{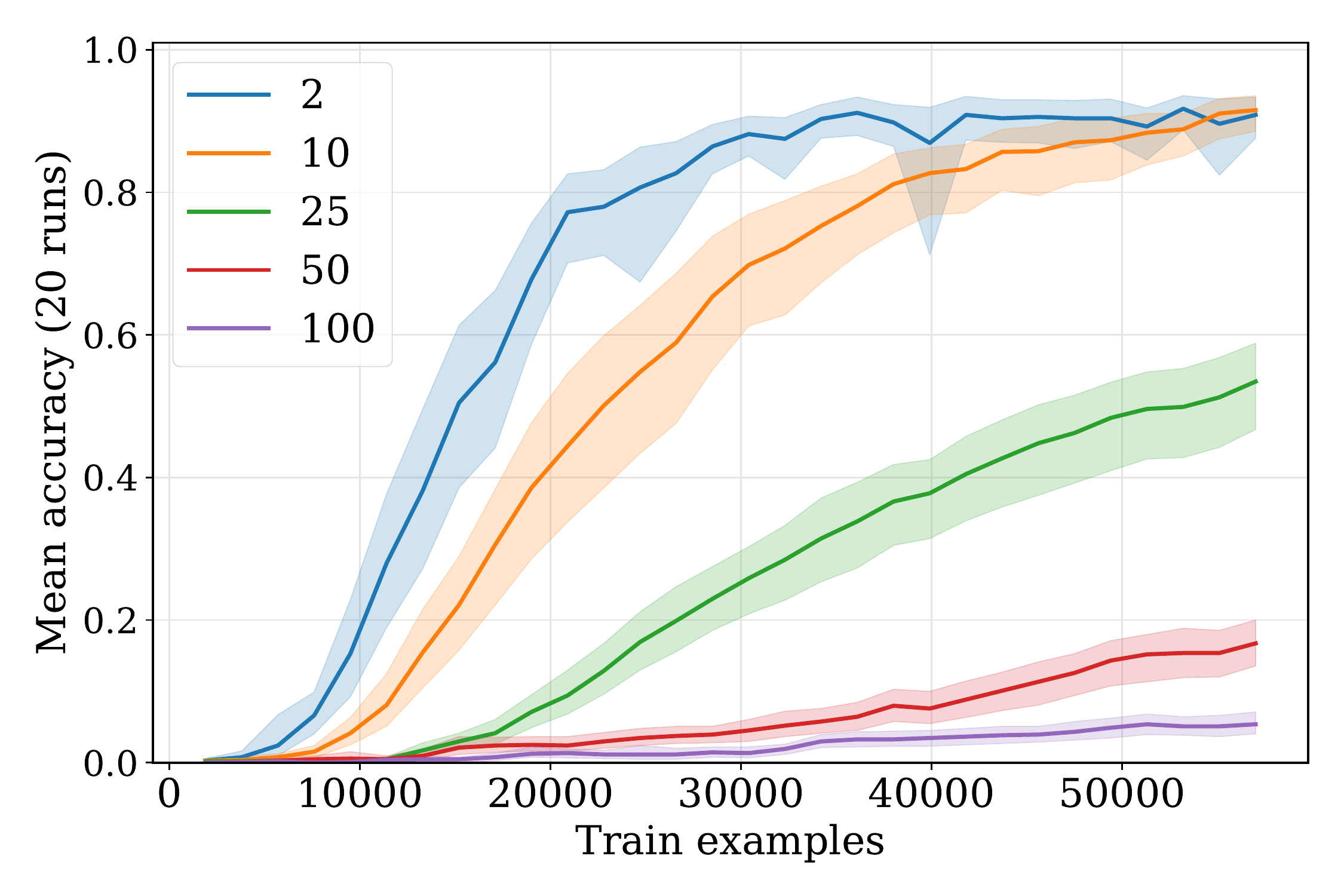}
    \caption{Hidden dimensionality 100.}
    \label{fig:model2-rep}
  \end{subfigure}

  \caption{Experiment~2 model for the sequential ABA task, with a vocabulary size of 20. Lines correspond to different input dimensions. These models were provided random input representations.}
  \label{fig:model2}
\end{figure}

\Figref{fig:model2} explores a wider range of hidden dimensionalities for our Experiment~2 model (LSTM; equations~\dasheg{eq:lstm-recur}{eq:lstm-predict}) applied to the sequential ABA task. The lines correspond to different embedding dimensionalities. The full training set is presented to the model in multiple epochs.

\subsection{Experiment~1 model applied to hierarchical same--different}\label{app:model1-premack}

\begin{figure}
  \centering

  \begin{subfigure}{0.45\linewidth}
    \includegraphics[width=1\textwidth]{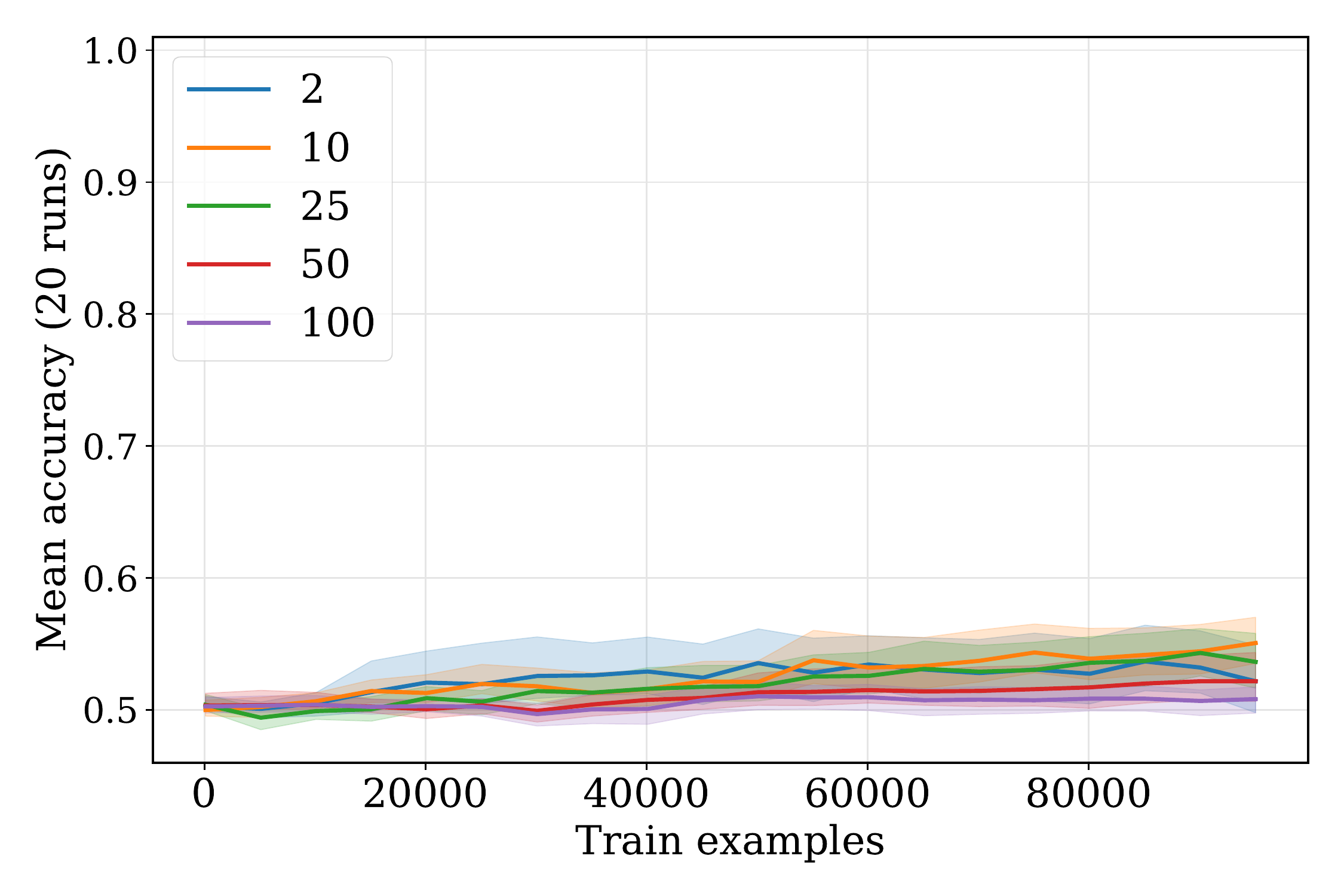}
    \caption{Hidden dimensionality 2.}
  \end{subfigure}
  \hfill
  \begin{subfigure}{0.45\linewidth}
    \includegraphics[width=1\textwidth]{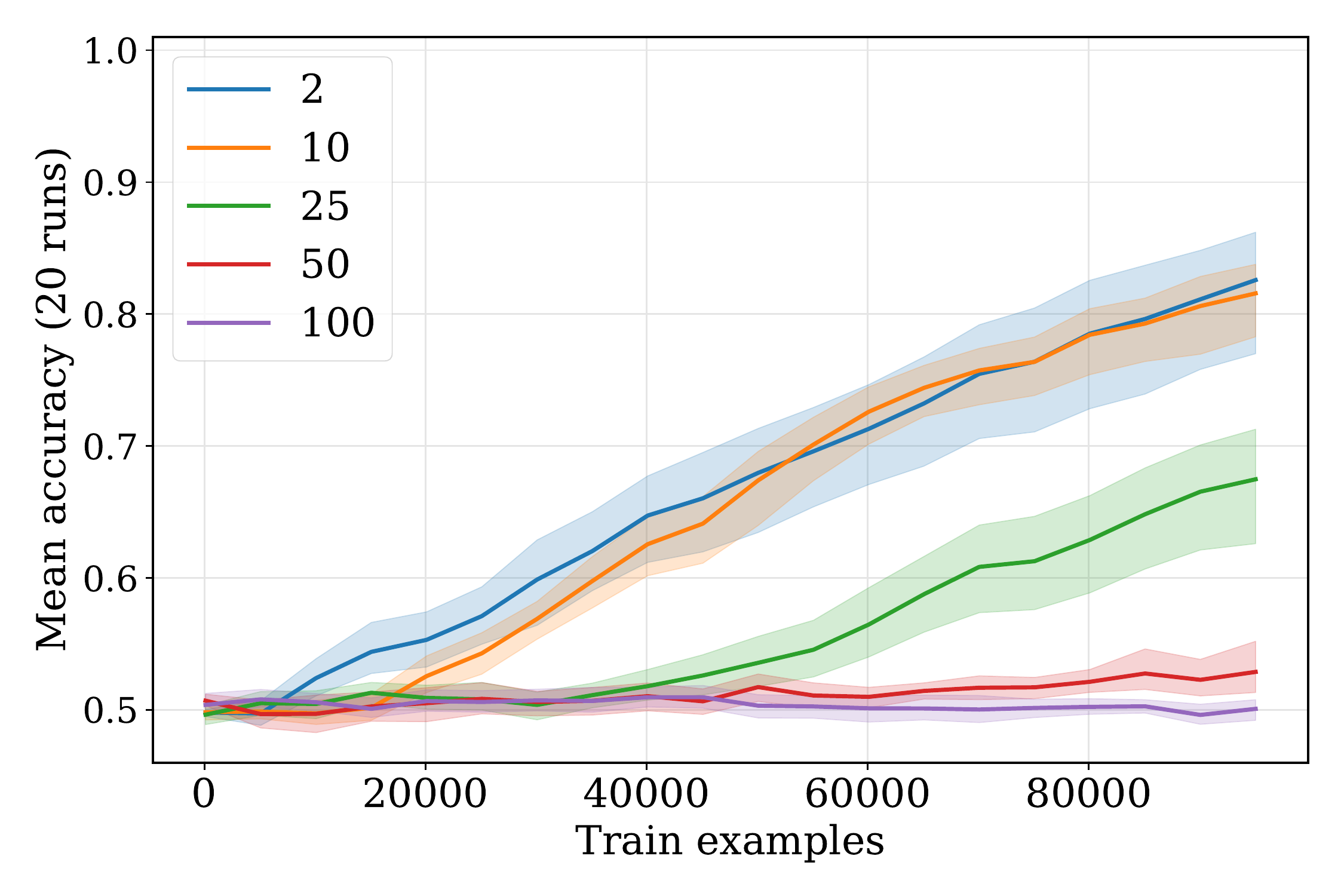}
    \caption{Hidden dimensionality 10.}
  \end{subfigure}

  \vspace{24pt}

  \begin{subfigure}{0.45\linewidth}
    \includegraphics[width=1\textwidth]{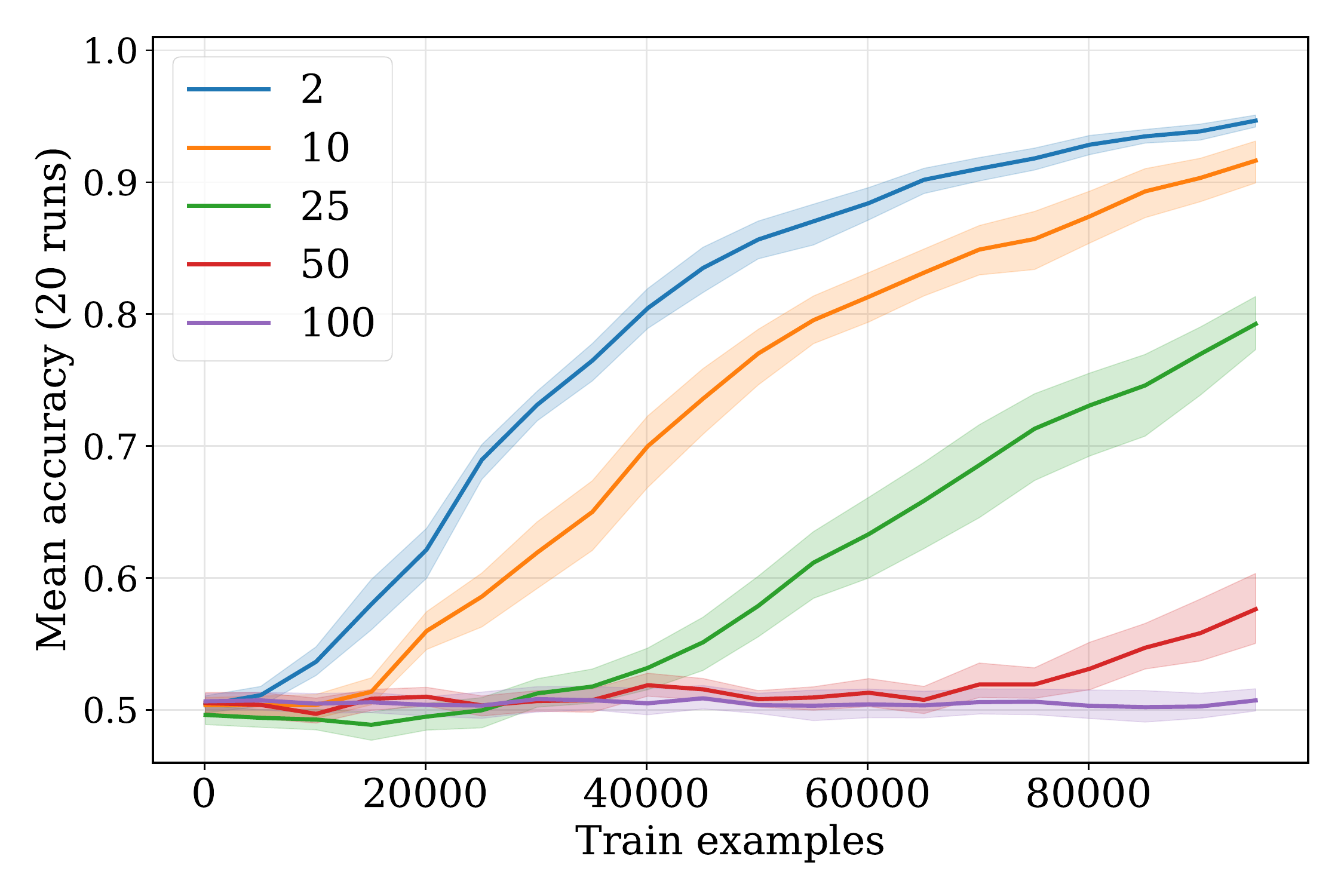}
    \caption{Hidden dimensionality 25.}
  \end{subfigure}
  \hfill
  \begin{subfigure}{0.45\linewidth}
    \includegraphics[width=1\textwidth]{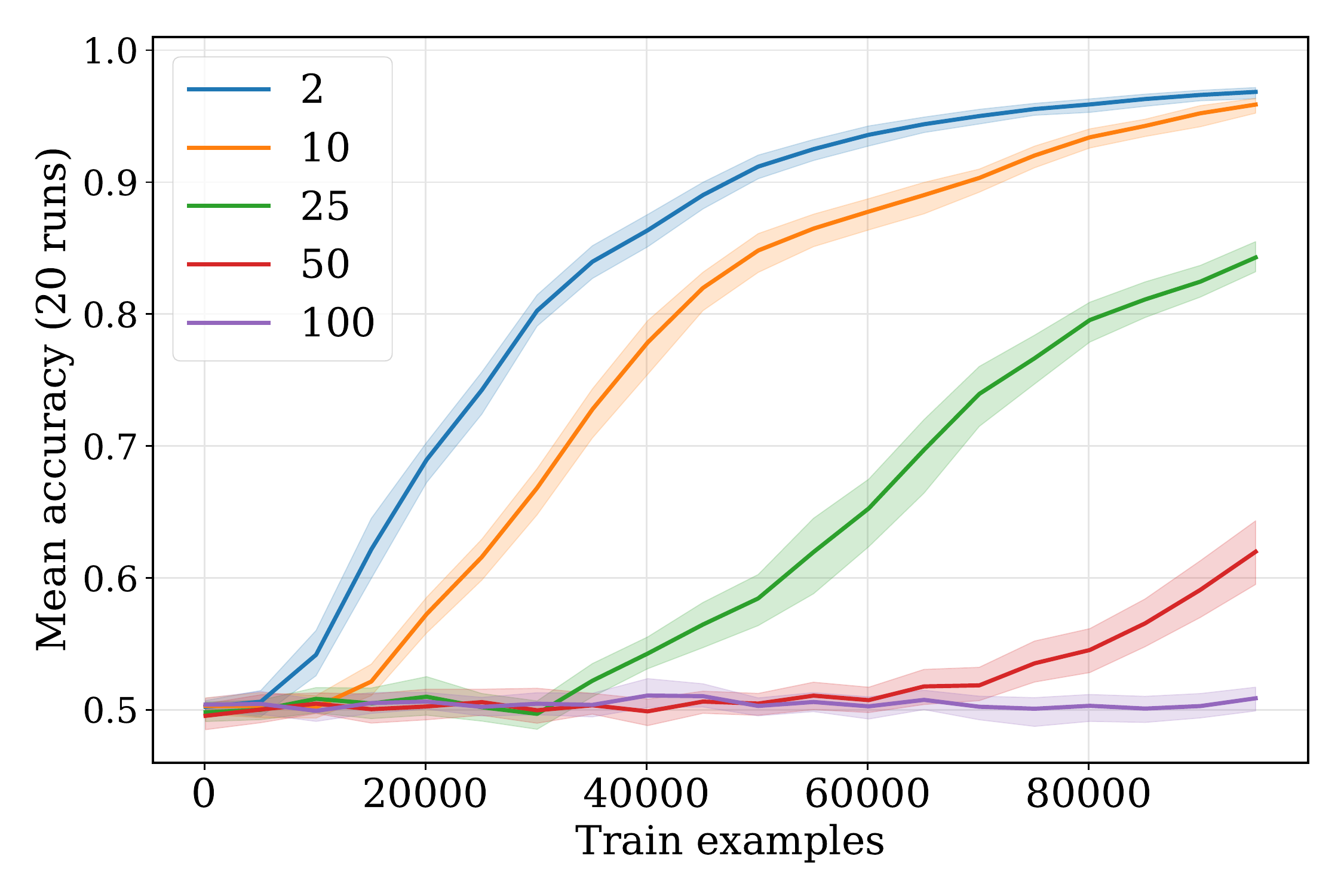}
    \caption{Hidden dimensionality 50.}
  \end{subfigure}

  \vspace{24pt}

  \begin{subfigure}{0.45\linewidth}
    \includegraphics[width=1\textwidth]{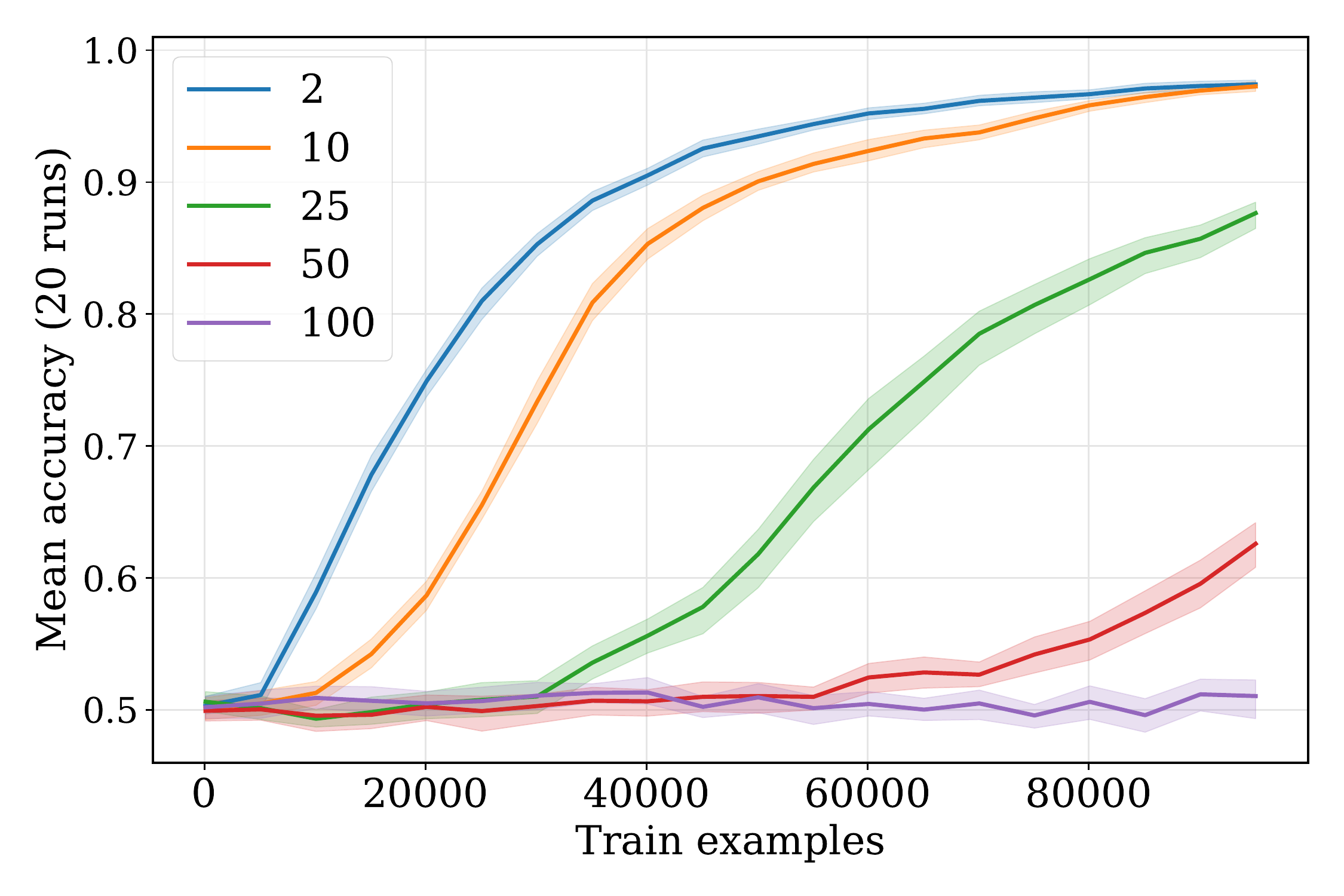}
    \caption{Hidden dimensionality 100.}
  \end{subfigure}
  \caption{Experiment~1 model applied to the hierarchical same--different task. Lines correspond to different input dimensions. These models were provided random input representations.}
  \label{fig:model1:premack}
\end{figure}

\Figref{fig:model1:premack} shows the results of applying the Experiment~1 model (equations~\dasheg{eq:x2h}{eq:h2y}) to the hierarchical same--different task. The only change from that model is that the inputs have dimensionality $4m$, since the four distinct representations in task inputs are simply concatenated. The lines correspond to different embedding dimensionalities.

\subsection{Experiment~3 results for different hidden dimensionalities}

\begin{figure}
  \centering

  \begin{subfigure}{0.45\linewidth}
    \includegraphics[width=1\textwidth]{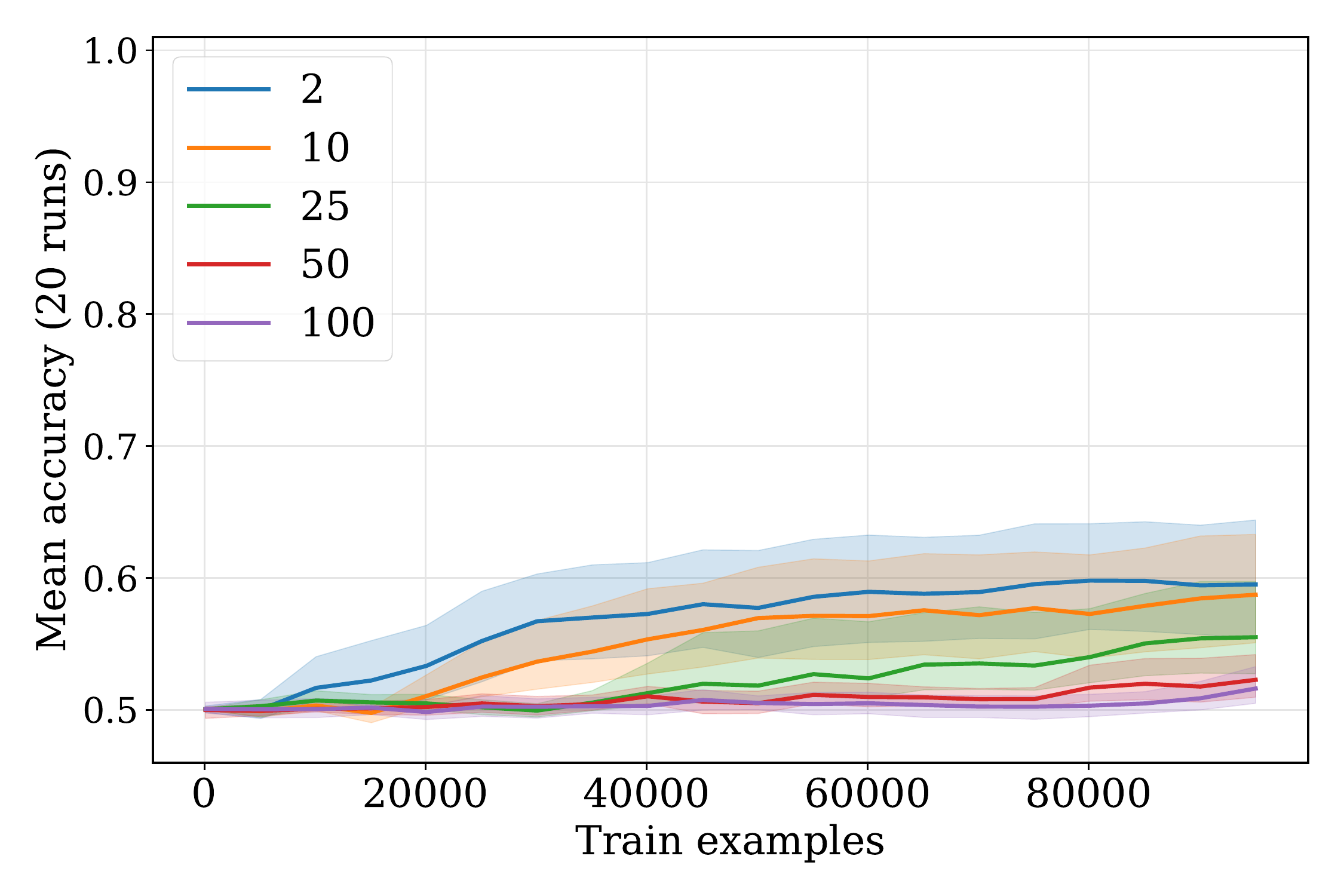}
    \caption{Hidden dimensionality 2.}
  \end{subfigure}
  \hfill
  \begin{subfigure}{0.45\linewidth}
    \includegraphics[width=1\textwidth]{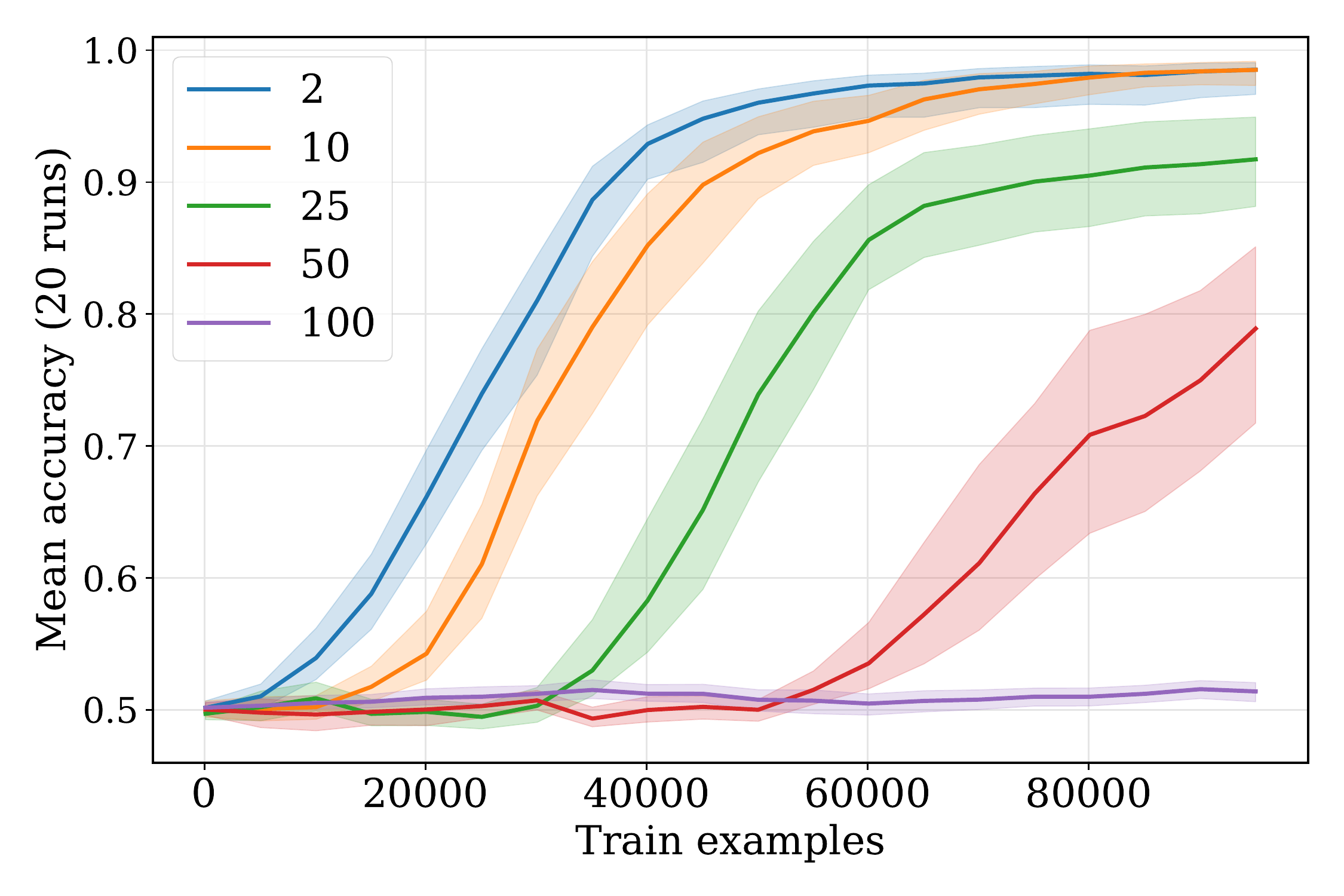}
    \caption{Hidden dimensionality 10.}
  \end{subfigure}

  \vspace{24pt}

  \begin{subfigure}{0.45\linewidth}
    \includegraphics[width=1\textwidth]{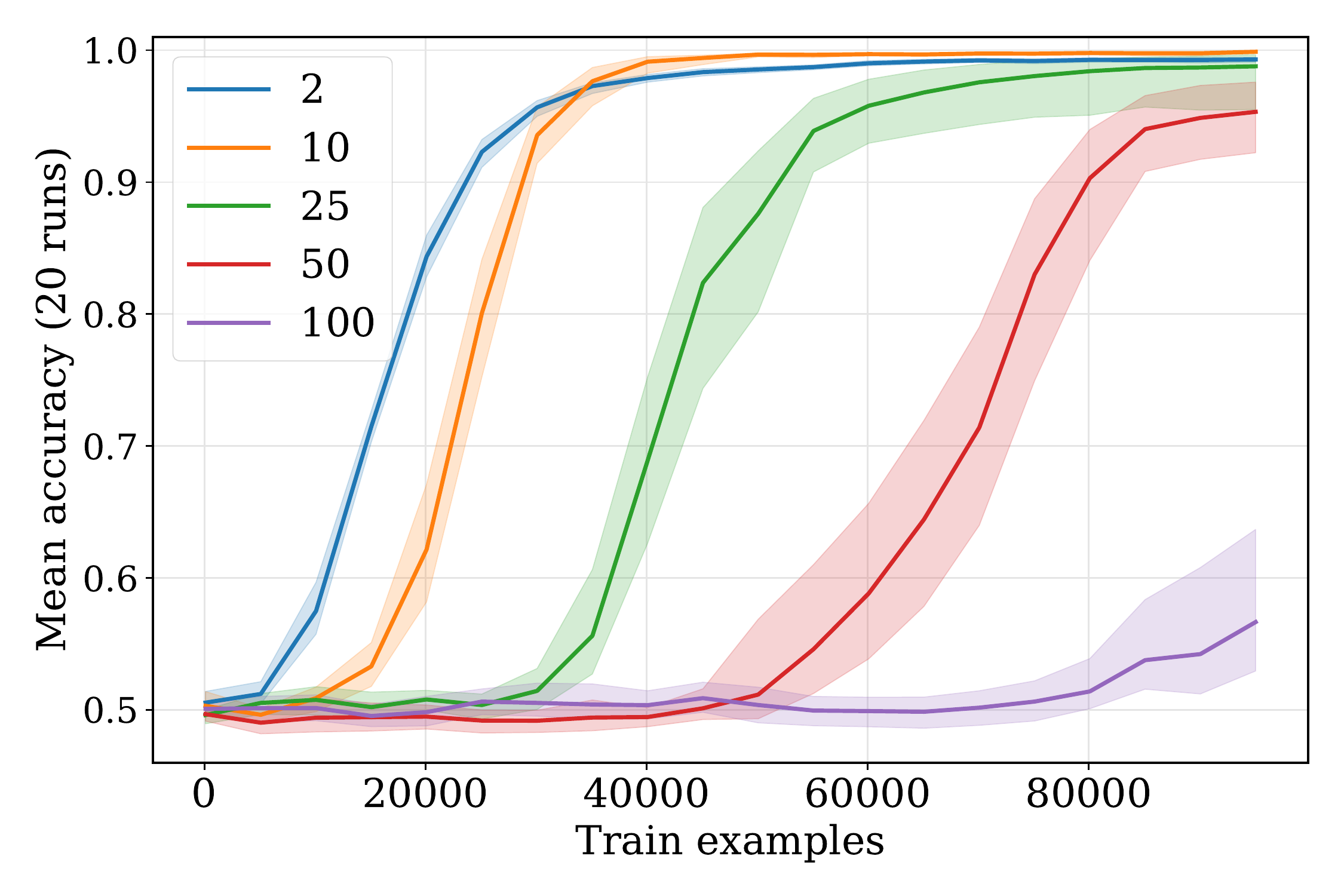}
    \caption{Hidden dimensionality 25.}
  \end{subfigure}
  \hfill
  \begin{subfigure}{0.45\linewidth}
    \includegraphics[width=1\textwidth]{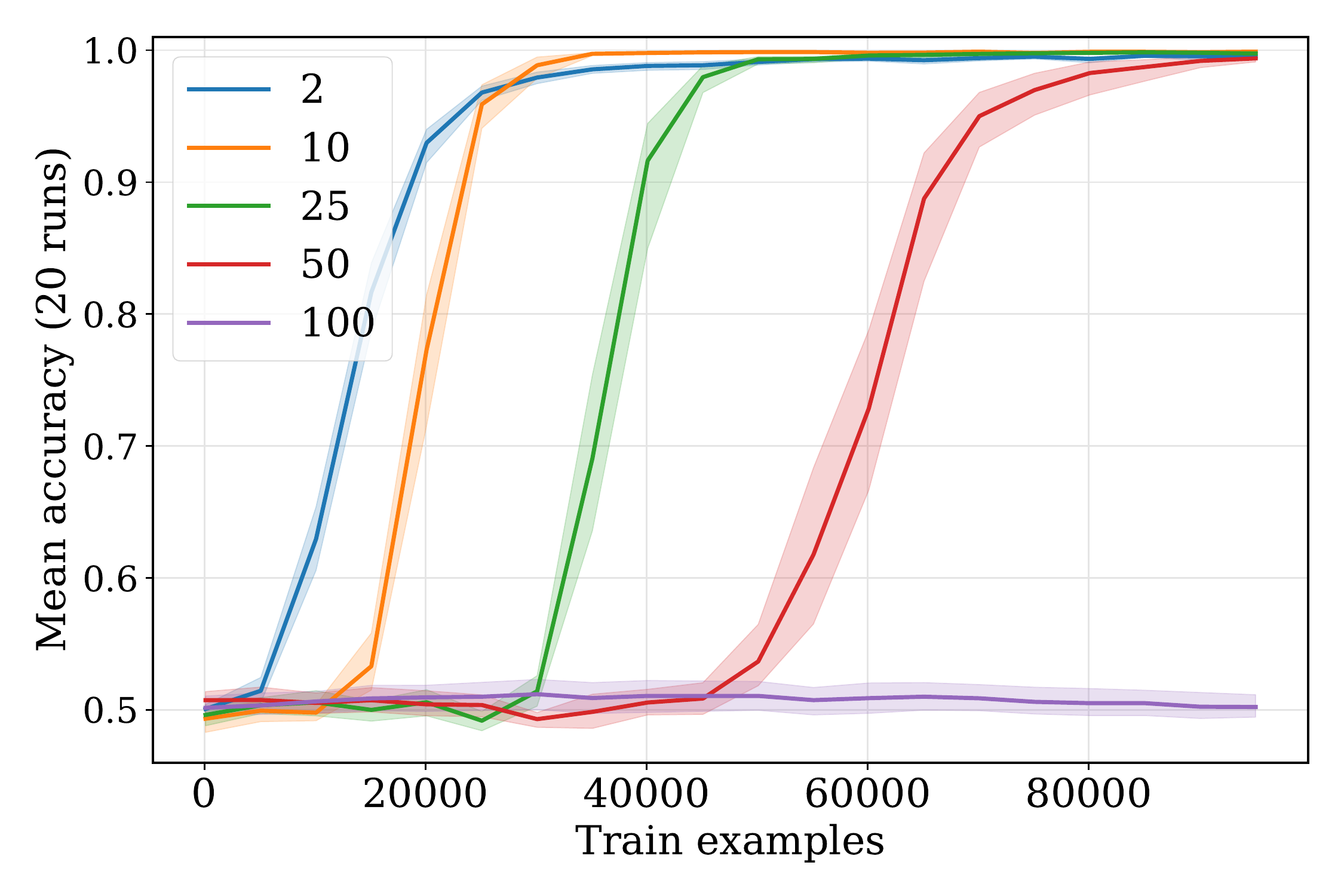}
    \caption{Hidden dimensionality 50.}
  \end{subfigure}

  \vspace{24pt}

  \begin{subfigure}{0.45\linewidth}
    \includegraphics[width=1\textwidth]{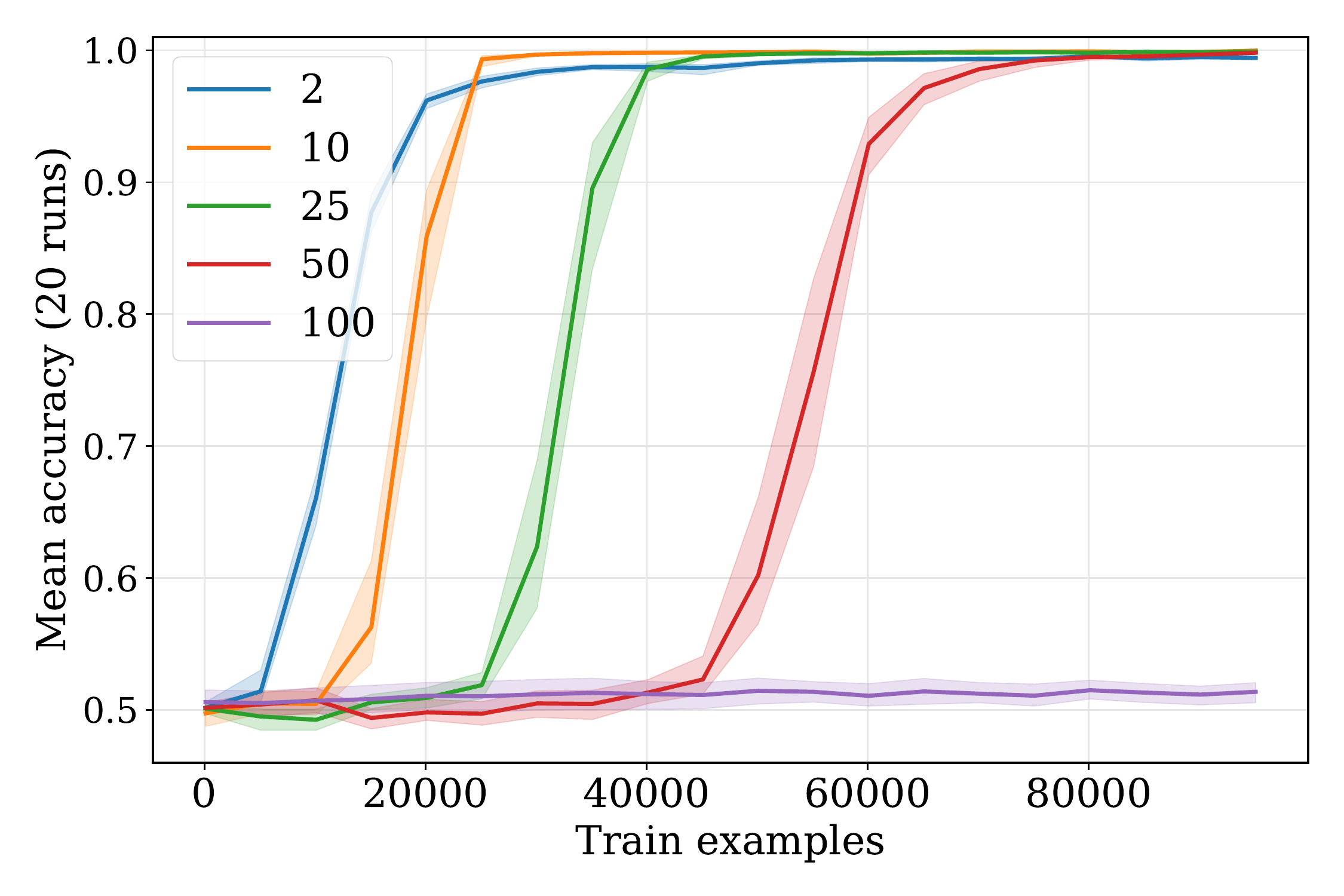}
    \caption{Hidden dimensionality 100.}
    \label{fig:model3a-rep}
  \end{subfigure}
  \caption{Experiment~3 model applied to the hierarchical same--different task. Lines correspond to different input dimensions. These models were provided random input representations.}
  \label{fig:model3a}
\end{figure}

\Figref{fig:model3a} shows the results of applying our Experiment~3 model (feed-forward network with two hidden layers; equations~\dasheg{eq:x2h1}{eq:h2y2}) to the hierarchical same--different task. The lines correspond to different embedding dimensionalities.

\subsection{Experiment~4 results for different input dimensionalities}

\begin{figure}
  \centering
  \includegraphics[width=0.6\textwidth]{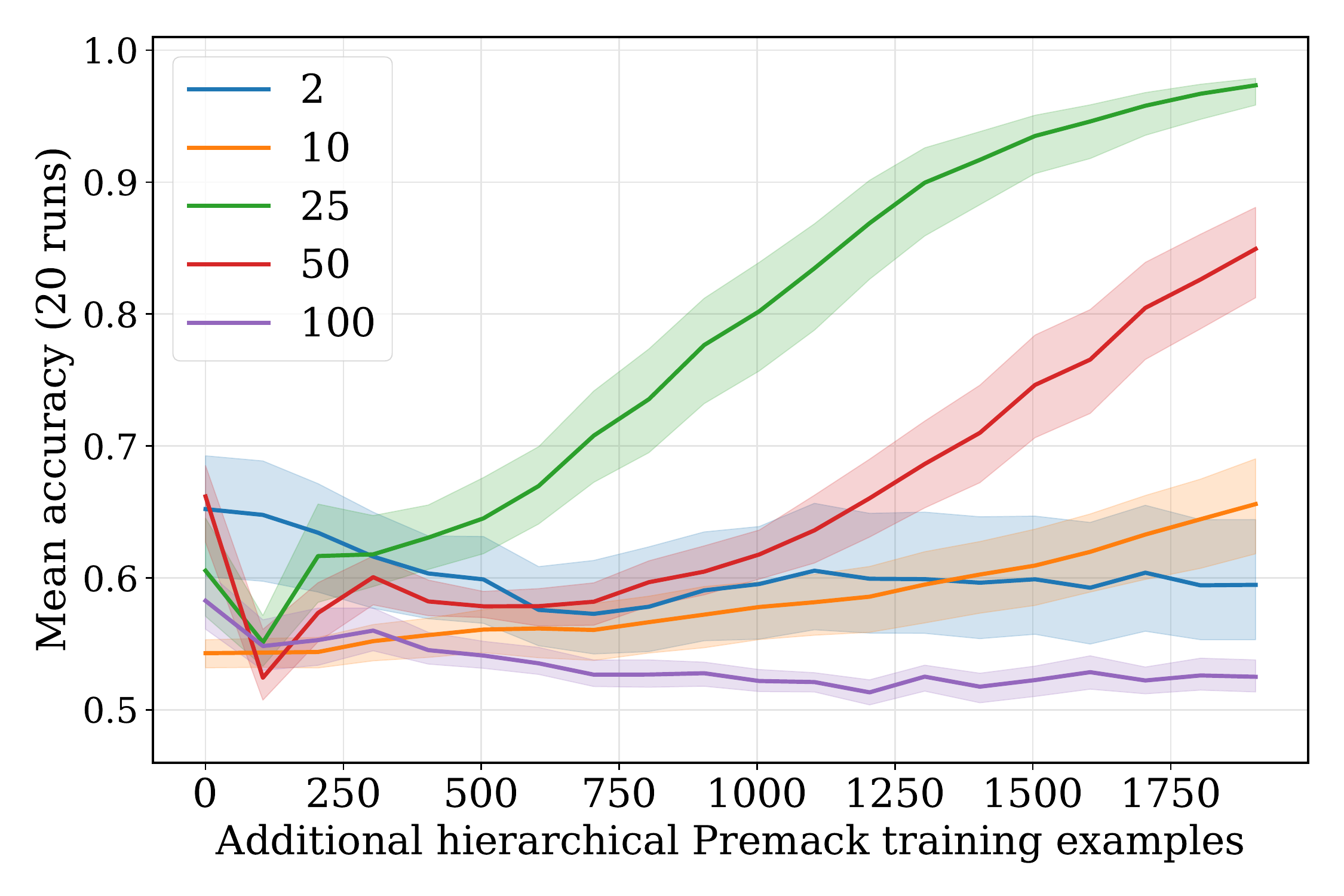}
  \caption{Experiment~4 model applied to the hierarchical same--different task. Lines correspond to different input dimensionality, which determine the hidden dimensionality. These models were provided random input representations.}
  \label{fig:model4}
\end{figure}

\Figref{fig:model4} shows the results of applying our Experiment~4 model (equations~\dasheg{eq:e4:1}{eq:e4:y}) to the hierarchical same--different task. In this model, the hidden dimensionality is required to be twice the input dimensionality, due to the way the components are the model are reused in a hierarchical fashion.

\subsection{Experiment~4 results by input class}

\begin{figure}
  \centering
  \includegraphics[width=0.6\textwidth]{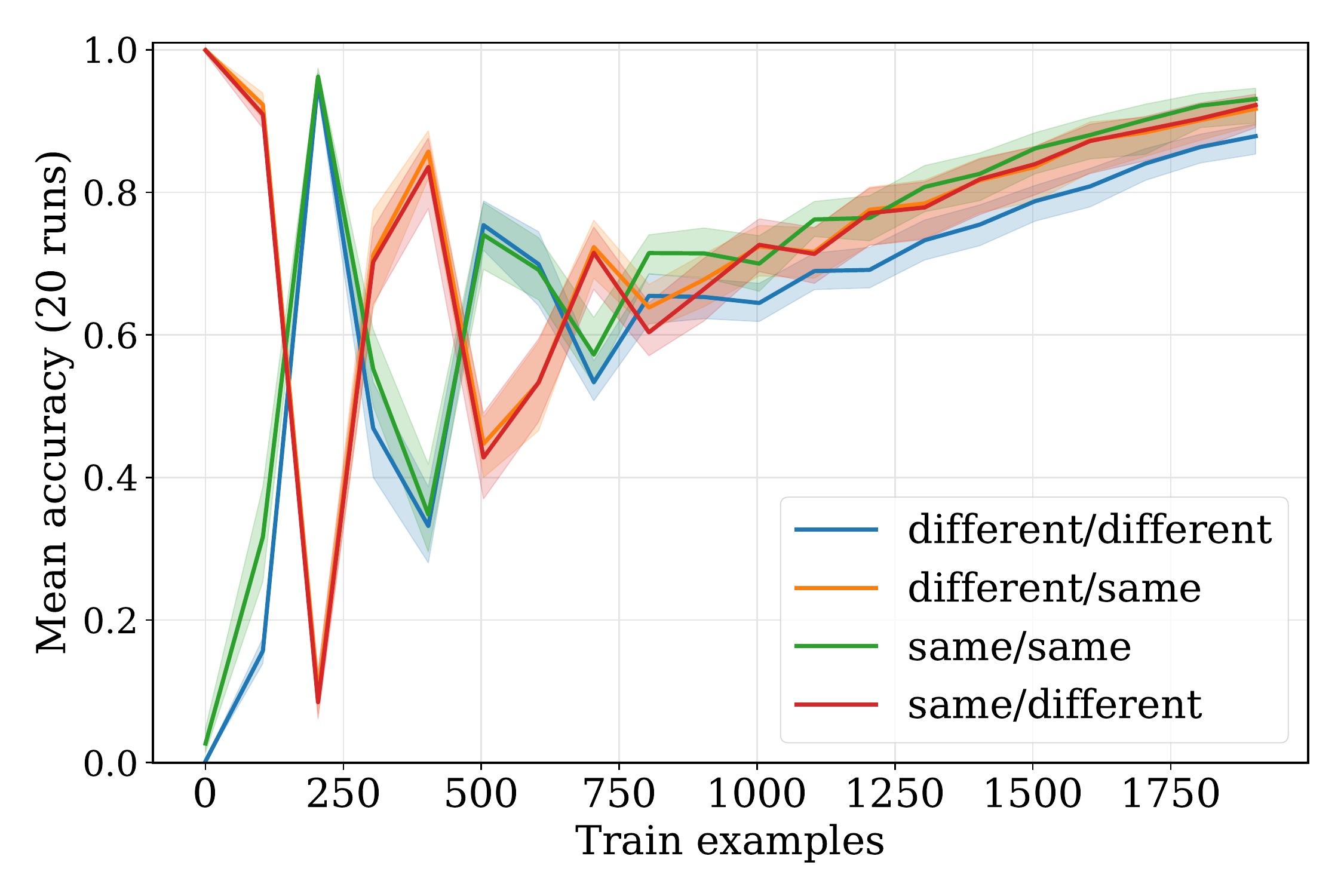}
  \caption{Experiment~4 model performance broken down by input class.}
  \label{fig:premack-by-class}
\end{figure}

The hierarchical same--different task intutively has four subclasses corresponding to different inputs:
\begin{itemize}
\item same/same: `positive'
\item different/different: `positive'
\item same/different: `negative'
\item different/same: `negative'
\end{itemize}
\Figref{fig:premack-by-class} depicts model accuracy for each of these input classes, for the `no pretraining' model depicted in \figref{fig:premack-pretraining-results}, which uses 25-dimensional input representations. It is noteworthy that the different/different class is the most difficult for the model.

\subsection{Performance on model train sets}\label{app:train-results}

\begin{figure}
  \centering

  \begin{subfigure}[t]{0.45\linewidth}
    \includegraphics[width=1\textwidth]{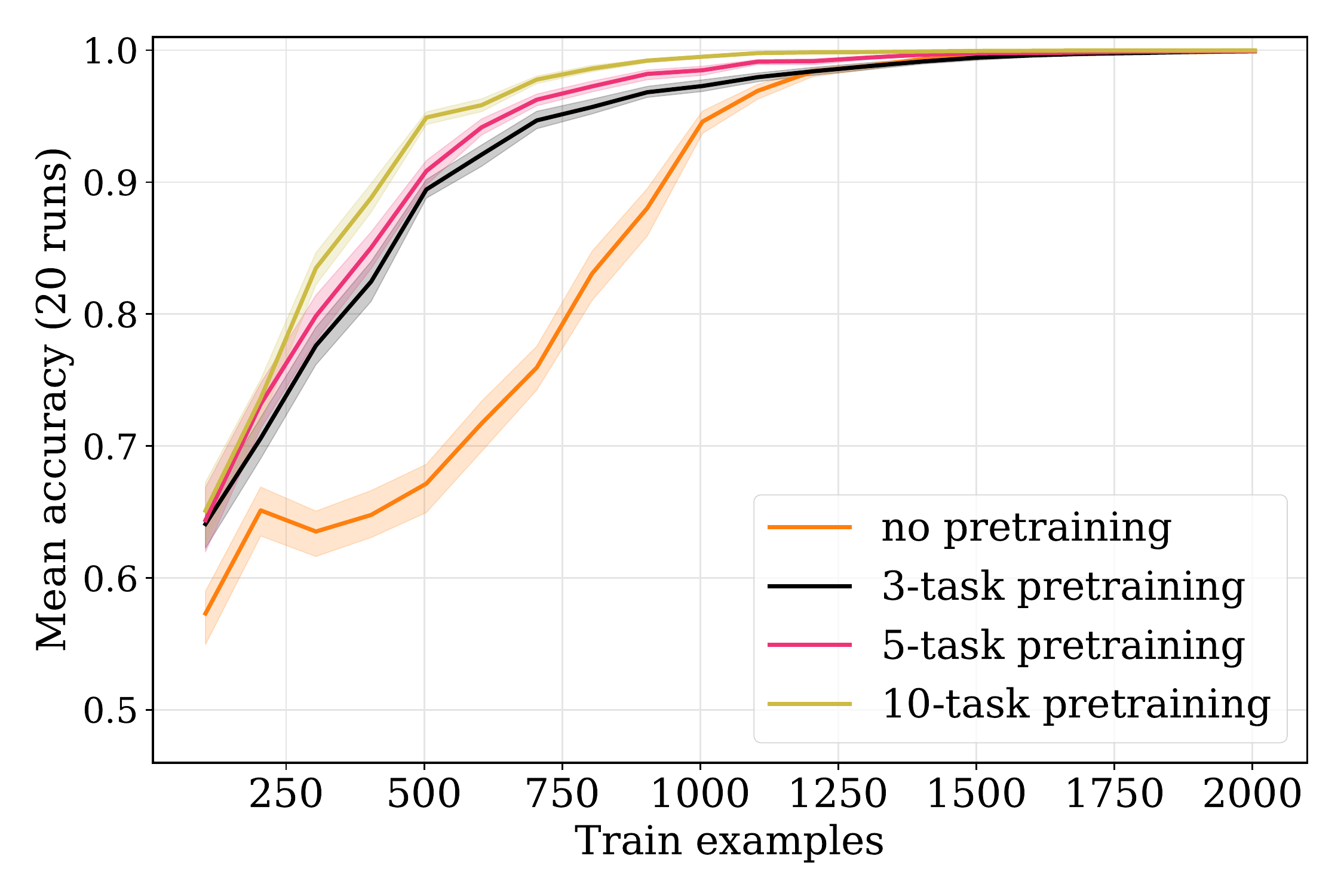}
    \caption{Experiment~1 training set results. These are qualitatively very much like our held-out assessment results (\figref{fig:basic-equality-pretrain}).}
    \label{fig:train-results:e1}
  \end{subfigure}
  \hfill
  \begin{subfigure}[t]{0.45\linewidth}
    \includegraphics[width=1\textwidth]{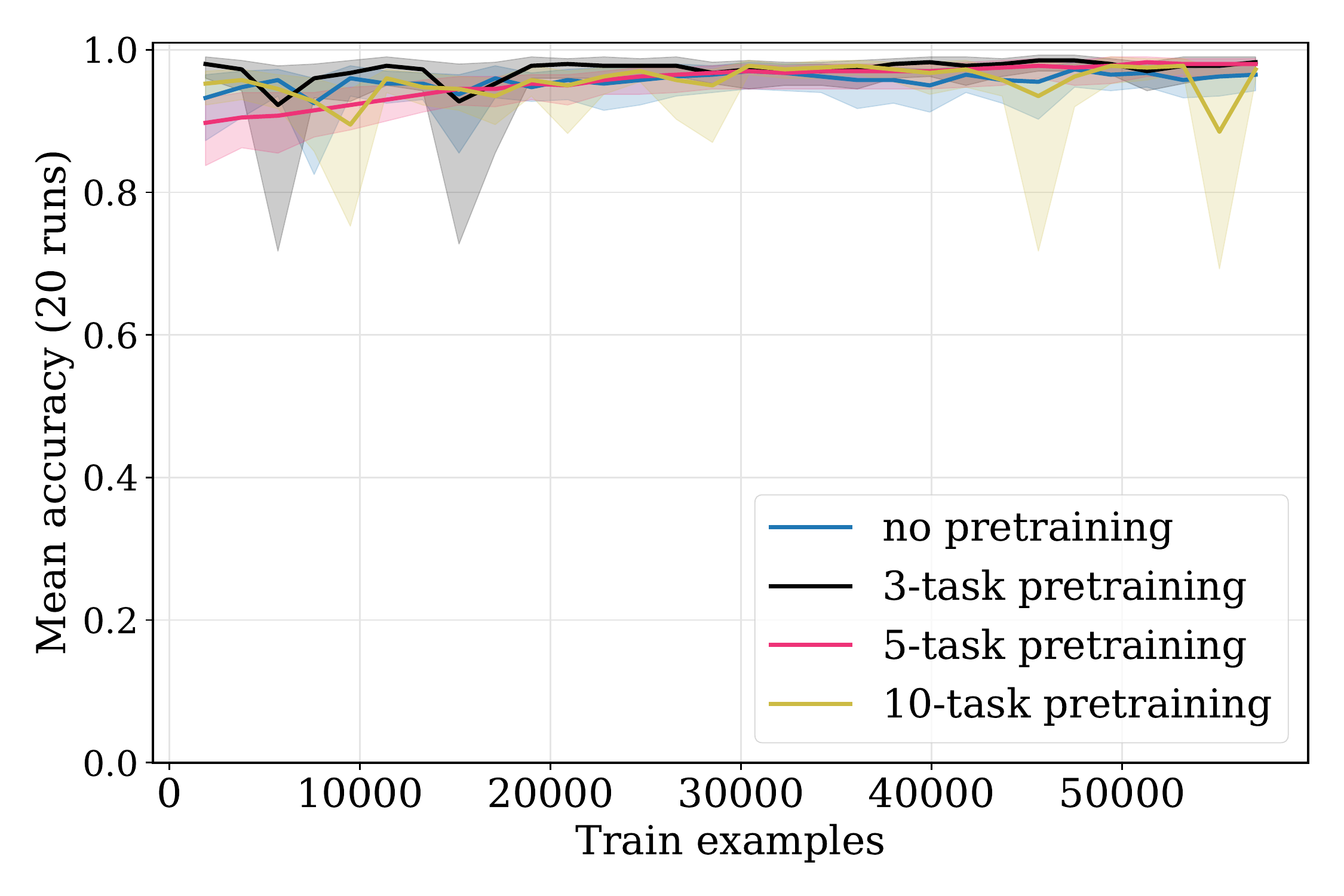}
    \caption{Experiment~2 training set results, indicating that these models very quickly memorize the training data, creating a large gap between training performance and test performance (cf.~\figref{fig:fuzzy-lm-pretrain-results}).}
    \label{fig:train-results:e2}
  \end{subfigure}

  \vspace{24pt}

  \begin{subfigure}[t]{0.45\linewidth}
    \includegraphics[width=1\textwidth]{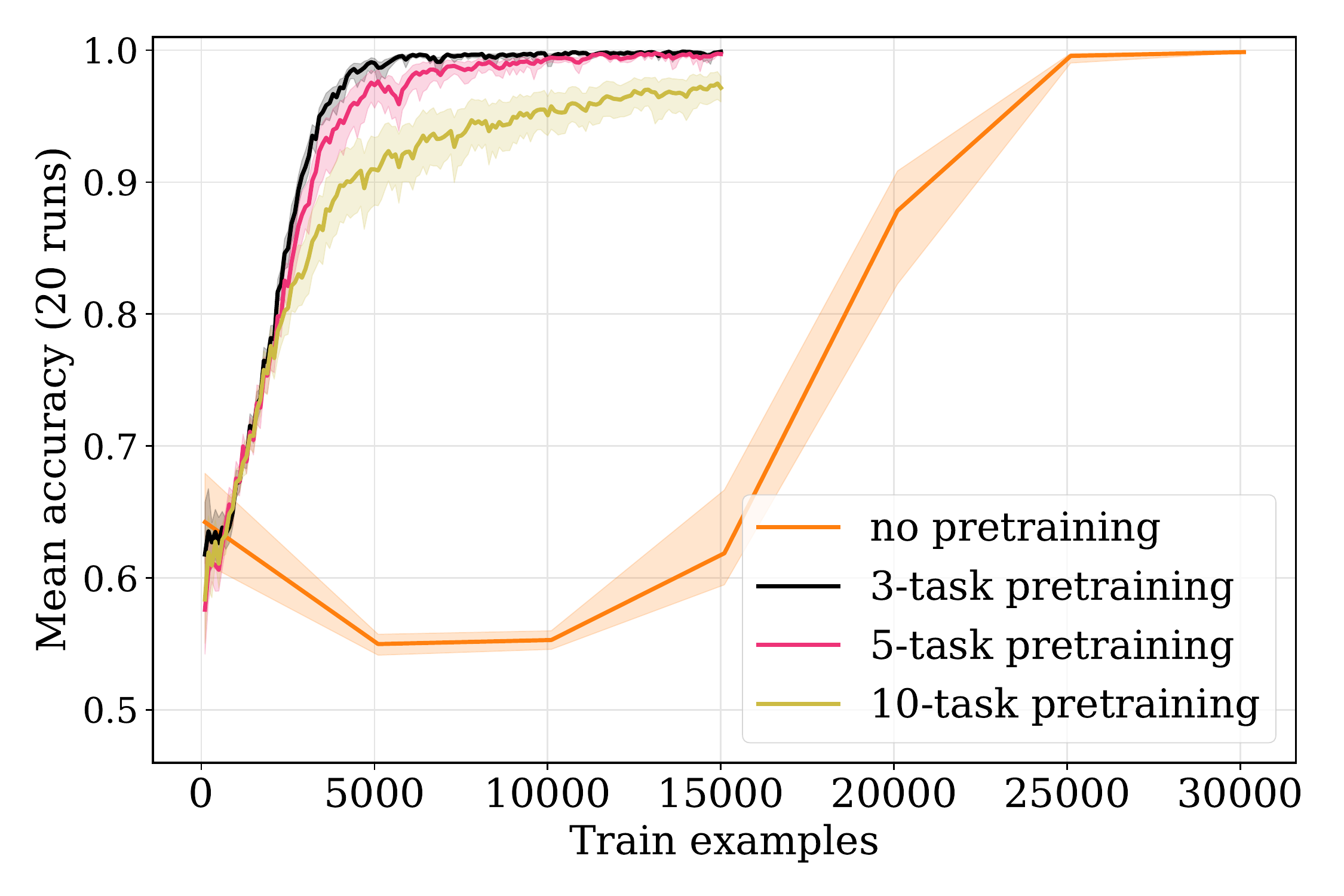}
    \caption{Experiment~3 training set results. These are qualitatively very much like our held-out assessment results (\figref{fig:premack-h2-pretrain}).}
    \label{fig:train-results:e3}
  \end{subfigure}
  \hfill
  \begin{subfigure}[t]{0.45\linewidth}
    \includegraphics[width=1\textwidth]{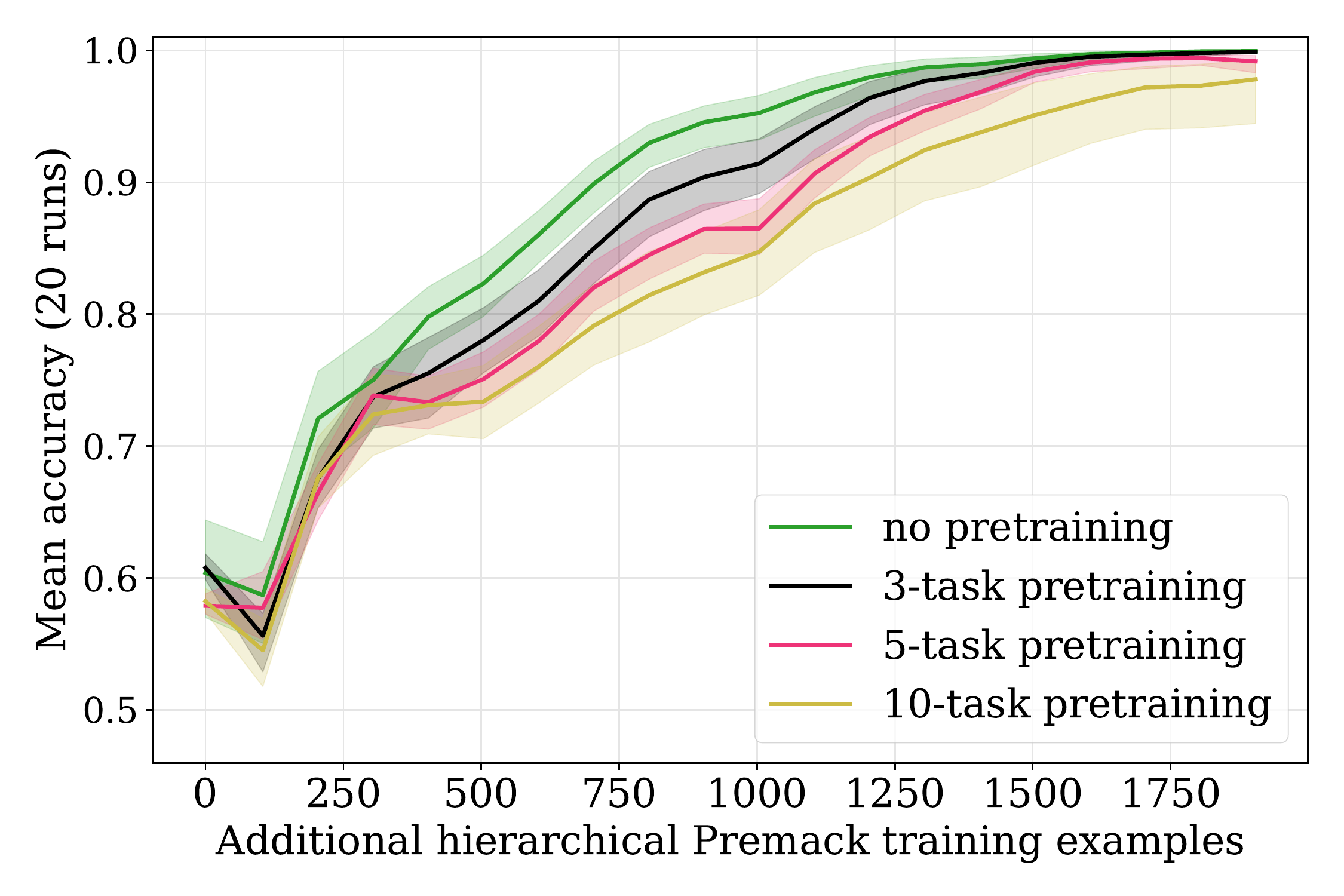}
    \caption{Experiment~4 training set results. These are very similar to our held-out assessments (\figref{fig:premack-pretraining-results}), with one noteworthy difference: pretraining provides small improvements in the held-out assessments, whereas `no pretraining' clearly provides the fastest learning in the training-set evaluations}
    \label{fig:train-results:e4}
  \end{subfigure}
  \caption{Training set results for the networks used in our primary assessments in the main text.}
  \label{fig:train-results}
\end{figure}

\Figref{fig:train-results} presents experimental results comparable to each of our main experiment figures, but with the evaluations now done on the training data.

\end{document}